\PassOptionsToPackage{table}{xcolor}

\documentclass[mlmain,onecolumn,cleveref]{jmlr}

\usepackage{longtable} 
\usepackage{booktabs}
\usepackage{ifthen}    

\theorembodyfont{\upshape}
\theoremheaderfont{\scshape}
\theorempostheader{:}
\theoremsep{\newline}

\newtheorem{assumption}[theorem]{Assumption}
\crefname{assumption}{assumption}{assumptions}
\Crefname{assumption}{Assumption}{Assumptions}

\jmlrvolume{}
\firstpageno{1}
\editors{Louisa Cornelis, Francisco Acosta, Alison Pouplin, David Klindt, Fatih Dinc, Nina Miolane}

\jmlrproceedings{PMLR}{Proceedings of Machine Learning Research}

\jmlryear{2026}
\jmlrworkshop{Symmetry and Geometry in Neural Representations}

\title[Heterogeneous Sheaf Neural Networks]{Heterogeneous Sheaf Neural Networks}

\makeatletter
\renewcommand*{\thanks}[1]{%
  \refstepcounter{mpfootnote}%
  \footnotemark[\number\value{mpfootnote}]%
  \xappto\@thanks{\noexpand\footnotetext[\number\value{mpfootnote}]{\unexpanded{#1}}}%
}
\makeatother

\author[Braithwaite, Borgi, Onorato, Tarantelli et al.]{
  \Name{Luke Braithwaite\nametag{\textsuperscript{1,}\footnotemark[1]}}\,
  \Name{Alessio Borgi\nametag{\textsuperscript{1,3,\textdagger}\thanks{Equal contribution. \textsuperscript{\textdagger}Corresponding authors: Alessio Borgi: \texttt{alessio.borgi@uniroma1.it, ab3352@cam.ac.uk}, Gabriele Onorato: \texttt{onorato.g@northeastern.edu}. Code is available at \url{https://github.com/alessioborgi/Heterogeneous-Sheaf-Neural-Networks}.}}}\,
  \Name{Gabriele Onorato\nametag{\textsuperscript{2,\textdagger,}\footnotemark[1]}}\,
  \Name{Kristjan \\ Tarantelli\nametag{\textsuperscript{3,}\footnotemark[1]}}\,
  \Name{Francesco Restuccia\nametag{\textsuperscript{2}}}\,
  \Name{Fabrizio Silvestri\nametag{\textsuperscript{3}}}\,
  \Name{Pietro Li\`o\nametag{\textsuperscript{1}}}
  \\
  \addr \textsuperscript{1}Department of Computer Science and Technology, University of Cambridge \\
  \addr \textsuperscript{2}Department of Electrical and Computer Engineering, Northeastern University \\
  \addr \textsuperscript{3}Department of Computer, Control and Management Engineering, Sapienza University of Rome
}

\usepackage[utf8]{inputenc}
\usepackage{lmodern}
\usepackage[T1]{fontenc}
\usefont{T1}{lmr}{m}{n}
\DeclareFontShape{T1}{lmr}{bx}{sc}{<->ssub*cmr/bx/sc}{}
\usepackage{amsfonts}
\usepackage{nicefrac}
\usepackage{microtype}
\usepackage{wrapfig}
\usepackage{import}
\usepackage{pifont}

\usepackage{mathtools}
\allowdisplaybreaks
\usepackage{bm}
\usepackage{fix-cm}
\usepackage[separate-uncertainty=true]{siunitx}
\usepackage{xfrac}

\usepackage{makecell}
\usepackage{multicol}
\usepackage{multirow}
\usepackage{adjustbox}
\usepackage{tikz}
\usetikzlibrary{arrows.meta,positioning,fit,calc,backgrounds,decorations.pathreplacing,shapes.geometric}
\usepackage{etoc}

\usepackage{enumitem}
\usepackage{csquotes}

\usepackage[most]{tcolorbox}
\tcbuselibrary{skins, breakable}

\definecolor{proofbar}{RGB}{225,225,225}
\definecolor{theorembar}{RGB}{128, 0, 128}
\definecolor{propositionbar}{RGB}{0, 100, 0}
\definecolor{definitionbar}{RGB}{0, 0, 200}
\definecolor{lemmabar}{RGB}{255, 140, 0}
\definecolor{corollarybar}{RGB}{204, 153, 255}
\definecolor{remarkbar}{RGB}{0, 139, 139}
\definecolor{assumptionbar}{RGB}{178, 34, 34}

\tcbset{
    mybarstyle/.style={
        enhanced,
        standard jigsaw, 
        sharp corners,
        breakable,
        boxrule=0pt,        
        leftrule=2pt,       
        colframe=#1,        
        opacityback=0,      
        colback=white,
        left=6pt,           
        right=0pt,
        top=0pt,
        bottom=0pt,
        before skip=10pt,
        after skip=10pt,
    }
}

\tcolorboxenvironment{theorem}{mybarstyle=theorembar}
\tcolorboxenvironment{proposition}{mybarstyle=propositionbar}
\tcolorboxenvironment{lemma}{mybarstyle=lemmabar}
\tcolorboxenvironment{corollary}{mybarstyle=corollarybar}
\tcolorboxenvironment{definition}{mybarstyle=definitionbar}
\tcolorboxenvironment{remark}{mybarstyle=remarkbar}
\tcolorboxenvironment{assumption}{mybarstyle=assumptionbar}
\tcolorboxenvironment{proof}{mybarstyle=proofbar}

\newboolean{showcomments}
\setboolean{showcomments}{true}

\definecolor{Top1}{HTML}{0072B2} 
\definecolor{Top2}{HTML}{E69F00} 
\definecolor{Top3}{HTML}{CC79A7}

\definecolor{ufill}{HTML}{E1D5E7}
\definecolor{uborder}{HTML}{9673A6}
\definecolor{vfill}{HTML}{D5E8D4}
\definecolor{vborder}{HTML}{82B366}

\makeatletter
\long\def\@makecaption#1#2{%
   \vskip 6pt
   \setbox\@tempboxa\hbox{#1: #2}%
   \ifdim \wd\@tempboxa >\hsize
       \begin{list}{#1:}{%
       \settowidth{\labelwidth}{#1:}
       \setlength{\leftmargin}{\labelwidth}
       \addtolength{\leftmargin}{\labelsep}
       \setlength{\topsep}{0pt}
        }\item #2 \end{list}\par
     \else
       \hbox to\hsize{\hfil\box\@tempboxa\hfil}
   \fi}
\renewcommand{\section}{\@startsection{section}{1}{\z@}%
   {-0.18in \@plus -0.5ex \@minus -.2ex}%
   {0.07in \@plus.2ex}%
   {\normalfont\rmfamily\bfseries\large\raggedright}}
\renewcommand\subsection{\@startsection{subsection}{2}{\z@}%
   {-0.15in \@plus -0.5ex \@minus -.2ex}%
   {0.05in \@plus .2ex}%
   {\normalfont\rmfamily\bfseries\normalsize\raggedright}}
\appto\normalsize{%
  \setlength{\abovedisplayskip}{6pt plus 2pt minus 2pt}%
  \setlength{\belowdisplayskip}{6pt plus 2pt minus 2pt}%
  \setlength{\abovedisplayshortskip}{2pt plus 2pt}%
  \setlength{\belowdisplayshortskip}{4pt plus 2pt minus 2pt}}
\makeatother
\setlist{topsep=3pt,itemsep=2pt,parsep=0pt}

\usepackage{bm}

\def\eqref#1{equation~\ref{#1}}

\def\1{\bm{1}}

\def\rva{{\mathbf{a}}}

\def\rve{{\mathbf{e}}}

\def\rvg{{\mathbf{g}}}
\def\rvh{{\mathbf{h}}}
\def\rvu{{\mathbf{i}}}

\def\rvo{{\mathbf{o}}}

\def\rvu{{\mathbf{u}}}

\def\rvx{{\mathbf{x}}}
\def\rvy{{\mathbf{y}}}
\def\rvz{{\mathbf{z}}}

\def\rmA{{\mathbf{A}}}

\def\rmC{{\mathbf{C}}}
\def\rmD{{\mathbf{D}}}

\def\rmG{{\mathbf{G}}}
\def\rmH{{\mathbf{H}}}
\def\rmI{{\mathbf{I}}}

\def\rmM{{\mathbf{M}}}

\def\rmP{{\mathbf{P}}}
\def\rmQ{{\mathbf{Q}}}
\def\rmR{{\mathbf{R}}}
\def\rmS{{\mathbf{S}}}
\def\rmT{{\mathbf{T}}}
\def\rmU{{\mathbf{U}}}
\def\rmV{{\mathbf{V}}}
\def\rmW{{\mathbf{W}}}
\def\rmX{{\mathbf{X}}}

\def\rmZ{{\mathbf{Z}}}

\DeclareMathAlphabet{\mathsfit}{\encodingdefault}{\sfdefault}{m}{sl}

\def\gE{{\mathcal{E}}}
\def\gF{{\mathcal{F}}}
\def\gG{{\mathcal{G}}}

\def\gV{{\mathcal{V}}}

\def\sR{{\mathbb{R}}}

\DeclareMathOperator*{\concat}{%
    \mathchoice%
        {\Big\Vert}%
        {\big\Vert}%
        {\Vert}%
        {\Vert}%
}

\DeclarePairedDelimiter{\parens}{\lparen}{\rparen}

\DeclarePairedDelimiter{\norm}{\lVert}{\rVert}
\DeclarePairedDelimiter{\kets}{\langle}{\rangle}
\DeclarePairedDelimiter{\brkts}{\lbrack}{\rbrack}
\DeclarePairedDelimiter{\braces}{\lbrace}{\rbrace}

\DeclareMathOperator{\MLP}{MLP}

\DeclareMathOperator{\x}{\mathbf{x}}

\DeclareMathOperator{\h}{\mathbf{h}}

\definecolor{slate}{HTML}{0F172A}
\definecolor{body}{HTML}{334155}
\definecolor{muted}{HTML}{475569}
\definecolor{linec}{HTML}{CBD5E1}

\definecolor{bluebg}{HTML}{EEF6FF}
\definecolor{bluebd}{HTML}{60A5FA}
\definecolor{greenbg}{HTML}{EFFCF6}
\definecolor{greenbd}{HTML}{34D399}
\definecolor{orangebg}{HTML}{FFF7ED}
\definecolor{orangebd}{HTML}{FB923C}
\definecolor{purplebg}{HTML}{F5F3FF}
\definecolor{purplebd}{HTML}{A78BFA}
\definecolor{redbg}{HTML}{FEF2F2}
\definecolor{redbd}{HTML}{F87171}
\definecolor{graybg}{HTML}{F8FAFC}
\definecolor{graybd}{HTML}{64748B}

\definecolor{typeA}{HTML}{A78BFA}
\definecolor{typeB}{HTML}{7CC96B}
\definecolor{typeC}{HTML}{F3A74F}
\definecolor{typeD}{HTML}{E87BB5}
\definecolor{typeX}{HTML}{76A7FA}

\tikzset{
  >={Latex[length=3mm,width=2mm]},
  bigpanel/.style={rounded corners=8pt, draw=linec, fill=white, line width=0.8pt},
  stepbox/.style 2 args={rounded corners=10pt, draw=#1, fill=white, line width=1pt},
  paneltitle/.style={font=\bfseries\fontsize{16}{18}\selectfont, text=slate},
  steptitle/.style={font=\bfseries\fontsize{8.7}{10.2}\selectfont, text=slate, align=center},
  bodytext/.style={font=\fontsize{6.9}{8.1}\selectfont, text=body, align=center},
  smalltext/.style={font=\fontsize{6.7}{7.9}\selectfont, text=body, align=center},
  tinytext/.style={font=\fontsize{6.2}{7.2}\selectfont, text=muted, align=center},
  formulatxt/.style={font=\fontsize{10}{11}\selectfont\ttfamily, text=slate, align=center},
  nodecircle/.style={circle, minimum size=8.5mm, inner sep=0pt, draw=#1!80!black, fill=#1!55, line width=0.8pt},
  mtx/.style={draw=#1!80!black, fill=#1!18, rounded corners=1.5pt, line width=0.7pt},
  subpanel/.style={rounded corners=6pt, draw=linec, fill=white, line width=0.7pt},
  propbox/.style 2 args={rounded corners=8pt, draw=#1, fill=white, line width=0.9pt}
}

\newcommand{\pmstd}[1]{{\scriptsize $\pm$#1}}
\newcommand{\ablationlegend}{Best restriction map per sheaf learner \colorbox{Top1!30!cyan!40}{highlighted}; best sheaf learner in \textbf{bold}.}
\setcitestyle{numbers,square,comma}

\begin{document}
\maketitle

\begin{abstract}
Heterogeneous graphs, whose nodes and edges can belong to different types and feature spaces, arise in many real-world domains, including biology, recommendation, social networks, and computer systems. Existing heterogeneous graph neural networks typically handle this heterogeneity at the architectural level through relation-specific modules, meta-path machinery or type-aware attention, which often leads to increasingly specialised, parameter-heavy designs. In this work, we propose \textsc{HetSheaf}, a framework for learning heterogeneous graphs through \emph{cellular sheaves}. Instead of encoding heterogeneity solely in the architecture, \textsc{HetSheaf} represents it directly in the underlying data structure by assigning type-aware local feature spaces and learning restriction maps conditioned on node features, node types, and edge types. To support graph-level prediction, we further introduce \textsc{SheafPool}, a universal stalk-space readout that aggregates node representations while being invariant to local changes of basis, thereby making graph classification with sheaf networks well-defined and achieving an F1 Score up to 42 percentage points higher than mean pooling. Across a diverse suite of benchmarks (node classification, link prediction and graph classification), \textsc{HetSheaf} consistently achieves up to 2 percentage points higher performance (up to 94.93\% Macro F1 Score on node classification and up to 99.64\% on link prediction) on the Heterogeneous Graph Benchmark (HGB) framework against homogeneous (GCN, GAT, GIN, GraphSAGE), heterogeneous (R-GCN, HAN, HGT) and type-agnostic sheaf baselines, while reducing the number of parameters by up to 10$\times$.
\end{abstract}


\section{Introduction}
Graphs are a natural representation in many scientific fields, with Graph Neural Networks (GNNs) being one of the main tools to model relational data~\citep{goller1996learning,gori2005new,scarselli2008graph,bruna2013spectral,defferrard2016convolutional,velickovicGraphAttentionNetworks2018,gilmerNeuralMessagePassing2017}. 
Traditional GNNs implicitly assume that the underlying graph is \emph{homogeneous}; i.e., all nodes belong to a single category and share the same feature space, and all edges represent the same kind of relationship. However, many real-world graphs are \emph{heterogeneous}; nodes and edges may have different types, different meanings, and even different input dimensionalities ~\citep{fanGraphNeuralNetworks2019,qiuDeepInfSocialInfluence2018,liEncodingSocialInformation2019,guoAttentionBasedSpatialTemporal2019,huGPTGNNGenerativePreTraining2020,yangMultiSageEmpoweringGCN2020}. Consequently, treating graphs as homogeneous limits the model's ability to capture complex patterns, leading to sub-optimal performances. To overcome these limitations, \emph{heterogeneous GNNs} (HGNNs) enrich standard GNN architectures with edge-specific transformations, hand-crafted meta-path guided aggregations or attention pipelines~\citep{schlichtkrullModelingRelationalData2018,zhangHeterogeneousGraphNeural2019,huHeterogeneousGraphTransformer2020,wangHeterogeneousGraphAttention2019a}. While effective, this work typically handles heterogeneity through \emph{ad-hoc modifications}, which often require more relation-specific components in the architecture, more hand-crafted design choices, or more parameters, as the as the number
of node and edge types grows. Second, the burden of representing heterogeneous structure is placed on the neural network itself instead of on the underlying data representation. In other words, the graph topology is converted to its homogeneous variant and the model must compensate for the heterogeneity of local spaces. 
\begin{figure}[!htbp]
\floatconts
  {fig:pipeline}
  {\caption{\emph{(a)} Conventional Heterogeneous GNNs encode type information through specialised architectural effects. \emph{(b)} Sheaf Neural Networks (SheafNNs) require a homogeneous graph representation; this can be artificially constructed by encoding node types into the feature vectors $\mathbf{x}_u$ and $\mathbf{x}_v$ to infer the restriction maps $\gF_{u \trianglelefteq e \coloneqq (u,v)}$  between stalks $\gF_u$ and $\gF_v$. \emph{(c)} \textsc{HetSheaf} maps type-dependent features $\mathbf{x}_{u|v}$ to a shared space $\rvz_{u|v}$ and natively handles heterogeneity by explicitly parameterising the sheaf's restriction maps with the type information of nodes $\phi(\cdot)$ and edges $\psi(\cdot)$.}}
  {\includegraphics[width=1.0\linewidth]{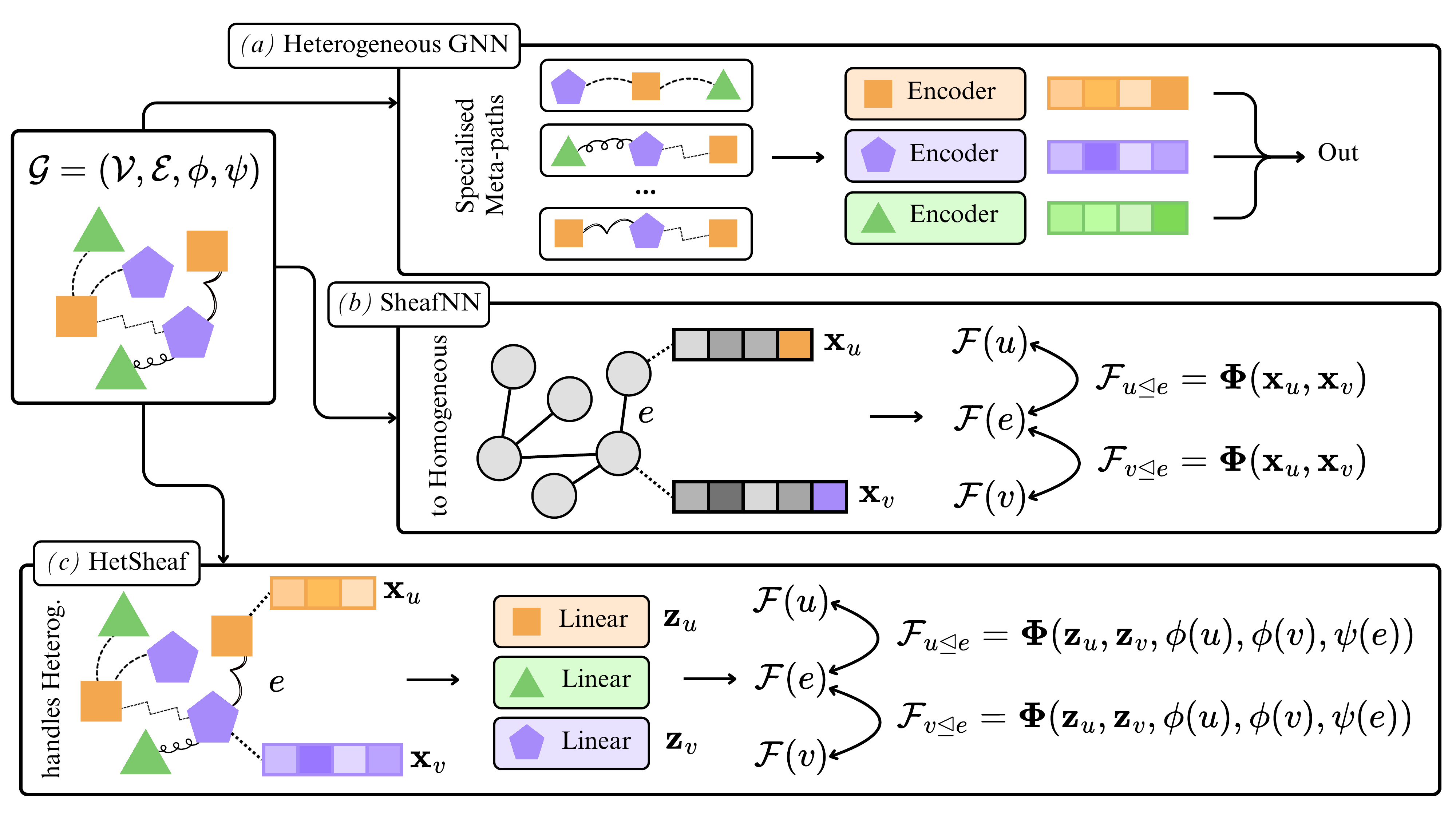}}
\end{figure}
\noindent To address these limitations, we propose \textsc{HetSheaf}, a framework specifically tailored for learning on heterogeneous graphs through \emph{cellular sheaves}. A \emph{sheaf} augments a graph by attaching a local vector space (or \emph{stalk}) to each node and edge, which acts as a dedicated feature space for that specific entity. Information is translated between these local spaces via linear transformations called \emph{restriction maps}. \textsc{HetSheaf}, first projects type-dependent input features into a shared channel dimension and then infers a cellular sheaf where restriction maps are conditioned on node features, node types, and edge types. Moreover, to execute graph-level inductive tasks on cellular sheaves a new challenge emerges in aggregating feature vectors from all nodes (i.e., graph pooling): in standard GNNs, all node embeddings lie in the same latent space, so graph classification can be performed by applying any permutation-invariant pooling operation (e.g., mean, sum, or max). Conversely, in sheaf networks, nodes and edges have their own stalk space and are only defined \emph{up to a change of basis}, i.e., two embeddings might represent the same underlying information but have different coordinate matrices. Consequently, flattening node embeddings and pooling them can yield different graph representations for mathematically equivalent sheaf states. Therefore, the graph pooling must be  \emph{gauge-invariant}, i.e.,  tolerate node-wise reparameterization of the stalk coordinates. To address this, we introduce \textsc{SheafPool}, the first stalk-space pooling for graph classification. \textsc{SheafPool} constructs a stalk graph representation by aggregating node  stalks independently of their arbitrary local coordinates. To achieve this, it resolves coordinate ambiguities and rotates the node embeddings into a common reference frame prior to the final aggregation step.


\noindent\textbf{Summary of main contributions}
\begin{enumerate}
    \item We propose \textsc{HetSheaf}, a model-agnostic framework for heterogeneous sheaf neural networks that preserves type-specific semantics in the data representation without specialised, parameter-heavy models (\Cref{fig:pipeline}). We instantiate it with Neural Sheaf Diffusion (NSD)~\citep{bodnarNeuralSheafDiffusion2022}, Sheaf Attention Networks (SheafAN)~\citep{barberoSheafAttentionNetworks2022}, and Polynomial Neural Sheaf Diffusion (PolyNSD)~\citep{borgi2026polynomialneuralsheafdiffusion}.
    \item We introduce a family of \emph{heterogeneous sheaf predictors} that infer restriction maps by explicitly conditioning on node features, node and edge types, allowing the learned sheaf structure to adapt to the typed nature of the underlying relational data (\Cref{fig:sheaf-predictors}).
    \item We introduce \textsc{SheafPool}, the first graph-level pooling for sheaf neural networks. Since stalk embeddings lie in distinct coordinate systems, standard pooling is not geometrically well defined. \textsc{SheafPool} resolves it through whitening, anchor-based alignment, invariant attention, and late-token processing, yielding an injective, architecture-agnostic pooling that improves Macro F1 by up to 42\% over mean pooling (\Cref{sec:universal_stalk_space_late_token}).
    \item We evaluate \textsc{HetSheaf} and \textsc{SheafPool} on node classification, link prediction, and graph classification benchmarks, following the HGB setup~\citep{lvAreWeReally2021a} (\Cref{sec:results}). \textsc{HetSheaf} outperforms type-agnostic sheaf models and specialised HGNNs by up to 2\%, while being up to $10\times$ more parameter efficient compared to R-GCN, HAN, and HGT~\citep{huHeterogeneousGraphTransformer2020, schlichtkrullModelingRelationalData2018, wangHeterogeneousGraphAttention2019a} (\Cref{app:computational-overhead}).
\end{enumerate}

\section{Background and Related Work}
\label{sec:related_work}
\noindent \textbf{Homogeneous and Heterogeneous Graph Neural Networks.}
Homogeneous GNNs~\citep{hamiltonGraphRepresentationLearning2020,kipf2016semi,liGatedGraphSequence2017,velickovicGraphAttentionNetworks2018,defferrard2016convolutional,corso2020pna} can be applied to heterogeneous graphs by discarding node and edge types, and consequently they do not model type-specific relational structure. Heterogeneous GNNs instead are also able to model the type of nodes and edges. Some of them use meta-paths, predefined type patterns~\citep{sunPathSimMetaPathbased2011,sunMiningHeterogeneousInformation2012,dongMetapath2vecScalableRepresentation2017}, or relation-specific processing. HAN~\citep{wangHeterogeneousGraphAttention2019a} and MAGNN~\citep{fuMAGNNMetapathAggregated2020} aggregate information through manually specified meta-paths; GTN~\citep{yunGraphTransformerNetworks2019} learns such paths, although they can be costly with many edge types. Relational approaches include RSHN~\citep{zhuRelationStructureAwareHeterogeneous2019}, which derives edge features on a coarsened line graph before MPNN propagation~\citep{gilmerNeuralMessagePassing2017}; R-GCN~\citep{schlichtkrullModelingRelationalData2018}, which applies GCNs to relation-induced subgraphs; HGT~\citep{huHeterogeneousGraphTransformer2020} and HetGNN~\citep{zhangHeterogeneousGraphNeural2019}, which respectively use sampled transformer attention and random-walk/RNN aggregation; and SlotGAT~\citep{zhouSlotGATSlotbasedMessage2023}, SimpleHGNN~\citep{lvAreWeReally2021a}, and Space4HGNN~\citep{zhaoSpace4HGNNNovelModularized2022}. Recent examples include HAGNN~\citep{zhu2025hagnn}, which fuses meta-path-based intra-type and meta-path-free inter-type aggregation, and HGNN-AR\textsuperscript{2}~\citep{lin2025adaptivehetero}, which adaptively reconstructs relations. Rather than adding specialised modules, we represent type-specific information through cellular sheaves.

\noindent\textbf{Sheaf Neural Networks.}
Cellular sheaves~\citep{shepard1985cellular,curry2014sheaves} attach local feature spaces (\emph{stalks}) and linear restriction maps to graphs, offering a geometric response to oversmoothing~\citep{caiNoteOverSmoothingGraph2020}, oversquashing~\citep{alon2021on}, and heterophily. SheafNNs progressed from fixed one-dimensional sheaves~\citep{hansen2020sheaf} to learned restriction maps~\citep{bodnarNeuralSheafDiffusion2022}, and now include attention~\citep{barberoSheafAttentionNetworks2022}, learned connection Laplacians~\citep{pmlr-v196-barbero22a}, nonlinear and joint diffusion~\citep{zaghen2024nonlinear,caralt2024joint}, polynomial filtering~\citep{borgi2026polynomialneuralsheafdiffusion}, directional models~\citep{fiorini2026sheaves}, and extensions to hypergraphs~\citep{dutaSheafHypergraphNetworks2023a,mule2026directional}, recommendation~\citep{purificatoSheaf4RecSheafNeural2025}, federated learning~\citep{nguyen2024federated}, cooperative dynamics~\citep{ribeiro2026cooperative}, and copresheaves~\citep{hajij2025copresheaf}. Most work concerns node-level tasks. For graph prediction, readout must be both permutation-invariant and invariant to node-wise stalk reparameterisations; standard pooling~\citep{grattarola2024poolings,huaHighorderPoolingGraph2022,liGatedGraphSequence2017,lee2019attPooling} only guarantees the former.

\noindent\textbf{Cellular Sheaves on Graphs.}
For an oriented graph $\gG=(\gV,\gE)$, a cellular sheaf $\gF$ assigns stalks $\gF(u)$ and $\gF(e)$ to nodes and edges, with restriction maps $\gF_{u \trianglelefteq e}:\gF(u)\to\gF(e)$ for every incidence. Its $0$- and $1$-cochains are $C^0(\gG,\gF)=\bigoplus_{u\in\gV}\gF(u)$ and $C^1(\gG,\gF)=\bigoplus_{e\in\gE}\gF(e)$. For $e=(u\to v)$, the coboundary $\delta(\rvx)_e=\gF_{v \trianglelefteq e}\rvx_v-\gF_{u \trianglelefteq e}\rvx_u$ measures disagreement in the shared edge space~\citep{hansen2019toward}; $L_{\gF}=\delta^\top\delta$ is the sheaf Laplacian. With $d$-dimensional stalks and block diagonal $D$ of $L_{\gF}$, its normalised action is $\Delta_{\gF}(\rvx)_u
    \coloneqq
    D^{-1/2}
    \left[
    \sum_{u,v \trianglelefteq e}
    \gF_{u \trianglelefteq e}^{\top}
    \bigl(
        \gF_{u \trianglelefteq e}\rvx_u
        -
        \gF_{v \trianglelefteq e}\rvx_v
    \bigr)
    \right]
    D^{-1/2}$. Thus $L_{\gF}\in\mathbb{R}^{dn\times dn}$ for $n=|\gV|$; scalar stalks with identity restrictions recover the graph Laplacian. For $f$ feature channels, $\rmX\in\mathbb{R}^{nd\times f}$, and NSD learns restrictions as $\gF_{u \trianglelefteq e\coloneqq(u,v)}=\boldsymbol{\Phi}(\rvx_u,\rvx_v)=\MLP(\rvx_u\|\rvx_v)$.

\section{Heterogeneous Sheaf Neural Networks}
\noindent\textbf{Heterogeneous Graph Setting.} We consider a heterogeneous graph~\citep{sunMiningHeterogeneousInformation2012} $\gG = (\gV,\gE,\phi,\psi)$, where $\gV$ and $\gE$ denote the sets of nodes and edges, while $\phi:\gV\to\Phi$ and $\psi:\gE\to\Psi$ are node and edge type mapping functions respectively. For an edge $e=(u,v)$, we interchangeably write its type as $\psi(e)$ or $\psi(u,v)$. 
Each node $u\in\gV$ is associated with a feature vector $\rvx_u$. Because feature sizes vary by node type, all nodes of a given type $t\in\Phi$ share a specific input dimension $f_t$, meaning $\rvx_u \in \mathbb{R}^{f_{\phi(u)}}$. 

\subsection{\textsc{HetSheaf}: A Sheaf-based framework for Heterogeneous data}
\label{sec:hetsheaf-framework}
Existing heterogeneous GNNs (\Cref{fig:pipeline}a) handle heterogeneity at the architectural level, often leading to increasingly specialised and parameter-heavy designs that scale with the number of entity types. Instead, we adopt a \emph{representation-centric view} by modifying the input structure to encode specific heterogeneous knowledge while preserving topological information. Unlike standard graphs, which assume a single shared feature space for all nodes, cellular sheaves naturally capture heterogeneous data through type-specific local vector spaces and restriction maps that preserve local structure while enabling coherent global message passing. Motivated by this perspective, we introduce \textsc{HetSheaf} (\Cref{fig:pipeline}c), a general framework for heterogeneous SheafNNs. Given a heterogeneous graph $\gG = (\gV,\gE,\phi,\psi)$, it first preprocesses type-dependent node features $\rvx$ into a common channel dimension $\rvz$, then infers a cellular sheaf whose restriction maps are conditioned on both features and type information, and finally applies a sheaf neural architecture for downstream prediction. More formally, a heterogeneous sheaf neural network consists of three components:
\begin{enumerate}
    \item \emph{Type-aware preprocessing}, which projects input features $\rvx$ into a shared channel space $\rvz$. Because each node type may have a different number of feature channels, we use a linear layer with bias for each node type to project node features into a common number of input channels. This can be combined with the sheaf projection step of the SheafNN to create the \emph{heterogeneous sheaf cochain} $\rvz \in C^0(\gG, \gF)$. The modified linear layer takes in the original feature channels $f_{t}$ and outputs $df$ features as the node stalk projection. We follow~\citep{lvAreWeReally2021a,zhouSlotGATSlotbasedMessage2023} to generate the input feature types $f$ by using either all of the node feature types (type-0), only the feature of the node target type (type-1) or replacing all node features with one-hot vectors (type-2). 
    \item \emph{Heterogeneous sheaf inference}, (\Cref{fig:sheaf-predictors}) which predicts restriction maps conditioned on node features and type information.
    \item \emph{Sheaf-based representation learning}, which applies a sheaf neural encoder followed by a task-specific decoder or readout. We apply task-specific representations (e.g., $\ell_2$ normalisation~\citep{lvAreWeReally2021a} or JKNet-style concatenations~\citep{xuRepresentationLearningGraphs2018}) before the final predictions, as detailed in \Cref{sec:hetero-sheaf-predictors}.
    
\end{enumerate}

\noindent Let $\mathcal P$ denote the preprocessing module, $\boldsymbol{\Phi}$ the sheaf predictor, $\mathcal M$ a sheaf neural encoder, and $\mathcal D$ a downstream decoder. The overall pipeline can be written abstractly as:
\begin{equation}
    \label{eq:hetsheaf_pipeline1}   
    \rvz_u = \mathcal P_{\phi(u)}(\rvx_u), \;\;\;\:
    \gF_{u \trianglelefteq e\coloneq(u,v)} = \boldsymbol{\Phi} (\rvz_u,\rvz_v,\phi(u),\phi(v),\psi(e)), \;\;\;\:
    \hat{\rvy} = \mathcal D\!\left(\mathcal M(\gG,\gF,\{\rvz_u\}_{u\in\gV})\right)
\end{equation}
Since heterogeneity is encoded into the inferred sheaf itself, once the sheaf has been constructed, it can be processed by any sheaf-based neural architecture, including NSD \cite{bodnarNeuralSheafDiffusion2022}, Sheaf Attention Networks \cite{barberoSheafAttentionNetworks2022}, Polynomial NSD \cite{borgi2026polynomialneuralsheafdiffusion} or other sheaf variants, without requiring ad hoc heterogeneous redesign.

\begin{figure}[!htbp]
    \centering
    \resizebox{1\columnwidth}{!}{%

\begin{tikzpicture}[x=1cm,y=1.48cm, trim left=-0.75cm, trim right=23.9cm]
\colorlet{ufill}{typeA!16}
\colorlet{uborder}{typeA!80!black}
\colorlet{vfill}{typeB!16}
\colorlet{vborder}{typeB!80!black}
\colorlet{eborder}{slate}

\newcommand{\sheafblock}[9]{
    \begin{scope}[shift={(#1,#2)}]
        \node[draw=slate!30, rounded corners=6pt, fill=white, 
              minimum width=6.0cm, minimum height=4.0cm, anchor=south west] (box) at (0,0) {};
        
        \node[font=\bfseries\fontsize{13}{15}\selectfont, text=slate, anchor=north] 
            at ([yshift=-6pt]box.north) {#3};
        
        
        \node[nodecircle=typeA, minimum size=36pt, font=\fontsize{13}{15}\selectfont]
            (u) at (1.55, 1.5) {$#4$}; 
        \node[nodecircle=typeB, minimum size=36pt, font=\fontsize{13}{15}\selectfont]
            (v) at (4.55, 1.5) {$#5$}; 
        
        \draw[eborder!40, line width=1.2pt] (u) -- (v) 
            node[font=\fontsize{13}{15}\selectfont, fill=white, inner sep=2pt, pos=0.5, yshift=13pt, text=slate] {$e$}
            node[font=\fontsize{13}{15}\selectfont, fill=white, inner sep=2pt, pos=0.5, yshift=-13pt, text=eborder] {$#8$};
        
        \node[left=5pt of u, font=\fontsize{13}{15}\selectfont, text=uborder] {$u$};
        \node[right=5pt of v, font=\fontsize{13}{15}\selectfont, text=vborder] {$v$};
        \node[font=\fontsize{13}{15}\selectfont, text=uborder, above=0.5pt of u, xshift=-25pt] {$#6$};
        \node[font=\fontsize{13}{15}\selectfont, text=vborder, above=0.5pt of v, xshift=25pt] {$#7$};
        
        \node[font=\fontsize{13}{15}\selectfont, text=slate, anchor=south] 
            at ([yshift=8pt]box.south) {$#9$};
    \end{scope}
}

\def\colA{-0.65}
\def\colB{5.50}
\def\colC{11.65}
\def\colD{17.80}
\def\rowOne{0}
\def\rowTwo{-2.8}


\sheafblock{\colA}{\rowOne}{Sheaf-NSD}{\mathbf{x}_u}{\mathbf{x}_v}{}{}{}{\mathcal{F}_{u\unlhd e} = \text{MLP}(\mathbf{z}_u \Vert \mathbf{z}_v)}
\sheafblock{\colB}{\rowOne}{HetSheaf-ensemble}{\mathbf{x}_u}{\mathbf{x}_v}{}{}{\psi(e)}{\mathcal{F}_{u\unlhd e} = \text{MLP}_{\psi(e)}(\mathbf{z}_u \Vert \mathbf{z}_v)}
\sheafblock{\colC}{\rowOne}{HetSheaf-NE}{\mathbf{x}_u}{\mathbf{x}_v}{\phi(u)}{\phi(v)}{}{\mathcal{F}_{u\unlhd e} = \text{MLP}(\mathbf{z}_u \Vert \dots \Vert \phi(v))}
\sheafblock{\colD}{\rowOne}{HetSheaf-EE}{\mathbf{x}_u}{\mathbf{x}_v}{}{}{\psi(e)}{\mathcal{F}_{u\unlhd e} = \text{MLP}(\mathbf{z}_u \Vert \mathbf{z}_v \Vert \psi(e))}

\sheafblock{\colA}{\rowTwo}{HetSheaf-TE}{\mathbf{x}_u}{\mathbf{x}_v}{\phi(u)}{\phi(v)}{\psi(e)}{\mathcal{F}_{u\unlhd e} = \text{MLP}(\dots \Vert \psi(e))}
\sheafblock{\colB}{\rowTwo}{HetSheaf-NT}{\mathbf{x}_u}{\mathbf{x}_v}{\phi(u)}{\phi(v)}{}{\mathcal{F}_{u\unlhd e} = \text{MLP}(\phi(u) \Vert \phi(v))}
\sheafblock{\colC}{\rowTwo}{HetSheaf-ET}{\mathbf{x}_u}{\mathbf{x}_v}{}{}{\psi(e)}{\mathcal{F}_{u\unlhd e} = \text{MLP}(\psi(e))}
\sheafblock{\colD}{\rowTwo}{HetSheaf-types}{\mathbf{x}_u}{\mathbf{x}_v}{\phi(u)}{\phi(v)}{\psi(e)}{\mathcal{F}_{u\unlhd e} = \text{MLP}(\phi(u) \Vert \phi(v) \Vert \psi(e))}
\end{tikzpicture}%
    }
    \caption{
        \textbf{Heterogeneous Sheaf Predictors.}
        For an edge $e=(u,v)$, $\phi(u)$ and $\phi(v)$ denote node types, $\psi(e)$ denotes the edge type, $x_u,x_v$ are the raw node features, projected in the shared channel via $\rvz_* = \mathcal P_{\phi(*)}(\rvx_*)$. $\|$ denotes concatenation, and $\mathrm{MLP}_{\psi(e)}$ denotes a separate MLP for each edge type.
    }
    \label{fig:sheaf-predictors}
\end{figure}
\vspace{-10pt}
\subsection{Heterogeneous Sheaf Predictors and Model Adaptations}
\label{sec:hetero-sheaf-predictors}
\noindent\textbf{Heterogeneous Sheaf Predictors.} 
A key component of \textsc{HetSheaf} is the construction of restriction maps that explicitly incorporate heterogeneous information. Unlike standard sheaf learning, where maps depend only on node features, \textsc{HetSheaf} conditions them on projected node features (into the shared channel via $\mathcal P_{\phi(*)}$), node types, and edge types, as defined in Eq.~\ref{eq:hetsheaf_pipeline1}. Following \cite{bodnarNeuralSheafDiffusion2022}, we consider diagonal, orthogonal, and general restriction-map families (see more details in \Cref{app:restriction_maps_impl}), and we instantiate the heterogeneous sheaf through a \emph{family of heterogeneous sheaf predictors} that vary along two axes. The former concerns \emph{which type signals are used}: from no type information (Sheaf-NSD, where we only concatenate the projected information relative to the node features $\rvz_u, \rvz_v$), to node-only (\textsc{HetSheaf-NE, HetSheaf-NT}) or edge-only (\textsc{HetSheaf-EE, HetSheaf-ET}) variants, up to predictors conditioned on both node and edge types (\textsc{HetSheaf-TE, HetSheaf-types, HetSheaf-ensemble}). The latter regards \emph{how type information is routed}: most variants use a shared MLP with type indicators as additional inputs, whereas \textsc{HetSheaf-ensemble} allocates a separate predictor to each edge type $\text{MLP}_{\psi(e)}$, increasing expressivity at the cost of more parameters. \Cref{fig:sheaf-predictors} summarises the predictor family, while full details are deferred to \Cref{app:hetero-sheaf-predictors}.

\noindent\textbf{Attention Restriction Maps.} A further design choice concerns how the raw output of $\boldsymbol{\Phi}$ is converted into a restriction map. \citet{bodnarNeuralSheafDiffusion2022} formulate an alternative attention-style post-processing step. For the general case, a row-wise softmax yields row-stochastic restriction maps, and this transformation depends only on the output of $\boldsymbol{\Phi}$, and we can apply it uniformly to all \textsc{HetSheaf} variants:
\begin{equation}
    \gF_{u \trianglelefteq e}
    =
    \rmI_d - \mathrm{Softmax}\!\left(\boldsymbol{\Phi}(\rvz_u,\rvz_v,\phi(u),\phi(v),\psi(e))\right)
    \label{eq:sheafan-general}
\end{equation}

\noindent\textbf{Task-specific Preprocessing.}
Because different node types may have different input dimensionalities, we use a type-specific linear projection with bias $\rvz_u = \mathcal P_{\phi(u)}(\rvx_u), \forall u \in \gV$ to map all node features into a shared channel dimension. These projections are \emph{learned jointly} with the rest of the model and can be combined with the initial stalk projection of the sheaf encoder to construct the input cochain $\rvz \in C^0(\gG,\gF)$. Following \citet{lvAreWeReally2021a,zhouSlotGATSlotbasedMessage2023}, we consider multiple feature initialisation strategies, including using all available node features, target-type-only features, and one-hot type indicators. For \emph{node classification}, we apply $\ell_2$ normalisation to the final node representation: $\rvo_u=    \frac{\rvh_u^{(L)}}{\norm{\rvh_u^{(L)}}}$, where $\rvh_u^{(L)}$ is the final hidden state of node $u$. For \emph{link prediction}, we follow a JKNet-style strategy~\citep{xuRepresentationLearningGraphs2018} and concatenate the normalised hidden states across layers: $\rvo_u=\concat_{l=1}^{L}\frac{\rvh_u^{(l)}}{\norm{\rvh_u^{(l)}}}$. 

\noindent\textbf{Sheaf Models Adaptations.}
A key advantage of \textsc{HetSheaf} is its model-agnosticity: once a heterogeneous sheaf has been inferred, the resulting typed geometric structure can be processed by a broad class of existing sheaf neural architectures with minimal modifications. To demonstrate this property, we adapt both SheafAN~\citep{barberoSheafAttentionNetworks2022} and PolyNSD~\citep{borgi2026polynomialneuralsheafdiffusion} to the heterogeneous setting. For SheafAN, we make the attention mechanism type-aware by augmenting the attention input with node-type and edge-type information. For PolyNSD, instead, we preserve the original polynomial spectral filtering scheme and apply it directly to the sheaf Laplacian, allowing to perform higher-order diffusion directly over the heterogeneous sheaf without requiring any ad hoc redesign. Full model-specific derivations and implementation details are deferred to \Cref{app:sheafan,app:polynsd}.

\section{\textsc{SheafPool}: Stalk-Space Late-token Readout}
\label{sec:universal_stalk_space_late_token}
Graph-level prediction is more delicate in SheafNNs than in GNNs. While relying on the common latent space assumption, for which we can apply a pooling hierarchy, sheaf node embeddings $\rmH_i \in \mathbb{R}^{d \times C}$ are meaningful only up to node-wise changes of basis. 

\begin{table}[!htbp]
    \centering
    \caption{
\emph{Pooling ablation on graph classification}, same \textsc{HetSheaf} backbone. Macro/Micro F1, mean $\pm$ std over five seeds; best per dataset in \textbf{bold}. Attention is the self-attention pooling of \citet{lee2019attPooling}.}
    \label{tab:pooling-ablation}
    \resizebox{0.9\textwidth}{!}{%
                \maxsizebox{0.999\textwidth}{!}{
	\begin{tabular}{lcccccccc}
		\toprule
		{} & \multicolumn{2}{c}{\thead{MUTAG}}
		   & \multicolumn{2}{c}{\thead{PROTEINS}}
		   & \multicolumn{2}{c}{\thead{ENZYMES}}
		   & \multicolumn{2}{c}{\thead{NCI1}} \\
		\cmidrule(lr){2-3}
		\cmidrule(lr){4-5}
		\cmidrule(lr){6-7}
		\cmidrule(lr){8-9}
		{\thead{Pooling}} & {\thead{Macro F1}} & {\thead{Micro F1}}
		                  & {\thead{Macro F1}} & {\thead{Micro F1}}
		                  & {\thead{Macro F1}} & {\thead{Micro F1}}
		                  & {\thead{Macro F1}} & {\thead{Micro F1}} \\
		\midrule
		Mean                 & 94.88\pmstd{1.11} & 94.92\pmstd{2.26} & 57.51\pmstd{1.76} & 67.04\pmstd{2.19} & 19.92\pmstd{1.77} & 23.28\pmstd{1.29} & 57.88\pmstd{1.78} & 57.97\pmstd{1.29} \\
		Max                  & 82.55\pmstd{1.16} & 82.41\pmstd{1.66} & 67.84\pmstd{1.82} & 73.15\pmstd{2.31} & 13.59\pmstd{1.83} & 15.06\pmstd{2.22} & 64.11\pmstd{1.82} & 64.02\pmstd{1.23} \\
		Add                  & 74.24\pmstd{1.40} & 74.32\pmstd{1.26} & 65.68\pmstd{1.75} & 67.02\pmstd{2.22} & 10.03\pmstd{1.75} & 16.61\pmstd{1.30} & 57.45\pmstd{1.75} & 57.54\pmstd{1.28} \\
		Attention~\cite{lee2019attPooling}           & 82.77\pmstd{2.31} & 84.94\pmstd{2.26} & 69.12\pmstd{1.84} & 76.85\pmstd{2.20} & 13.74\pmstd{1.81} & 16.72\pmstd{2.24} & 65.71\pmstd{1.84} & 65.61\pmstd{2.21} \\
		\midrule
		\textsc{SheafPool} (ours)    & \bfseries 98.69\pmstd{1.31} & \bfseries 98.68\pmstd{1.32} & \bfseries 80.55\pmstd{1.55} & \bfseries 82.14\pmstd{1.42} & \bfseries 62.36\pmstd{2.14} & \bfseries 63.30\pmstd{1.94} & \bfseries 79.42\pmstd{0.87} & \bfseries 79.42\pmstd{0.83} \\
		\bottomrule
	\end{tabular}
    }

    }
\end{table}

\noindent Consequently, a valid graph readout must satisfy two requirements simultaneously: it must be permutation-invariant over nodes, and it must be compatible with stalk-wise gauge reparameterisations $\rmH_i \mapsto \rmG_i \rmH_i, \rmG_i \in \mathbb{G}$, where $\mathbb{G}$ denotes the gauge family induced by $\gF$ (e.g.\ diagonal, orthogonal, or general maps). However, the standard flatten-and-pool procedure on $\mathrm{vec}(\rmH_i)\in\mathbb{R}^{dC}$ does not satisfy this requirement: they depend on arbitrary stalk coordinates chosen for each stalk, so equivalent sheaf representations may produce different graph embeddings, (as also shown in \Cref{tab:pooling-ablation}). 
To address this issue, we introduce \textsc{SheafPool}, a \emph{Stalk-Space Late-token Readout} graph pooling mechanism that first canonicalises node embeddings (i.e., maps to a common reference stalk frame), then aggregates them directly in stalk space, collapsing the stalk dimension at the very end. The construction preserves the stalk geometry, while remaining applicable to all sheaf families.

\begin{figure}[!htbp]
    \centering
    \resizebox{\textwidth}{!}{
    \input{figures/figure2}
    }
    \caption{\textbf{The \textsc{SheafPool} Architecture.} Our readout processes node stalk embeddings in five steps. \emph{(1)} \emph{Per-node Whitening} removes GL(d) scaling and shearing. \emph{(2)} \emph{Anchor Alignment} resolves residual O(d) rotations. \emph{(3)} \emph{Attention Weights} are computed via stalk energies. \emph{(4)} \emph{Stalk Pooling} aggregates into a \emph{Late Receive-only Token} without altering message passing. \emph{(5)} \emph{Feature Extraction} collapses the stalk dimension, yielding a permutation- and gauge-invariant graph embedding.}
    \label{fig:sheafpool_architecture}
\end{figure}

\noindent\textbf{Stalk-Space Late-Token Formalisation.}
Formally, we define a graph-level readout as $\mathcal{R}:\{\rmH_i\}_{i\in V_g} \longmapsto \rvg_g \in \mathbb{R}^C$, whose output is permutation-invariant over nodes and robust to stalk reparameterisations. The readout consists in five stages that can be summarised (\cref{fig:sheafpool_architecture}) as:
\begin{equation}
    \rmH_i
    \xrightarrow{\text{whiten}}
    \rmZ_i
    \xrightarrow{\text{align}}
    \widetilde{\rmZ}_i
    \xrightarrow{\text{attn pool}}
    \rmP_g
    \xrightarrow{\text{late token}}
    \widehat{\rmT}_g
    \xrightarrow{\text{energy}}
    \rvg_g
\end{equation}

\noindent\emph{(1) Per-node whitening.}
We first remove node-wise scaling and shearing in the stalk dimension by forming $\rmS_i = \rmH_i \rmH_i^\top + \varepsilon \rmI_d$, and defining $\rmZ_i = \rmS_i^{-1/2}\rmH_i$. This makes the stalk covariance approximately isotropic, thus reducing remaining ambiguity to an orthogonal transform.

\noindent\emph{(2) Anchor-based alignment.}
After whitening, different nodes may still be represented in different orthogonal frames. We therefore align each node to a shared learnable anchor $\rmA \in \mathbb{R}^{d\times C}$ via $\rmR_i 
    =
    \arg\min_{\rmR\in O(d)}
    \|\rmR^\top \rmZ_i - \rmA\|_F^2$, and set $\widetilde{\rmZ}_i = \rmR_i^\top \rmZ_i$. This places all node embeddings in a common stalk frame before pooling.

\noindent\emph{(3) Invariant attention weights.}
To weight node contributions without using basis-dependent coordinates, we score each node through the channel-wise energies of its whitened representation $\rvu_i = \mathrm{diag}(\rmZ_i^\top \rmZ_i) \in \mathbb{R}^C$. Since $\rvu_i$ is invariant to residual orthogonal reparameterisations, a learnable scorer $s_\theta$ can produce graph-wise attention weights $\alpha_i
    =
    \frac{\exp(s_\theta(\rvu_i))}
    {\sum_{j\in V_g}\exp(s_\theta(\rvu_j))}.
$

\noindent\emph{(4) Stalk-space pooling with a late receive-only token.}
We then aggregate the aligned embeddings directly in stalk space $\rmP_g = \sum_{i\in V_g} \alpha_i \widetilde{\rmZ}_i \in \mathbb{R}^{d\times C}$. To perform readout without inserting a token into the encoder during the diffusion process, which would cause oversmoothing, we use a \emph{late, receive-only token} $\rmT \in \mathbb{R}^{d\times C}$ updated \emph{only} after pooling $\widehat{\rmT}_g = \phi([\rmT,\rmP_g])$, where $\phi$ is a small channel-wise MLP with layer normalisation. 

\noindent\emph{(5) Invariant graph feature extraction.}
Finally, we collapse the stalk dimension through channel-wise energies: $\rvg_g=\mathrm{diag}(\widehat{\rmT}_g^\top \widehat{\rmT}_g)\in\mathbb{R}^C$. The resulting graph embedding is then fed to a standard MLP decoder for graph classification. 

\noindent Full derivations, invariance arguments, and stability remarks are deferred to \Cref{app:universal_stalk_space_late_token}. 

\section{Experimental Evaluation}
\label{sec:results}
We evaluate \textsc{HetSheaf} on heterogeneous node-, edge-, and graph-level tasks, complemented by ablations on sheaf predictors, restriction-map families, attention mechanisms, \textsc{SheafPool}, and different sheaf architectures. These experiments test four claims: (i) encoding heterogeneity directly in the inferred sheaf improves performance; (ii) conditioning restriction maps on typed information outperforms type-agnostic sheaf learning (\Cref{app:external_comparisons}); (iii) \textsc{HetSheaf} is model-agnostic, i.e., can be instantiated across multiple sheaf variants~\citep{bodnarNeuralSheafDiffusion2022, barberoSheafAttentionNetworks2022, borgi2026polynomialneuralsheafdiffusion}; and (iv) these gains are achieved with favourable parameter efficiency while preserving heterogeneous structure through \textsc{SheafPool}.
\begin{table*}[!htbp]
    \centering
    \caption{\emph{Heterogeneous node classification.} Macro/Micro F1, mean $\pm$ std over five seeds. Top three by mean: \textbf{\textcolor{Top1}{first}}, \textbf{\textcolor{Top2}{second}}, \textbf{\textcolor{Top3}{third}}.}
    \label{tab:nc_results}
    \resizebox{0.8\textwidth}{!}{%
        \begin{tabular}{l cc cc cc}
            \toprule
            {} & \multicolumn{2}{c}{\thead{ACM}} & \multicolumn{2}{c}{\thead{DBLP}} & \multicolumn{2}{c}{\thead{IMDB}} \\
            \cmidrule(lr){2-3}\cmidrule(lr){4-5}\cmidrule(lr){6-7}
            {} & {\thead{Macro F1}} & {\thead{Micro F1}}
               & {\thead{Macro F1}} & {\thead{Micro F1}}
               & {\thead{Macro F1}} & {\thead{Micro F1}} \\
            \midrule
            \multicolumn{7}{l}{\textit{Homogeneous}} \\
            GCN
                & 92.25\pmstd{0.82} & 91.25\pmstd{0.84} & 90.87\pmstd{1.77} & 91.68\pmstd{1.83} & 58.26\pmstd{0.36} & 60.89\pmstd{0.26} \\
            GAT
                & 92.16\pmstd{1.55} & 91.95\pmstd{0.87} & 92.33\pmstd{1.13} & 92.81\pmstd{0.99} & 60.23\pmstd{0.67} & \textbf{\textcolor{Top3}{64.06\pmstd{0.80}}} \\
            GIN
                & 88.70\pmstd{2.20} & 88.78\pmstd{1.98} & \textbf{\textcolor{Top2}{94.83\pmstd{0.43}}} & \textbf{\textcolor{Top2}{95.24\pmstd{0.43}}} & 60.30\pmstd{0.75} & 63.96\pmstd{0.52} \\
            SAGE
                & 92.18\pmstd{1.61} & 92.07\pmstd{1.51} & 90.23\pmstd{1.31} & 90.69\pmstd{1.26} & \textbf{\textcolor{Top3}{60.60\pmstd{1.75}}} & 63.73\pmstd{0.63} \\
            \midrule
            \multicolumn{7}{l}{\textit{Heterogeneous}} \\
            R-GCN
                & \textbf{\textcolor{Top3}{93.27\pmstd{0.82}}} & \textbf{\textcolor{Top3}{93.19\pmstd{0.86}}} & \textbf{\textcolor{Top3}{94.30\pmstd{0.23}}} & \textbf{\textcolor{Top3}{94.70\pmstd{0.20}}} & 59.67\pmstd{0.21} & 62.36\pmstd{0.20} \\
            HAN
                & \textbf{\textcolor{Top2}{93.67\pmstd{0.26}}} & \textbf{\textcolor{Top2}{93.61\pmstd{0.28}}} & 94.07\pmstd{0.41} & 94.47\pmstd{0.38} & 56.85\pmstd{2.91} & 61.68\pmstd{1.08} \\
            HGT
                & 92.38\pmstd{0.51} & 92.28\pmstd{0.55} & 92.82\pmstd{1.14} & 93.37\pmstd{1.08} & 56.59\pmstd{0.89} & 60.25\pmstd{0.41} \\
            \midrule
            Sheaf-NSD
                & 92.07\pmstd{2.09} & 91.90\pmstd{2.09} & 92.91\pmstd{1.21} & 93.15\pmstd{1.18} & \textbf{\textcolor{Top2}{61.94\pmstd{0.34}}} & \textbf{\textcolor{Top2}{65.10\pmstd{0.20}}} \\
            \textbf{HetSheaf (ours)}
                & \textbf{\textcolor{Top1}{94.30\pmstd{0.24}}} & \textbf{\textcolor{Top1}{94.22\pmstd{0.24}}} & \textbf{\textcolor{Top1}{94.93\pmstd{0.36}}} & \textbf{\textcolor{Top1}{95.27\pmstd{0.34}}} & \textbf{\textcolor{Top1}{64.51\pmstd{0.68}}} & \textbf{\textcolor{Top1}{66.57\pmstd{0.17}}} \\
            \bottomrule
        \end{tabular}%
    }%
\end{table*}

\noindent\textbf{Datasets, Baselines and Evaluation Protocol.}
We evaluate \textsc{HetSheaf} on a broad suite of heterogeneous benchmarks spanning \emph{node classification}, \emph{link prediction}, and \emph{graph classification}, including citation, knowledge, user--item, biological, molecular, biomedical, and networking graphs~\citep{lvAreWeReally2021a,zhouSlotGATSlotbasedMessage2023,wangHeterogeneousGraphAttention2019a,huHeterogeneousGraphTransformer2020,yangHeterogeneousNetworkRepresentation2022,morris2020tudataset,hu2025heterobiomedical,BioERP2021,mcauleyHiddenFactors2013,leskovec2005graphs,ferriol2022routenet-erlang,ferriol2023routenet-fermi}. We compare against representative homogeneous (GCN, GAT, GIN, GraphSAGE), heterogeneous (R-GCN, HAN, HGT), and sheaf-based baselines, with additional comparisons to recent heterogeneous models in \Cref{app:external_comparisons}. We follow the standard evaluation protocol of each benchmark, and report \emph{Macro/Micro-F1} for classification and \emph{AUPR/AUROC} for link prediction. Hyperparameters are selected through 150-run sweeps per model type, after which the best validation configuration is re-trained over five random seeds (42--46); results are reported as mean $\pm$ standard deviation. Dataset details, preprocessing, hyperparameter spaces, hardware, and additional comparisons are provided in \Cref{app:exp_protocol_datasets,app:external_comparisons,app:implementation_details}.

\noindent\textbf{Experimental Results.}
\Cref{tab:nc_results,tab:gc_results,tab:lp_results} report the main results across tasks, where \textsc{HetSheaf} emerges as the most consistently strong method. It achieves the best performance on all three node-classification benchmarks, ranks first on the two largest link-prediction datasets, and remains highly competitive on the remaining relation-prediction tasks. On graph classification, where the challenge is no longer only to learn typed local interactions but also to compress them into a stable graph-level representation, \textsc{HetSheaf} again performs strongly. \Cref{app:external_comparisons} situates \textsc{HetSheaf} within a broader landscape of heterogeneous GNNs from the \emph{recent literature}, including a matched-protocol re-run of HAGNN, SeHGNN, and SlotGAT under the same splits and metric code (\Cref{tab:matched_protocol}), while \Cref{app:task-specific-datasets} and \ref{app:networking-datasets} extend the study to biomedical networks and networking tasks, where \textsc{HetSheaf} and Sheaf-NSD consistently occupy the top ranks. \Cref{app:model_analysis} then dissects the source of these gains through a detailed analysis of heterogeneous sheaf predictors and restriction-map families, including extensions of \textsc{HetSheaf} to additional sheaf architectures such as \textsc{PolyNSD} and \textsc{SheafAN}. Further sensitivity analyses over the stalk dimension and the number of layers are reported in \Cref{app:stalk_ablation,app:layer_ablation}. 

\begin{table*}[!htbp]
    \centering
    \caption{\emph{Heterogeneous link prediction.} Binary AUPR/AUROC; mean $\pm$ std over five seeds. Top three by mean: \textbf{\textcolor{Top1}{first}}, \textbf{\textcolor{Top2}{second}}, \textbf{\textcolor{Top3}{third}}.}
    \label{tab:lp_results}
    \setlength{\tabcolsep}{3pt}
    \resizebox{0.95\textwidth}{!}{%
        \begin{tabular}{lcccccccccc}
            \toprule
            {} & \multicolumn{2}{c}{\thead{LastFM}} & \multicolumn{2}{c}{\thead{MovieLens}} & \multicolumn{2}{c}{\thead{PubMed}} & \multicolumn{2}{c}{\thead{Amazon}} & \multicolumn{2}{c}{\thead{YouTube}} \\
            \cmidrule(lr){2-3}\cmidrule(lr){4-5}\cmidrule(lr){6-7}\cmidrule(lr){8-9}\cmidrule(lr){10-11}
            {} & {\thead{AUPR}} & {\thead{AUROC}} & {\thead{AUPR}} & {\thead{AUROC}} & {\thead{AUPR}} & {\thead{AUROC}} & {\thead{AUPR}} & {\thead{AUROC}} & {\thead{AUPR}} & {\thead{AUROC}} \\
            \midrule
            \multicolumn{11}{l}{\textit{Homogeneous}} \\
            GCN                      & 97.05\pmstd{0.25} & 96.95\pmstd{0.17} & \textbf{\textcolor{Top3}{99.67\pmstd{0.04}}} & \textbf{\textcolor{Top2}{99.62\pmstd{0.04}}} & 96.03\pmstd{0.11} & 92.88\pmstd{0.28} & 92.44\pmstd{0.28} & 89.61\pmstd{0.27} & 89.44\pmstd{0.18} & 83.25\pmstd{0.17} \\
            GAT                      & 63.03\pmstd{0.22} & 51.11\pmstd{0.50} & 97.50\pmstd{0.22} & 97.83\pmstd{0.18} & 96.06\pmstd{0.14} & 93.23\pmstd{0.27} & 88.59\pmstd{0.19} & 86.11\pmstd{0.25} & 66.79\pmstd{0.39} & 58.31\pmstd{0.28} \\
            GIN                      & \textbf{\textcolor{Top2}{97.98\pmstd{0.19}}} & \textbf{\textcolor{Top3}{97.73\pmstd{0.30}}} & 99.58\pmstd{0.14} & 99.54\pmstd{0.04} & 96.11\pmstd{0.07} & 92.84\pmstd{0.12} & 92.04\pmstd{0.28} & 89.12\pmstd{0.29} & 88.62\pmstd{0.58} & 82.92\pmstd{0.49} \\
            SAGE                     & 95.07\pmstd{1.76} & 92.70\pmstd{2.89} & 99.21\pmstd{0.09} & 99.12\pmstd{0.08} & 95.53\pmstd{0.52} & 88.66\pmstd{0.97} & 90.82\pmstd{0.39} & 87.82\pmstd{0.32} & 84.52\pmstd{0.09} & 80.22\pmstd{0.32} \\
            \midrule
            \multicolumn{11}{l}{\textit{Heterogeneous}} \\
            R-GCN                    & \textbf{\textcolor{Top3}{97.84\pmstd{0.27}}} & \textbf{\textcolor{Top2}{97.88\pmstd{0.14}}} & 99.37\pmstd{0.10} & 99.34\pmstd{0.09} & 95.02\pmstd{0.09} & 93.09\pmstd{0.09} & 93.28\pmstd{0.11} & 91.48\pmstd{0.14} & \textbf{\textcolor{Top3}{91.58\pmstd{0.21}}} & \textbf{\textcolor{Top3}{88.38\pmstd{0.14}}} \\
            HAN                      & 88.24\pmstd{2.41} & 83.34\pmstd{2.12} & 63.58\pmstd{0.06} & 52.11\pmstd{0.13} & 95.79\pmstd{0.37} & 93.64\pmstd{0.60} & \textbf{\textcolor{Top2}{95.19\pmstd{0.68}}} & \textbf{\textcolor{Top2}{93.59\pmstd{0.87}}} & \textbf{\textcolor{Top2}{93.48\pmstd{0.26}}} & \textbf{\textcolor{Top2}{90.14\pmstd{0.36}}} \\
            HGT                      & 86.01\pmstd{0.80} & 74.55\pmstd{0.78} & 77.61\pmstd{1.46} & 73.90\pmstd{1.50} & \textbf{\textcolor{Top1}{97.56\pmstd{0.30}}} & \textbf{\textcolor{Top1}{96.43\pmstd{0.52}}} & \textbf{\textcolor{Top1}{96.13\pmstd{0.26}}} & \textbf{\textcolor{Top1}{95.65\pmstd{0.16}}} & \textbf{\textcolor{Top1}{94.96\pmstd{0.58}}} & \textbf{\textcolor{Top1}{92.55\pmstd{0.45}}} \\
            \midrule
            Sheaf-NSD                & 97.16\pmstd{0.16} & 96.58\pmstd{0.17} & \textbf{\textcolor{Top2}{99.69\pmstd{0.04}}} & \textbf{\textcolor{Top3}{99.58\pmstd{0.04}}} & \textbf{\textcolor{Top2}{96.41\pmstd{0.07}}} & \textbf{\textcolor{Top3}{93.69\pmstd{0.16}}} & 92.58\pmstd{0.24} & 90.02\pmstd{0.25} & 89.36\pmstd{0.40} & 83.22\pmstd{0.30} \\
            \textbf{HetSheaf (ours)} & \textbf{\textcolor{Top1}{98.51\pmstd{0.34}}} & \textbf{\textcolor{Top1}{98.33\pmstd{0.35}}} & \textbf{\textcolor{Top1}{99.71\pmstd{0.04}}} & \textbf{\textcolor{Top1}{99.64\pmstd{0.04}}} & \textbf{\textcolor{Top3}{96.35\pmstd{0.59}}} & \textbf{\textcolor{Top2}{93.82\pmstd{0.68}}} & \textbf{\textcolor{Top3}{93.72\pmstd{0.32}}} & \textbf{\textcolor{Top3}{91.71\pmstd{0.37}}} & 91.25\pmstd{0.59} & 87.32\pmstd{0.48} \\
            \bottomrule
        \end{tabular}%
    }%
\end{table*}
\begin{table*}[!htbp]
    \centering
    \caption{\emph{Heterogeneous graph classification.} Macro/Micro F1; mean $\pm$ std over five seeds. Top three by mean: \textbf{\textcolor{Top1}{first}}, \textbf{\textcolor{Top2}{second}}, \textbf{\textcolor{Top3}{third}}.}
    \label{tab:gc_results}
    \setlength{\tabcolsep}{3pt}
    \resizebox{0.85\textwidth}{!}{%
        \begin{tabular}{l cc cc cc cc}
            \toprule
            {} & \multicolumn{2}{c}{\thead{MUTAG}} & \multicolumn{2}{c}{\thead{PROTEINS}} & \multicolumn{2}{c}{\thead{ENZYMES}} & \multicolumn{2}{c}{\thead{NCI1}} \\
            \cmidrule(lr){2-3}\cmidrule(lr){4-5}\cmidrule(lr){6-7}\cmidrule(lr){8-9}
            {} & {\thead{Macro F1}} & {\thead{Micro F1}}
               & {\thead{Macro F1}} & {\thead{Micro F1}}
               & {\thead{Macro F1}} & {\thead{Micro F1}}
               & {\thead{Macro F1}} & {\thead{Micro F1}} \\
            \midrule
            \multicolumn{9}{l}{\textit{Homogeneous}} \\
            GCN
                & 89.56\pmstd{1.04} & 90.00\pmstd{1.35} & 76.82\pmstd{2.45} & 80.34\pmstd{2.47} & 12.71\pmstd{1.92} & 28.30\pmstd{2.71} & 64.13\pmstd{2.12} & 64.94\pmstd{2.31} \\
            GAT
                & 89.56\pmstd{1.02} & 90.00\pmstd{1.32} & 75.76\pmstd{1.51} & 78.54\pmstd{1.75} & 14.12\pmstd{1.29} & 16.60\pmstd{2.27} & 64.19\pmstd{2.44} & 64.90\pmstd{2.07} \\
            GIN
                & \textbf{\textcolor{Top3}{94.88\pmstd{0.72}}} & \textbf{\textcolor{Top3}{95.00\pmstd{0.78}}} & 70.64\pmstd{1.38} & 72.32\pmstd{1.32} & 20.10\pmstd{2.46} & 28.33\pmstd{2.43} & 64.17\pmstd{2.18} & 64.91\pmstd{2.86} \\
            SAGE
                & 89.56\pmstd{1.04} & 90.00\pmstd{1.35} & 74.59\pmstd{1.56} & 78.55\pmstd{1.35} & 8.55\pmstd{1.80} & 18.33\pmstd{2.26} & 63.64\pmstd{0.85} & 65.41\pmstd{1.09} \\
            \midrule
            \multicolumn{9}{l}{\textit{Heterogeneous}} \\
            R-GCN
                & 84.97\pmstd{1.86} & 85.00\pmstd{1.45} & 75.71\pmstd{2.73} & 78.54\pmstd{0.91} & 10.35\pmstd{2.07} & 16.67\pmstd{2.43} & 64.83\pmstd{2.11} & 65.60\pmstd{3.62} \\
            HAN
                & 82.90\pmstd{4.25} & 85.00\pmstd{1.16} & \textbf{\textcolor{Top2}{79.01\pmstd{1.94}}} & \textbf{\textcolor{Top1}{83.02\pmstd{1.04}}} & \textbf{\textcolor{Top3}{46.13\pmstd{2.28}}} & \textbf{\textcolor{Top3}{44.97\pmstd{2.59}}} & \textbf{\textcolor{Top3}{77.94\pmstd{1.59}}} & \textbf{\textcolor{Top3}{78.15\pmstd{1.56}}} \\
            HGT
                & 84.64\pmstd{1.44} & 85.00\pmstd{1.40} & \textbf{\textcolor{Top3}{77.54\pmstd{1.60}}} & \textbf{\textcolor{Top3}{81.26\pmstd{1.35}}} & \textbf{\textcolor{Top2}{48.30\pmstd{1.83}}} & \textbf{\textcolor{Top2}{48.33\pmstd{2.87}}} & \textbf{\textcolor{Top2}{79.29\pmstd{2.79}}} & \textbf{\textcolor{Top2}{79.31\pmstd{3.73}}} \\
            \midrule
            Sheaf-NSD
                & \textbf{\textcolor{Top2}{98.23\pmstd{1.77}}} & \textbf{\textcolor{Top2}{98.26\pmstd{1.74}}} & 73.58\pmstd{1.75} & 75.84\pmstd{1.65} & 33.66\pmstd{2.08} & 34.97\pmstd{1.90} & 72.32\pmstd{1.87} & 72.32\pmstd{1.82} \\
            \textbf{HetSheaf (ours)}
                & \textbf{\textcolor{Top1}{98.69\pmstd{1.31}}} & \textbf{\textcolor{Top1}{98.68\pmstd{1.32}}} & \textbf{\textcolor{Top1}{80.55\pmstd{1.55}}} & \textbf{\textcolor{Top2}{82.14\pmstd{1.42}}} & \textbf{\textcolor{Top1}{62.36\pmstd{2.14}}} & \textbf{\textcolor{Top1}{63.30\pmstd{1.94}}} & \textbf{\textcolor{Top1}{79.42\pmstd{0.87}}} & \textbf{\textcolor{Top1}{79.42\pmstd{0.83}}} \\
            \bottomrule
        \end{tabular}%
    }%
\end{table*}

\noindent Finally, \Cref{app:universal_stalk_space_late_token,app:implementation_details} provides a complementary systems-level view, including runtime and parameter-count comparisons, showing that the empirical gains are obtained with modest computational overhead. Taken together, the results support three conclusions. First, explicitly encoding heterogeneity in the inferred sheaf yields systematic gains over both type-agnostic sheaf models and strong heterogeneous GNN baselines. Moreover these results extend to graph-level inductive reasoning, providing empirical support for the proposed \textsc{SheafPool}. Third, the advantage of \textsc{HetSheaf} is robust across different dataset regimes.


\section{Conclusion}
\label{sec:conclusion}
We presented \textsc{HetSheaf}, a model-agnostic framework that captures graph heterogeneity through cellular sheaves to enable principled type-specific interactions. By explicitly conditioning restriction maps on node and edge types, \textsc{HetSheaf} avoids the need for increasingly specialized architectures. To enable graph-level learning, we introduced \textsc{SheafPool}, a stalk-space readout that makes sheaf-network graph classification effective, improving Macro F1 by up to 42 percentage points over mean pooling. Empirically, \textsc{HetSheaf} achieves state-of-the-art results on the Heterogeneous Graph Benchmark, reaching up to 95.39\% Macro F1 in node classification and 99.62\% in link prediction. It outperforms specialized heterogeneous GNNs by up to 2 percentage points while being up to $10\times$ more parameter-efficient. Despite the limitations discussed in \Cref{sec:limitations}, \textsc{HetSheaf} establishes a robust, highly efficient foundation for heterogeneous sheaf learning that naturally extends to hypergraphs and broader topological domains.









\vspace{-0.2cm}
\acks{
This work was supported in part by NSF under Grants CNS-2312875 and OAC-2530896, DARPA under Cooperative Agreement D25AC00374-00, ONR under Award N00014-23-1-2221, and AFOSR under Award FA9550-23-1-0261. Alessio Borgi acknowledges travel support from ELIAS (GA no 101120237), G-Research Early Career Award November 2025, Sapienza Project EVOLVE, TROPHY ultrafast holographic FTIR microscopy Horizon Europe (2022–26) and CHARM coherent Raman imaging for precision medicine Horizon Europe, for funding. }

\bibliography{references}
\clearpage
\appendix
\crefalias{section}{appendix}
\crefalias{subsection}{appendix}
\crefalias{subsubsection}{appendix}
\begin{center}
    {\LARGE \bfseries Heterogeneous Sheaf Neural Networks\par}
    \vspace{0.4em}
    {\Large Supplementary Material\par}
\end{center}

\vspace{0.4em}
\begingroup
    \etocsetlocaltop{chapter}
    \etocsetnexttocdepth{subsection}
    \localtableofcontents*
\endgroup
\vspace{1em}
\clearpage
\section{Extended Related Work}
\noindent This appendix provides the extended version of the discussion in Section~\ref{sec:related_work}.

\subsection{Homogeneous and Heterogeneous Graph Neural Networks}
A wide range of GNNs have been proposed for \emph{homogeneous} graphs~\citep{hamiltonGraphRepresentationLearning2020,kipf2016semi,liGatedGraphSequence2017,velickovicGraphAttentionNetworks2018,defferrard2016convolutional,corso2020pna}. Although these models can be directly applied to heterogeneous graphs by ignoring node and edge types, they are not explicitly designed to capture the type-specific structure of heterogeneous relational data. While in homogeneous graphs all nodes/edges encode the same type of data, in \emph{Heterogeneous GNNs} nodes and edges can represent different types of information. Several HGNNs have been proposed: a specific typology of them uses \emph{meta-paths}~\citep{sunPathSimMetaPathbased2011,sunMiningHeterogeneousInformation2012,dongMetapath2vecScalableRepresentation2017}, i.e., pre-defined node- or edge-type patterns that aim to capture important semantic information. HAN~\citep{wangHeterogeneousGraphAttention2019a} employs a hierarchical attention mechanism to capture node-level importance among nodes and semantic-level importance across meta-paths. MAGNN~\citep{fuMAGNNMetapathAggregated2020} combines several meta-path encoders to aggregate all information along the meta-path. Both HAN and MAGNN require meta-paths to be generated manually in advance. Graph Transformer Networks (GTNs)~\citep{yunGraphTransformerNetworks2019} automate this process by learning meta-paths with graph transformation layers. However, Meta-paths are computationally expensive, making them infeasible for graphs with many distinct edge types.
Instead, the graph may be treated as a relational graph. RSHN~\citep{zhuRelationStructureAwareHeterogeneous2019} constructs a coarsened line graph to generate edge features, and then applies a Message Passing Neural Network (MPNN)~\citep{gilmerNeuralMessagePassing2017} to propagate node and edge features. R-GCN~\citep{schlichtkrullModelingRelationalData2018} splits the heterogeneous graph into the subgraph induced by each edge type, then applies a standard GCN to each subgraph. HGT~\citep{huHeterogeneousGraphTransformer2020} is a graph transformer that can process large heterogeneous graphs using subgraph sampling. HetGNN~\citep{zhangHeterogeneousGraphNeural2019} uses random walks to sample strongly correlated heterogeneous neighbours and then applies an RNN for representation learning. SlotGAT~\citep{zhouSlotGATSlotbasedMessage2023} uses a slot-based message passing approach with an attention mechanism for aggregation. SimpleHGNN~\citep{lvAreWeReally2021a}, applies a series of additional techniques to lift GATs to heterogeneous domains and Space4HGNN~\citep{zhaoSpace4HGNNNovelModularized2022} defines a unified design space of modular components for heterogeneous GNNs and allows exhaustive evaluation of many combinations of techniques.
More recently, HAGNN~\citep{zhu2025hagnn}  combines meta-path-based intra-type aggregation with meta-path-free inter-type aggregation through a fused meta-path graph, while HGNN-AR\textsuperscript{2}~\citep{lin2025adaptivehetero} adaptively reconstructs relations to address same-category connection deficiencies in meta-path methods. Instead of introducing a new architecture for heterogeneous graphs, we propose shifting from representing heterogeneous relational data as graphs to cellular sheaves, which preserve all type-specific information and enable powerful processing with sheaf-specific tools.

\subsection{Sheaf Neural Networks}
Standard GNNs are known to suffer from limitations such as oversmoothing~\citep{caiNoteOverSmoothingGraph2020}, oversquashing~\citep{alon2021on} and degradation of performances in heterophilic settings, i.e., when adjacent nodes belong to different classes or carry contrasting attributes. Cellular sheaf theory~\citep{shepard1985cellular,curry2014sheaves} equips graphs with local feature spaces, or \emph{stalks}, together with linear restriction maps along edges, thereby providing a principled geometric framework for modelling and alleviating these pathologies. Sheaf Neural Networks were originally introduced using a hand-crafted sheaf with a single dimensionality in \cite{hansen2020sheaf}, and further improved by allowing the sheaf to be learned using a learnable parametric function in \citet{bodnarNeuralSheafDiffusion2022}. Since then, sheaf-based graph learning has been extended in several directions: attention mechanisms over sheaves~\citep{barberoSheafAttentionNetworks2022}, connection Laplacians learned during preprocessing~\citep{pmlr-v196-barbero22a}, sheaf-based positional encodings~\citep{pmlr-v228-he24a}, nonlinear diffusion processes~\citep{zaghen2024nonlinear}, joint diffusion biases~\citep{caralt2024joint}, polynomial spectral filtering \citep{borgi2026polynomialneuralsheafdiffusion}, and directional extensions~\citep{fiorini2026sheaves}. Beyond standard graphs, the sheaf framework has also been extended to hypergraphs~\citep{dutaSheafHypergraphNetworks2023a, mule2026directional}, recommendation systems~\citep{purificatoSheaf4RecSheafNeural2025}, federated learning~\citep{nguyen2024federated}, cooperative sheaf dynamics~\citep{ribeiro2026cooperative}, and more general topological constructions via copresheaves~\citep{hajij2025copresheaf}. Existing sheaf neural networks have primarily focused on node-level tasks, leaving graph-level prediction largely unexplored. Indeed, graph-level readout in sheaf networks is more subtle than standard aggregation, since a valid pooling operator must not only be permutation-invariant over nodes, but also invariant to node-wise stalk reparameterisations. Existing graph pooling methods~\citep{grattarola2024poolings,huaHighorderPoolingGraph2022,liGatedGraphSequence2017,lee2019attPooling} satisfy the former requirement but not the latter, and therefore cannot be directly applied in a principled way to sheaf representations.

\subsection{Cellular Sheaves on Graphs}
Let $\gG = (\gV,\gE)$ be a finite undirected graph endowed with an arbitrary orientation on its edges. A \emph{cellular sheaf} $\gF$ on $\gG$ assigns a finite-dimensional vector space, or \emph{stalk}, $\gF(u)$ to each node $u \in \gV$ and $\gF(e)$ to each edge $e \in \gE$, together with linear \emph{restriction maps} $\gF_{u \trianglelefteq e} : \gF(u) \to \gF(e)$ for every incident node-edge pair $u \trianglelefteq e$. By stacking the node and edge stalks, we obtain the spaces of $0$- and $1$-cochains, $C^0(\gG,\gF) \coloneqq \bigoplus_{u \in \gV} \gF(u), \quad C^1(\gG,\gF) \coloneqq \bigoplus_{e \in \gE} \gF(e)$. Given an oriented edge $e=(u \to v)$, these spaces induce the \emph{coboundary map} $\delta : C^0(\gG,\gF) \to C^1(\gG,\gF)$, defined edgewise by $\delta(\rvx)_e \coloneqq \gF_{v \trianglelefteq e}\rvx_v - \gF_{u \trianglelefteq e}\rvx_u$. The coboundary measures disagreement between neighbouring node assignments after they are mapped into the shared edge space~\citep{hansen2019toward}. The corresponding \emph{sheaf Laplacian} $L_{\gF} \coloneqq \delta^\top \delta$ and its normalised version is then defined node-wise, as: 
\vspace{-0.2cm}
\begin{equation}
    \Delta_{\gF}(\rvx)_u
    \coloneqq
    D^{-1/2}
    \left[
    \sum_{u,v \trianglelefteq e}
    \gF_{u \trianglelefteq e}^{\top}
    \bigl(
        \gF_{u \trianglelefteq e}\rvx_u
        -
        \gF_{v \trianglelefteq e}\rvx_v
    \bigr)
    \right]
    D^{-1/2}
\end{equation}
where $D$ is the block diagonal of $L_{\gF}$. As for the other sheaf neural networks, we assume all stalks have the same dimension $d$, i.e.\ $\gF(u)=\mathbb{R}^d$ and $\gF(e)=\mathbb{R}^d$, so that $L_{\gF} \in \mathbb{R}^{dn \times dn}$, with $n=|\gV|$. In the special case where all stalks are $\mathbb{R}$, and all restriction maps are the identity, the sheaf Laplacian reduces to the standard graph Laplacian. Generalizing the 0-cochains to $f$ feature channels results in a feature matrix $\rmX \in \mathbb{R}^{nd \times f}$, enabling \emph{Neural Sheaf Diffusion (NSD)}~\citep{bodnarNeuralSheafDiffusion2022}, to learn restriction maps via a parametric function $\gF_{u \trianglelefteq e \coloneqq (u,v)}
=
\boldsymbol{\Phi}(\rvx_u,\rvx_v)
=
\MLP(\rvx_u \| \rvx_v)$, with $\|$ denoting vector concatenation.

\section{Stalk-Space Late-token Readout: Formalisation and Properties}
\label{app:universal_stalk_space_late_token}

This appendix provides a formal treatment of \textsc{SheafPool}, our graph-level readout for sheaf neural networks. We proceed in four steps. First, we formalise graph classification in the presence of stalk gauges and explain why standard flatten-and-pool readouts are not well-defined in this setting. Second, we derive the proposed \emph{Stalk-Space Late-token Readout} and prove its permutation invariance and gauge compatibility under natural identifiability assumptions. Third, we compare the proposed \emph{late, receive-only token} to the \emph{classification token} used in Vision Transformers (ViTs), clarifying why the latter cannot be imported naively into sheaf encoders. Finally, following the multiset perspective of expressive GNN theory, we state and prove an \emph{injective readout theorem} for the canonicalised stalk descriptor space. This theorem formalises that injective graph-level pooling is possible in principle after gauge canonicalisation, and our practical late-token energy head should be viewed as a structured, low-overhead instantiation of this broader family.

\subsection{Problem Setting and Gauge Quotient Viewpoint}
\label{app:problem_setting_graph_classification}
Let $G_g=(V_g, E_g)$ be a graph and let a sheaf neural encoder produce node embeddings:
\begin{equation}
    \rmH_i \in \mathbb{R}^{d\times C} \quad \text{for} 
    \quad i\in V_g
\end{equation}
where $d$ is the stalk dimension and $C$ is the channel dimension. For graph classification, the goal is to construct a readout that summarises the graph into a single graph representation:
\begin{equation}
    \mathcal R : \{\rmH_i\}_{i\in V_g} \mapsto \rvg_g \in \mathbb{R}^C
\end{equation}

In ordinary GNNs, all node embeddings live in the same latent space, usually $\mathbb{R}^{C}$. To obtain a graph-level representation, we can simply pool the set of node embeddings ${\rvh_i},{i\in V_g}$ using a permutation-invariant operation such as sum, mean, or max pooling. In a sheaf model, the situation is different: the hidden state of node $i$ lives in its own local vector space, called the stalk $\gF(i)$. After choosing a basis for this stalk, the node state can be written as a matrix $\rmH_i \in \mathbb{R}^{d \times C}$. However, this coordinate representation is not intrinsic: different nodes may use different bases, and there is generally no canonical basis shared by all stalks.
Therefore, if we change the basis of the stalk $\gF(i)$ by a matrix $\rmG_i \in \mathbb{G}_i \subseteq GL(d)$, the same underlying sheaf signal is represented in different coordinates as:
\begin{equation}
    \rmH_i \mapsto \rmH_i' := \rmG_i \rmH_i \quad \text{with} \quad \rmG_i \in \mathbb G
\end{equation}
where $\mathbb G$ depends on the restriction-map family: a diagonal subgroup of $GL(d)$ in the diagonal case, $O(d)$ or a metric-preserving subgroup in the orthogonal/bundle case, and $GL(d)$ in the general case. We refer to this action as a \emph{local gauge action} on the stalk.

\begin{definition}[Gauge Orbit]
For a node embedding $\rmH\in\mathbb{R}^{d\times C}$, define its gauge orbit as:
\begin{equation}
    [\rmH]_{\mathbb G}
    :=
    \{\rmG\rmH : \rmG\in\mathbb G\}
\end{equation}
A graph-level readout is \emph{gauge-invariant} if it depends only on the multiset of node-wise gauge orbits.
\end{definition}
Thus, a principled sheaf readout must satisfy two requirements simultaneously:
\begin{enumerate}
    \item \textbf{Permutation invariance over nodes:} reordering the nodes of a graph must not change the output.
    \item \textbf{Gauge compatibility / invariance:} equivalent local coordinate realisations of the same stalk signal must produce the same graph descriptor.
\end{enumerate}
\begin{definition}[Gauge-compatible graph readout]
Let $\mathbb G := \prod_{i\in V_g}\mathbb G_i$ be the product of the admissible node-wise gauge groups, where $\mathbb G_i$ is determined by the restriction-map family (e.g.\ diagonal, orthogonal, or general invertible transforms). A graph-level readout
\begin{equation}
    \mathcal R : \{\rmH_i\}_{i\in V_g} \mapsto \rvg_g
\end{equation}
is said to be \emph{gauge-compatible} if
\begin{equation}
    \mathcal R(\{\rmG_i\rmH_i\}_{i\in V_g})
    =
    \mathcal R(\{\rmH_i\}_{i\in V_g})
    \qquad
    \forall (\rmG_i)_{i\in V_g} \in \mathbb G,
\end{equation}
and \emph{permutation-invariant} if its value is unchanged under any reordering of the node index set $V_g$.
\end{definition}

\begin{remark}
In our implementation, gauge compatibility is achieved through a normalisation-and-alignment procedure, similar to \cite{sevetlidis2026gauge}. We first apply node-wise whitening, which places each stalk representation in a normalised coordinate system and reduces general $GL(d)$ gauge freedom to a residual orthogonal ambiguity, namely, removes scale and shear effects. However, each whitened stalk may still differ by an orthogonal rotation or reflection. To make node representations comparable before pooling, we then align all whitened stalks to a common learnable reference, or anchor, by solving an anchor-based Procrustes problem. This step selects, for each node, the orthogonal transformation that best matches its whitened stalk representation to the anchor. Once all stalks are expressed in this shared aligned frame, we apply a permutation-invariant pooling operator to obtain the graph-level descriptor. This design makes the readout robust to local changes of basis. The invariance we obtain here is approximate in finite samples and may degrade when the stalk covariance matrices are ill-conditioned.
\end{remark}

\subsection{Flatten-and-Pool fails for Sheaf Neural Networks}
\label{app:why_naive_pooling_fails}

In sheaf neural networks, since each node representation $\rmH_i \in \mathbb{R}^{d\times C}$ lives in a \emph{ stalk space}, two tensors that represent the same underlying local signal may differ by a transformation of the form $\rmH_i \mapsto \rmG_i \rmH_i$, with $\rmG_i$ taken from the relevant gauge group. Any pooling rule that acts directly on the raw coordinates of $\rmH_i$ therefore risks depending on arbitrary local frame choices rather than on the intrinsic sheaf signal itself. This issue is particularly important after flattening. Indeed, vectorising $\rmH_i$ identifies the stalk and channel axes with a single ambient Euclidean coordinate system, thereby implicitly assuming that all nodes are already expressed in a common stalk basis. But this assumption is precisely what fails in the sheaf setting: different nodes may carry equivalent information in different local coordinates. As a result, applying a standard graph readout to $\mathrm{vec}(\rmH_i)\in\mathbb{R}^{dC}$ mixes together quantities that are not canonically comparable across nodes.

From a \emph{geometric viewpoint}, flatten-and-pool collapses the quotient structure induced by the local gauge action. A correct sheaf readout should first construct canonical representatives of these classes before aggregation, and naive flattening does not do that. Therefore, even if the encoder has learned meaningful stalk-structured node states, a basis-dependent readout may destroy this structure at the final stage and produce graph embeddings that vary under purely representational changes. 

\begin{proposition}[Flatten-and-pool is not gauge-invariant]
\label{prop:flatten_fails}
Let $\mathrm{vec}(\cdot)$ denote vectorization. Consider any readout that pools flattened node embeddings through a coordinate-wise operator such as sum, mean, max, or standard attention on $\mathrm{vec}(\rmH_i)\in\mathbb{R}^{d\times C}$. Then, this readout is not invariant under node-wise gauge transformations $\rmH_i\mapsto \rmG_i\rmH_i$.
\end{proposition}

\begin{proof}
Using the standard Kronecker identity, we have that: 
\begin{equation}
    \mathrm{vec}(\rmG\rmH)
    =
    (\rmI_C \otimes \rmG)\,\mathrm{vec}(\rmH)
\end{equation}
Hence, under a node-wise gauge transform we obtain:
\begin{equation}
    \mathrm{vec}(\rmH_i')
    =
    (\rmI_C \otimes \rmG_i)\,\mathrm{vec}(\rmH_i)
\end{equation}
If we consider, for instance, sum pooling, we get:
\begin{equation}
    \rvg_{\mathrm{sum}}
    :=
    \sum_{i\in V_g}\mathrm{vec}(\rmH_i)
\end{equation}
After the reparameterisation, we then have: 
\begin{equation}
    \rvg_{\mathrm{sum}}'
    =
    \sum_{i\in V_g}(\rmI_C\otimes \rmG_i)\,\mathrm{vec}(\rmH_i)
\end{equation}
which is generically different from $\rvg_{\mathrm{sum}}$, unless all $\rmG_i=\rmI_d$ or a non-generic cancellation occurs. Mean pooling is just a rescaled version of sum pooling and fails for the same reason. Max pooling is not equivariant under general invertible linear maps, and thus coordinate-wise maxima are also basis-dependent. Likewise, attention on flattened features changes because both attention logits and aggregated values depend on the chosen coordinates unless the scoring and value maps are themselves gauge-invariant. Therefore, direct flatten-and-pool is not a principled sheaf readout.
\end{proof}

The proposition shows that one must either pool purely gauge-invariant node descriptors or first canonicalise stalk frames before pooling in stalk space. The former is robust but collapses the stalk structure too early, while the latter is the route we adopt.

\subsection{Design Principle: Canonicalise and Pool}
\label{app:design_trajectory}

A valid readout for sheaf graph classification should satisfy the following desiderata: it must support graph-level prediction, it must remain meaningful for nontrivial stalk dimension $d>1$, it must remain valid across diagonal, orthogonal, and general restriction-map families and it should preserve stalk-structured information as long as possible before the final graph compression.

We have seen that the natural alternatives are indeed insufficient. First, standard pooling on flattened embeddings is computationally simple but fails Proposition~\ref{prop:flatten_fails}. Second, direct stalk-space pooling with energy-based weights is valid only if the residual gauge is orthogonal, and therefore does not extend to the general $GL(d)$ case. Third, invariant-only pooling, for example pooling per-node channel energies, is universal but collapses the stalk dimension before cross-node interactions in stalk space can be exploited.

Our solution is a \emph{two-stage canonicalisation followed by stalk-space aggregation}. We first use per-node whitening to remove the non-orthogonal part of the local gauge. We then use anchor-based orthogonal Procrustes alignment to express all nodes in a common stalk frame. Only after this canonicalisation, we pool in $\mathbb R^{d\times C}$. A late, receive-only token is then used to add graph-level expressive capacity without altering the encoder dynamics. The proposed \emph{Stalk-Space Late-token Readout} has already been described in \Cref{sec:universal_stalk_space_late_token}, but now we are going to see it again together with mathematical derivations and proofs. 

\subsection{Proposed Stalk-Space Late-token Readout}
\label{app:readout_definition}

Given node embeddings $\rmH_i\in\mathbb{R}^{d\times C}$, we define the readout: 
\begin{equation}
    \mathcal R : \{\rmH_i\}_{i\in V_g} \longmapsto \rvg_g\in\mathbb{R}^C
\end{equation}
through the five-stage pipeline: 
\begin{equation}
    \rmH_i
    \xrightarrow{\text{whiten}}
    \rmZ_i
    \xrightarrow{\text{align}}
    \widetilde{\rmZ}_i
    \xrightarrow{\text{attn pool}}
    \rmP_g
    \xrightarrow{\text{late token}}
    \widehat{\rmT}_g
    \xrightarrow{\text{energy}}
    \rvg_g
\end{equation}

\paragraph{Step 1: Per-node whitening.}
For each node, we define the regularised stalk-side Gram matrix as: 
\begin{equation}
    \rmS_i
    =
    \rmH_i\rmH_i^\top + \varepsilon \rmI_d
    \in\mathbb{R}^{d\times d}
    \label{eq:whitening}
\end{equation}
where the term $\varepsilon \mathbf I_d$ acts as a Tikhonov-style regulariser that yields a slightly smoothed, but well-defined, whitening transform. Its purpose is to ensure that Eq.~\ref{eq:whitening} remains positive definite, even when $\mathbf H_i \mathbf H_i^\top$ is singular or ill-conditioned. In particular, if some directions in stalk space have very small variance, the inverse square root $\mathbf S_i^{-1/2}$ may otherwise be unstable or undefined. By shifting all eigenvalues of $\mathbf H_i \mathbf H_i^\top$ by $\varepsilon$, this term stabilises the whitening operation, prevents the amplification of near-zero directions, and makes the computation numerically robust. Then, the whitened representation is: 
\begin{equation}
    \rmZ_i
    =
    \rmS_i^{-1/2}\rmH_i
    \in\mathbb{R}^{d\times C}
\end{equation}
When $\varepsilon=0$ and $\rmH_i\rmH_i^\top$ is invertible, we have $\rmZ_i\rmZ_i^\top=\rmI_d$. Thus, whitening removes anisotropic scaling and shearing in stalk space and reduces the ambiguity to an orthogonal one.

\paragraph{Step 2: Anchor-based Procrustes alignment.}
After whitening, different nodes may still differ by a residual orthogonal transform. We fix a shared learnable anchor $\rmA\in\mathbb{R}^{d\times C}$ and solve:
\begin{equation}
    \rmR_i^\star
    =
    \arg\min_{\rmR\in O(d)}
    \|\rmR^\top \rmZ_i - \rmA\|_F^2
\end{equation}
The aligned representation is then done by: 
\begin{equation}
    \widetilde{\rmZ}_i
    =
    {\rmR_i^\star}^\top \rmZ_i
\end{equation}

\paragraph{Step 3: Invariant attention weights.}
We define a per-node invariant descriptor using channel-wise energies of the whitened features:
\begin{equation}
    \rvu_i
    =
    \mathrm{diag}(\rmZ_i^\top\rmZ_i)
    \in\mathbb{R}^C
\end{equation}
A scorer $s_\theta:\mathbb{R}^C\to\mathbb{R}$ gives graph-wise attention weights:
\begin{equation}
    \alpha_i
    =
    \frac{\exp(s_\theta(\rvu_i))}
    {\sum_{j\in V_g}\exp(s_\theta(\rvu_j))}
\end{equation}

\paragraph{Step 4: Stalk-space pooling and a late receive-only token.}
We pool directly in the stalk space:
\begin{equation}
    \rmP_g
    =
    \sum_{i\in V_g}\alpha_i \widetilde{\rmZ}_i
    \in\mathbb{R}^{d\times C}
\end{equation}
A learnable late token $\rmT\in\mathbb{R}^{d\times C}$ is then updated only \emph{after} encoder message passing:
\begin{equation}
    \widehat{\rmT}_g
    =
    \phi([\rmT,\rmP_g])
    \in\mathbb{R}^{d\times C}
\end{equation}
where $\phi$ is a small channel-wise MLP with layer normalisation.

\paragraph{Step 5: Final invariant graph descriptor.}
We finally collapse the stalk dimension by channel-wise energies:
\begin{equation}
    \rvg_g
    =
    \mathrm{diag}(\widehat{\rmT}_g^\top\widehat{\rmT}_g)
    \in\mathbb{R}^C,
    \qquad
    [\rvg_g]_c
    =
    \sum_{a=1}^d (\widehat{\rmT}_g)_{a,c}^2
\end{equation}

\subsubsection{Whitening Removes the \texorpdfstring{$GL(d)$}{GL(d)} Gauge up to an Orthogonal Factor}
\label{app:whitening_gauge}

We first state the exact whitening equivariance result in the ideal full-row-rank setting, and then discuss the practically relevant regularised case.

\begin{theorem}[Exact whitening canonicalises up to $O(d)$ in the full-rank case]
\label{thm:whitening}
Fix a node and let $\rmH\in\mathbb{R}^{d\times C}$. Assume that $\rmH$ has full row rank, so that:
\begin{equation}
    \rmS := \rmH\rmH^\top \succ 0
\end{equation}
Let's define:
\begin{equation}
    \rmZ = \rmS^{-1/2}\rmH
\end{equation}
Let $\rmH'=\rmG\rmH$ with $\rmG\in GL(d)$, and define:
\begin{equation}
    \rmS'=\rmH'{\rmH'}^\top = \rmG\rmS\rmG^\top,
    \qquad
    \rmZ' = (\rmS')^{-1/2}\rmH'
\end{equation}
Then there exists $\rmQ\in O(d)$ such that:
\begin{equation}
    \rmZ' = \rmQ \rmZ
\end{equation}
\end{theorem}

\begin{proof}
Let $\rmM := \rmG\rmS^{1/2}$. Then, we have that: 
\begin{equation}
    \rmS' = \rmG\rmS\rmG^\top = \rmM\rmM^\top
\end{equation}
By the polar decomposition, there exist $\rmQ\in O(d)$ and $\rmP\succ 0$ such that:
\begin{equation}
    \rmM = \rmQ\rmP
\end{equation}
Hence:
\begin{equation}
    \rmS' = \rmQ\rmP^2\rmQ^\top,
    \qquad
    (\rmS')^{-1/2} = \rmQ\rmP^{-1}\rmQ^\top
\end{equation}
Therefore, we get:
\begin{align}
    \rmZ'
    &=
    (\rmS')^{-1/2}\rmH'
    =
    (\rmS')^{-1/2}\rmG\rmH
    =
    (\rmS')^{-1/2}\rmG\rmS^{1/2}\rmS^{-1/2}\rmH
    \nonumber\\
    &=
    (\rmS')^{-1/2}\rmM\rmZ
    =
    \rmQ\rmP^{-1}\rmQ^\top \rmQ\rmP \rmZ
    =
    \rmQ\rmZ
\end{align}
Thus, whitening removes the non-orthogonal part of the $GL(d)$ gauge, leaving only a residual orthogonal ambiguity.
\end{proof}

\begin{remark}[On the full-rank assumption]
The matrix $\rmH\rmH^\top$ is always positive semidefinite, but it is positive definite only when $\rmH$ has full row rank. This need not hold automatically in learned sheaf embeddings, since channel mixing and feature collapse may produce rank-deficient or ill-conditioned stalk-side Gram matrices.
\end{remark}

\begin{remark}[Regularized case used in practice]
In practice, we use the regularised matrix:
\begin{equation}
    \rmS_\varepsilon := \rmH\rmH^\top + \varepsilon \rmI_d,
    \qquad \varepsilon > 0
\end{equation}
which is always symmetric positive definite, regardless of the rank of $\rmH$. The corresponding whitening transform:
\begin{equation}
    \rmZ_\varepsilon := \rmS_\varepsilon^{-1/2}\rmH
\end{equation}
is therefore always well defined and numerically stable. However, because:
\begin{equation}
    \rmH'\rmH'^\top + \varepsilon \rmI_d
\neq
\rmG(\rmH\rmH^\top + \varepsilon \rmI_d)\rmG^\top
\end{equation}
in general, the exact identity $\rmZ'_\varepsilon=\rmQ\rmZ_\varepsilon$ no longer holds. The regularised whitening should therefore be viewed as a stable approximation to the ideal canonicalisation result of Theorem~\ref{thm:whitening}.
\end{remark}

\subsubsection{Orthogonal Procrustes Alignment}
\label{app:procrustes}

We next derive the alignment step.

\begin{proposition}[Closed-form Procrustes solution]
\label{prop:procrustes}
Given $\rmZ\in\mathbb{R}^{d\times C}$ and $\rmA\in\mathbb{R}^{d\times C}$, consider:
\begin{equation}
    \rmR^\star
    =
    \arg\min_{\rmR\in O(d)}
    \|\rmR^\top \rmZ - \rmA\|_F^2
\end{equation}
Let $\rmC:=\rmZ\rmA^\top$ and let its singular value decomposition be $\rmC=\rmU\bm{\Sigma}\rmV^\top$. Then:
\begin{equation}
    \rmR^\star = \rmU\rmV^\top
\end{equation}
is an optimiser. If one restricts to $\rmR\in SO(d)$, then an optimizer is: 
\begin{equation}
    \rmR^\star = \rmU \rmD \rmV^\top
\end{equation}
where $\rmD$ is diagonal with last entry $\det(\rmU\rmV^\top)$ and all other diagonal entries equal to $1$.
\end{proposition}

\begin{proof}
Expanding the objective here, we get:
\begin{align}
    \|\rmR^\top \rmZ - \rmA\|_F^2
    &=
    \mathrm{tr}(\rmZ\rmZ^\top)
    +
    \mathrm{tr}(\rmA\rmA^\top)
    -
    2\,\mathrm{tr}(\rmR^\top \rmZ\rmA^\top)
\end{align}
The first two terms do not depend on $\rmR$, so minimising is equivalent to maximising:
\begin{equation}
    \mathrm{tr}(\rmR^\top \rmC),
    \qquad \rmC=\rmZ\rmA^\top
\end{equation}
Let's write $\rmC=\rmU\bm{\Sigma}\rmV^\top$. Then, we get:
\begin{equation}
    \mathrm{tr}(\rmR^\top \rmC)
    =
    \mathrm{tr}(\rmV^\top \rmR^\top \rmU \bm{\Sigma})
    =
    \mathrm{tr}(\rmM \bm{\Sigma})
\end{equation}
where $\rmM:=\rmV^\top\rmR^\top\rmU\in O(d)$. Since $\bm{\Sigma}$ is diagonal with nonnegative entries, the maximum is attained when $\rmM=\rmI_d$, which gives $\rmR^\star=\rmU\rmV^\top$. The proper-rotation case is obtained by the usual determinant correction.
\end{proof}

\subsubsection{Exact Invariance of the Canonicalised Representative}
\label{app:exact_canonicalization}

The key point is that whitening and alignment together can produce an exact canonical representative in the idealised regime.
\begin{theorem}[Exact gauge invariance of the aligned representative]
\label{thm:exact_aligned_invariance}
Assume $\varepsilon=0$, $\rmH_i\rmH_i^\top\succ 0$ for every node $i$, and that for every node $i$, the cross-covariance $\rmC_i := \rmZ_i \rmA^\top$ has full rank and a simple singular spectrum, so that the orthogonal Procrustes solution is unique. Let $\rmH_i'=\rmG_i\rmH_i$ with $\rmG_i\in GL(d)$. Let $\widetilde{\rmZ}_i$ and $\widetilde{\rmZ}_i'$ denote the aligned representatives obtained from $\rmH_i$ and $\rmH_i'$ using the same anchor $\rmA$. Then:
\begin{equation}
    \widetilde{\rmZ}_i' = \widetilde{\rmZ}_i
    \qquad
    \text{for all } i
\end{equation}
\end{theorem}

\begin{proof}
By Theorem~\ref{thm:whitening}, there exists $\rmQ_i\in O(d)$ such that:
\begin{equation}
    \rmZ_i' = \rmQ_i \rmZ_i
\end{equation}
Now let:
\begin{equation}
    \rmC_i = \rmZ_i \rmA^\top,
    \qquad
    \rmC_i' = \rmZ_i' \rmA^\top = \rmQ_i \rmC_i
\end{equation}
Suppose $\rmC_i=\rmU_i \bm{\Sigma}_i \rmV_i^\top$ is the SVD of $\rmC_i$. Then an SVD of $\rmC_i'$ is:
\begin{equation}
    \rmC_i' = (\rmQ_i\rmU_i)\bm{\Sigma}_i \rmV_i^\top
\end{equation}
By Proposition~\ref{prop:procrustes}, the unique Procrustes solutions are: 
\begin{equation}
    \rmR_i^\star = \rmU_i\rmV_i^\top,
    \qquad
    {\rmR_i'}^\star = (\rmQ_i\rmU_i)\rmV_i^\top = \rmQ_i \rmR_i^\star
\end{equation}
Therefore, we get: 
\begin{equation}
    \widetilde{\rmZ}_i'
    =
    ({\rmR_i'}^\star)^\top \rmZ_i'
    =
    (\rmQ_i\rmR_i^\star)^\top \rmQ_i \rmZ_i
    =
    {\rmR_i^\star}^\top \rmZ_i
    =
    \widetilde{\rmZ}_i
\end{equation}
Hence, the aligned representative is invariant to the original $GL(d)$ gauge action.
\end{proof}

\begin{corollary}[Invariance of the energy descriptor]
\label{cor:ui_invariant}
Under the assumptions of Theorem~\ref{thm:exact_aligned_invariance}, the descriptor $\rvu_i = \mathrm{diag}(\rmZ_i^\top \rmZ_i)$ is gauge-invariant.
\end{corollary}

\begin{proof}
By Theorem~\ref{thm:whitening}, $\rmZ_i'=\rmQ_i\rmZ_i$ with $\rmQ_i\in O(d)$, hence:
\begin{equation}
    {\rmZ_i'}^\top \rmZ_i'
    =
    \rmZ_i^\top \rmQ_i^\top \rmQ_i \rmZ_i
    =
    \rmZ_i^\top \rmZ_i
\end{equation}
Taking diagonals proves the claim.
\end{proof}

\subsubsection{Permutation Invariance and Gauge Invariance of the Readout}
\label{app:perm_invariance}

\begin{theorem}[Permutation and gauge invariance of the readout]
\label{thm:readout_invariance}
Assume the setting of Theorem~\ref{thm:exact_aligned_invariance}. Then the readout can be written as: 
\begin{equation}
    \rvg_g
    =
    \mathrm{diag}\!\left(
    \phi\!\left[
    \rmT,
    \sum_{i\in V_g}\alpha_i \widetilde{\rmZ}_i
    \right]^\top
    \phi\!\left[
    \rmT,
    \sum_{i\in V_g}\alpha_i \widetilde{\rmZ}_i
    \right]
    \right)
\end{equation}
with:
\begin{equation}
    \alpha_i
    =
    \frac{\exp(s_\theta(\rvu_i))}
    {\sum_{j\in V_g}\exp(s_\theta(\rvu_j))}
\end{equation}
This is permutation-invariant over nodes and gauge-invariant under node-wise $GL(d)$ reparameterisations.
\end{theorem}

\begin{proof}
Permutation invariance follows because the logits $s_\theta(\rvu_i)$ are computed independently at each node and the normalisation is a symmetric sum over nodes. Therefore the multiset $\{(\alpha_i,\widetilde{\rmZ}_i)\}_{i\in V_g}$ is merely reindexed under a node permutation, and:
\begin{equation}
    \sum_{i\in V_g}\alpha_i \widetilde{\rmZ}_i
\end{equation}
is unchanged. Since $\phi$, concatenation with $\rmT$, and the final energy map are deterministic functions of this pooled tensor, the output is permutation-invariant. Gauge invariance follows from Theorem~\ref{thm:exact_aligned_invariance} and Corollary~\ref{cor:ui_invariant}: under node-wise $GL(d)$ reparameterization, $\widetilde{\rmZ}_i$ and $\rvu_i$ are unchanged. Therefore the attention weights $\alpha_i$ are unchanged, so $\rmP_g$, $\widehat{\rmT}_g$, and finally $\rvg_g$ are all unchanged.
\end{proof}

\subsection{Classification Token in Vision Transformers Comparison}
\label{app:relation_vit_cls}

In order to better understand the role of the late token, we can make a comparison with \emph{classification token} (\texttt{[CLS]}) in \emph{Vision Transformers}~\citep{dosovitskiy2020image}, even if the two mechanisms play fundamentally different roles.

The reason for this difference is structural: while in a ViT, all patch tokens already live in a single global embedding space, so a global token may interact with them directly throughout the encoder, in sheaf neural networks, node states live in stalk spaces and are only defined up to node-wise basis changes. Hence a token cannot be inserted meaningfully into the encoder without first resolving or otherwise respecting the stalk gauge. Moreover, in graph message passing, inserting a token node connected to all nodes changes the propagation graph itself. This effectively introduces global shortcuts, reduces the effective graph diameter, and may accelerate oversmoothing by forcing repeated mixing through a global hub. Our token is therefore intentionally \emph{late} and \emph{receive-only}.

\begin{proposition}[The late token does not alter encoder propagation]
\label{prop:no_encoder_change}
Let $\mathcal E_\theta$ denote the sheaf encoder, so that:
\begin{equation}
    \{\rmH_i\}_{i\in V_g} = \mathcal E_\theta(G_g, X_g)
\end{equation}
Let $\mathcal R_\omega$ denote the late-token readout. Then the full model factors as:
\begin{equation}
    \widehat y_g
    =
    \mathcal D_\eta \circ \mathcal R_\omega \circ \mathcal E_\theta (G_g,X_g)
\end{equation}
and the encoder mapping $\mathcal E_\theta$ is independent of the late token parameters.
\end{proposition}

\begin{proof}
By construction, the late token is introduced only after the encoder has already produced $\{\rmH_i\}_{i\in V_g}$. It is therefore not part of the message passing graph and does not enter any propagation, diffusion, or attention update inside $\mathcal E_\theta$. Consequently the propagation operator of the encoder is unchanged.
\end{proof}

\begin{remark}
In this sense, our token is closer to a \emph{post-encoder set summarisation token} than to a standard transformer class token. Its job is not to mediate message passing, but to add expressive readout capacity after gauge-compatible aggregation has already taken place.
\end{remark}

\subsection{Injective Stalk-Space Readout on the Canonicalised Descriptor Space}
\label{app:injective_readout}

As concerns the \emph{injectivity property}, the correct place to state an injectivity theorem is \emph{after canonicalisation}, i.e.\ on the space of gauge-compatible node descriptors. This follows the same multiset philosophy used in expressive GNN theory: if one can map multisets of node descriptors injectively, then graph-level pooling can be made discriminative.

\paragraph{Canonical descriptor space.}
For each node, define the canonical descriptor as: 
\begin{equation}
    \xi_i
    :=
    \big(\mathrm{vec}(\widetilde{\rmZ}_i), \rvu_i\big)
\end{equation}
This descriptor contains both the aligned stalk tensor and the invariant energy information. After exact canonicalisation, $\xi_i$ is gauge-invariant. 

\begin{assumption}[Bounded graphs and countable descriptor universe]
\label{ass:countable}
There exists a countable set $\mathfrak X$ such that all canonical node descriptors satisfy $\xi_i\in\mathfrak X$, and there exists $B\in\mathbb N$ such that every graph satisfies $|V_g|\leq B$.
\end{assumption}
This assumption mirrors the standard bounded-multiset/countable-universe assumption used in expressivity theory for injective multiset aggregation. It is exact when descriptors are discrete, quantised, or drawn from a countable universe; in practice, with continuous features, it should be interpreted as a theorem about the idealised discrete case and as guidance for approximate learnable implementations on compact domains.

\begin{theorem}[Existence of an injective gauge-aware multiset sum]
\label{thm:injective_multiset_sum}
Under the previous assumptions, we know that there exists a function:
\begin{equation}
    \psi : \mathfrak X \to \mathbb R
\end{equation}
such that the multiset map: 
\begin{equation}
    \mathcal S(\{\xi_i\}_{i\in V_g})
    :=
    \sum_{i\in V_g}\psi(\xi_i)
\end{equation}
is injective on multisets of canonical descriptors of size at most $B$.
\end{theorem}

\begin{proof}
Because $\mathfrak X$ is countable, there exists an injective indexing map $Z : \mathfrak X \to \mathbb N$. Let's choose any integer $M>B$. We can define then:
\begin{equation}
    \psi(\xi) := M^{-Z(\xi)}
\end{equation}
We claim that:
\begin{equation}
    \mathcal S(X) = \sum_{\xi\in X}\psi(\xi)
\end{equation}
is injective on multisets $X\subset\mathfrak X$ of cardinality at most $B$. Indeed, write a multiset $X$ through its multiplicity function $m_X:\mathfrak X\to\{0,1,\dots,B\}$. Then, we get: 
\begin{equation}
    \mathcal S(X)
    =
    \sum_{\xi\in \mathfrak X} m_X(\xi)\,M^{-Z(\xi)}
\end{equation}
This is a base-$M$ expansion whose ``digits'' are exactly the multiplicities $m_X(\xi)$, each strictly smaller than $M$. Such an expansion is unique. Therefore if $\mathcal S(X)=\mathcal S(Y)$, then all digits coincide, i.e.\ $m_X(\xi)=m_Y(\xi)$ for every $\xi\in\mathfrak X$. Hence $X=Y$ as multisets, proving injectivity.
\end{proof}

\begin{corollary}[Representation of bounded multiset functions]
\label{cor:universal_multiset}
Under the same assumptions, for any function $g : \{\text{bounded multisets over }\mathfrak X\}\to \mathcal Y$, there exists a function $\rho$ such that: 
\begin{equation}
    g(X)
    =
    \rho\!\left(\sum_{\xi\in X}\psi(\xi)\right)
\end{equation}
where $\psi$ is the injective map from Theorem~\ref{thm:injective_multiset_sum}.
\end{corollary}

\begin{proof}
By Theorem~\ref{thm:injective_multiset_sum}, the quantity $\sum_{\xi\in X}\psi(\xi)$ uniquely identifies the multiset $X$. Therefore, we may also define:
\begin{equation}
    \rho\!\left(\sum_{\xi\in X}\psi(\xi)\right) := g(X)
\end{equation}
This is well-defined because the sum is injective.
\end{proof}

The previous theorems/corollaries are the exact analogues, on the canonicalised sheaf descriptor space, of the classical injective-sum multiset theorems used to analyse expressive GNNs.

\subsubsection{Injective Readout on Gauge Orbits}
\label{app:injective_on_orbits}

We now strengthen the previous result by making explicit the quotient-space structure induced by the local gauge action and by stating precisely the condition under which the canonicalisation map is not only gauge-invariant, but also \emph{orbit-separating}. This allows us to prove injectivity at the level of bounded multisets of node-wise gauge orbits, rather than only at the level of bounded multisets of canonical descriptors.

\begin{definition}[Orbit multiset]
For a graph $g$, we can define the multiset of node-wise gauge orbits as:
\begin{equation}
    \Omega_g
    :=
    \big\{[\rmH_i]_{\mathbb G} : i\in V_g \big\}
\end{equation}
where $[\rmH_i]_{\mathbb G}$ denotes the orbit of $\rmH_i$ under the local gauge group $\mathbb G$.
\end{definition}

\begin{assumption}[Orbit-separating canonicalization]
\label{ass:orbit_separating}
Let
\begin{equation}
    \mathcal C(\rmH)
    :=
    \xi
    =
    \big(\mathrm{vec}(\widetilde{\rmZ}), \rvu\big)
\end{equation}
denote the canonical descriptor produced by whitening and alignment. We assume that, on the domain of interest, $\mathcal C$ is \emph{orbit-separating}, i.e.
\begin{equation}
    \mathcal C(\rmH)=\mathcal C(\rmH')
    \quad\Longrightarrow\quad
    [\rmH]_{\mathbb G}=[\rmH']_{\mathbb G}.
\end{equation}
Equivalently, the induced map on the quotient space
\begin{equation}
    \overline{\mathcal C}:
    \mathbb{R}^{d\times C}/\mathbb G
    \to
    \mathfrak X,
    \qquad
    \overline{\mathcal C}\big([\rmH]_{\mathbb G}\big):=\mathcal C(\rmH),
\end{equation}
is injective, where $\mathfrak X$ is the canonical descriptor space.
\end{assumption}

\begin{theorem}[Injective readout on bounded multisets of gauge orbits]
\label{thm:injective_orbit_readout}
Assume Theorem~\ref{thm:exact_aligned_invariance}, Assumption~\ref{ass:countable}, and Assumption~\ref{ass:orbit_separating}. Let:
\begin{equation}
    \mathcal C(\rmH_i)
    :=
    \xi_i
    =
    \big(\mathrm{vec}(\widetilde{\rmZ}_i), \rvu_i\big)
\end{equation}
be the canonical descriptor map. Then there exists a graph readout of the form: 
\begin{equation}
    \mathcal R_{\mathrm{inj}}(\{\rmH_i\}_{i\in V_g})
    :=
    \rho\!\left(\sum_{i\in V_g}\psi(\mathcal C(\rmH_i))\right)
\end{equation}
such that:
\begin{enumerate}
    \item $\mathcal R_{\mathrm{inj}}$ is permutation-invariant over nodes;
    \item $\mathcal R_{\mathrm{inj}}$ is gauge-invariant under node-wise $GL(d)$ actions;
    \item $\mathcal R_{\mathrm{inj}}$ is injective on bounded multisets of node-wise gauge orbits, i.e.:
    \begin{equation}
        \Omega_g \neq \Omega_{g'}
        \quad\Longrightarrow\quad
        \mathcal R_{\mathrm{inj}}(\{\rmH_i\}_{i\in V_g})
        \neq
        \mathcal R_{\mathrm{inj}}(\{\rmH_j'\}_{j\in V_{g'}})
    \end{equation}
\end{enumerate}
In particular, one may choose $\rho$ to be the identity on the image of the multiset sum.
\end{theorem}

\begin{proof}
We proceed in four steps.

\medskip\noindent\textbf{Step 1: The canonical descriptor induces a well-defined map on gauge orbits.}
By Theorem~\ref{thm:exact_aligned_invariance}, if two node embeddings belong to the same gauge orbit, i.e. $\rmH' = \rmG \rmH$ for some $\rmG \in \mathbb G$, then their aligned representatives and invariant descriptors coincide. Therefore, we have $\mathcal C(\rmH') = \mathcal C(\rmH)$. Hence $\mathcal C$ is constant on gauge orbits, and so it induces a well-defined quotient map:
\begin{equation}
    \overline{\mathcal C}:
    \mathbb{R}^{d\times C}/\mathbb G
    \to
    \mathfrak X,
    \qquad
    \overline{\mathcal C}\big([\rmH]_{\mathbb G}\big)
    :=
    \mathcal C(\rmH)
\end{equation}

\noindent\textbf{Step 2: The quotient map is injective on the domain of interest.} 
By Assumption~\ref{ass:orbit_separating}, the induced map $\overline{\mathcal C}$ is injective. Therefore, two distinct gauge orbits are mapped to two distinct canonical descriptors. In particular, if $[\rmH]_{\mathbb G} \neq [\rmH']_{\mathbb G}$, then, we have:
\begin{equation}
    \overline{\mathcal C}\big([\rmH]_{\mathbb G}\big)
    \neq
    \overline{\mathcal C}\big([\rmH']_{\mathbb G}\big)
\end{equation}
Thus, on the domain of interest, the canonical descriptor does not merely forget the gauge, but it provides a faithful representative of the gauge orbit.

\noindent\textbf{Step 3: Canonical orbit multisets are injectively encoded by a multiset sum.}
Let $\Xi_g := \big\{\xi_i : i\in V_g\big\} = \big\{\mathcal C(\rmH_i): i\in V_g\big\}$ be the multiset of canonical descriptors associated with graph $g$. Since $\mathfrak X$ is countable and graph sizes are bounded by Assumption~\ref{ass:countable}, Theorem~\ref{thm:injective_multiset_sum} guarantees the existence of a function $\psi:\mathfrak X\to\mathbb{R}^m$ for some $m$ such that the multiset sum $\mathcal S(\Xi_g) := \sum_{\xi\in \Xi_g}\psi(\xi)$ is injective on bounded multisets of descriptors. Equivalently, we have that: 
\begin{equation}
    \mathcal S(\Xi_g)=\mathcal S(\Xi_{g'})
    \quad\Longrightarrow\quad
    \Xi_g=\Xi_{g'}
\end{equation}

\noindent\textbf{Step 4: Injectivity transfers from descriptor multisets to orbit multisets.}
Now suppose that two graphs $g$ and $g'$ satisfy $\mathcal R_{\mathrm{inj}}(\{\rmH_i\}_{i\in V_g}) = \mathcal R_{\mathrm{inj}}(\{\rmH_j'\}_{j\in V_{g'}})$. Choose $\rho$ to be injective on the image of the multiset sum and, in particular, one may simply take $\rho$ to be the identity. Then:
\begin{equation}
    \sum_{i\in V_g}\psi(\mathcal C(\rmH_i))
    =
    \sum_{j\in V_{g'}}\psi(\mathcal C(\rmH_j'))
\end{equation}
By injectivity of the multiset sum from Step 3, this implies that $\Xi_g = \Xi_{g'}$. Since $\overline{\mathcal C}$ is injective by Step 2, equality of descriptor multisets implies equality of the corresponding multisets of gauge orbits, i.e., $\Omega_g = \Omega_{g'}$. Taking the contrapositive, if $\Omega_g\neq\Omega_{g'}$, then:
\begin{equation}
    \mathcal R_{\mathrm{inj}}(\{\rmH_i\}_{i\in V_g})
    \neq
    \mathcal R_{\mathrm{inj}}(\{\rmH_j'\}_{j\in V_{g'}})
\end{equation}
This proves injectivity on bounded multisets of node-wise gauge orbits.

Finally, permutation invariance follows immediately from the commutativity of summation: $\sum_{i\in V_g}\psi(\mathcal C(\rmH_i))$ depends only on the multiset $\{\mathcal C(\rmH_i)\}_{i\in V_g}$ and not on any ordering of the nodes. Gauge invariance follows from Step 1, since $\mathcal C$ is constant on each gauge orbit. Therefore, the readout is permutation-invariant, gauge-invariant, and injective on bounded multisets of node-wise gauge orbits.
\end{proof}

\begin{remark}[Practical Interpretation]
Theorem~\ref{thm:injective_orbit_readout} shows that, once the local gauge has been removed through an orbit-separating canonicalisation, graph-level readout reduces to injective multiset encoding over canonical node descriptors. This is the precise sheaf analogue of injective multiset readout results in expressive graph representation learning: the only additional subtlety is that injectivity must now be established on \emph{gauge orbits}, not on raw node coordinates.
\end{remark}

\newpage
\section{Additional Experimental, Ablation, and Implementation Details}
\label{app:additional_details}

This appendix consolidates all supplementary material related to the empirical study and implementation of \textsc{HetSheaf}. We begin by describing the experimental protocol, hyperparameter search space, and datasets used throughout the paper. We then present a detailed model analysis, including ablations over heterogeneous sheaf predictors, restriction-map families, attention-based variants, and architectural choices such as depth and stalk dimension. Next, we provide broader comparisons with recent heterogeneous GNNs reported in the literature, in order to situate \textsc{HetSheaf} within the wider benchmark landscape. Finally, we collect the main implementation details, including hardware, decoders, restriction-map parameterisations, the full family of heterogeneous sheaf predictors, computational overhead, graph-level readout costs, and the heterogeneous extensions of Sheaf Attention Networks and Polynomial Neural Sheaf Diffusion. Together, these experiments complement the main paper by clarifying both the empirical behaviour and the practical design choices behind the model.

\subsection{Experimental Protocol, Hyperparameters, and Datasets}
\label{app:exp_protocol_datasets}

We gather here the information necessary to reproduce the empirical setup used in the main paper. We first describe the hyperparameter search procedure shared across the evaluated models, and then summarise both the benchmark datasets and the task-specific datasets used in our study. 

\noindent For this paper, we rely on a broad collection of heterogeneous benchmarks \citep{lvAreWeReally2021a,zhouSlotGATSlotbasedMessage2023,wangHeterogeneousGraphAttention2019a,huHeterogeneousGraphTransformer2020,yangHeterogeneousNetworkRepresentation2022, morris2020tudataset, hu2025heterobiomedical,BioERP2021,mcauleyHiddenFactors2013, leskovec2005graphs, ferriol2022routenet-erlang, ferriol2023routenet-fermi} spanning \emph{node classification}, \emph{link prediction}, and \emph{graph classification}. The suite covers citation graphs (e.g., DBLP, ACM), knowledge graphs (e.g., IMDB), user-item interaction networks (e.g., LastFM, MovieLens, Amazon, YouTube), link prediction graphs (e.g., PubMed), biological graphs (e.g., PROTEINS, ENZYMES), chemical compounds (e.g., MUTAG, NCI1), biomedical networks (e.g., CTD-DDA, DrugBank, NDFRT-DDA, String-PPI), and networking tasks (e.g., AS773, RouteNet Traffic). We note that, although PROTEINS, ENZYMES, and NCI1 contain a single edge type, they comprise multiple node types (3, 3, and 37, respectively): they therefore directly probe heterogeneous node/stalk spaces and the gauge-compatible \textsc{SheafPool} readout, while MUTAG combines several node types with four bond types, exercising both node- and edge-type conditioning.
We compare against representative homogeneous (GCN~\citep{kipf2016semi}, GAT~\citep{velickovicGraphAttentionNetworks2018}, GIN~\citep{xuHowPowerfulAre2018}, GraphSAGE~\citep{hamiltonInductiveRepresentationLearning2017}), heterogeneous (R-GCN~\citep{schlichtkrullModelingRelationalData2018}, HAN~\citep{wangHeterogeneousGraphAttention2019a}, HGT~\citep{huHeterogeneousGraphTransformer2020}), and sheaf-based baselines~\citep{bodnarNeuralSheafDiffusion2022}. In \Cref{app:external_comparisons}, we compare against recent state-of-the-art heterogeneous models, HAGNN~\citep{zhu2025hagnn}, HINormer~\citep{mao2023hinormer}, SeHGNN~\citep{yang2023sehgnn}, and SlotGAT~\citep{zhouSlotGATSlotbasedMessage2023}, re-run under a fully matched protocol, together with contextual results from the literature. We follow the standard protocols for each benchmark and use the HGB evaluation setup~\citep{lvAreWeReally2021a}. We report \emph{Macro/Micro-F1} for classification and \emph{AUPR/AUROC} for link prediction. Negative edges for link prediction are generated following prior work on heterogeneous relation prediction~\citep{fuMAGNNMetapathAggregated2020,zhangHeterogeneousGraphNeural2019,cenRepresentationLearningAttributed2019}. Model selection is performed through hyperparameter sweeps (150 runs per model type, both for baselines and for restriction map type for Sheaf-NSD and \textsc{HetSheaf}). The best validation-selected configuration of each model is then re-trained across five random seeds (42--46), and all reported results are the mean $\pm$ standard deviation across these seeds. In the following \Cref{app:exp_protocol_datasets,app:external_comparisons,app:implementation_details}, we provide detailed dataset descriptions, preprocessing choices, additional baseline comparisons from the literature, hyperparameter search spaces, hardware details, and other implementation notes. 

\subsubsection{Model Hyperparameters}
\Cref{tab:hyperparam_search_space} reports the search space used to optimise all models. For each method, we adopt the same approach usually followed by the main Sheaf Neural Networks frameworks \cite{bodnarNeuralSheafDiffusion2022}, \cite{barberoSheafAttentionNetworks2022}, \cite{borgi2026polynomialneuralsheafdiffusion} and we run large-scale Bayesian hyperparameter sweeps  (150 runs per model type, both for baselines, and for restriction map type for Sheaf-NSD and \textsc{HetSheaf})~\citep{snoekPracticalBayesianOptimization2012, onorato2024bayesianoptimizationhyperparameterstuning} and select the best configuration according to the validation metric appropriate to the task. This shared protocol allows a fair comparison between homogeneous baselines, heterogeneous GNNs, and sheaf-based models under a common tuning budget.
\begin{table}[!htbp]
    \sisetup{
        list-final-separator={,},
        list-separator={,},
        list-pair-separator={,}
    }
    \centering
    \caption{\textbf{Hyperparameter search space.}}
    \label{tab:hyperparam_search_space}
    \resizebox{0.5\textwidth}{!}{%
        \begin{tabular}{cc}
        \toprule
        \thead{Hyperparameter}& \thead{Search space}\\
        \midrule
            Sheaf type & $\braces*{\text{general}, O(d), \text{diagonal}, \text{SheafAN}}$\\
            Sheaf learner & $\braces{\text{NSD}, \text{TE}, \text{EE},  \text{NE}, \text{types}, \text{NT}, \text{ET}}$\\
            Attention Learner$^\dagger$ & $\braces{\text{Yes}, \text{No}}$ \\
            $d$ & $\braces*{\numlist{2;3;4;5}}$\\
            \# Layers & $\braces{\numlist{2;3;4;5;6;7;8}}$\\
            input\_dropout & $\brkts*{\numlist{0;0.9}}$\\
            dropout & $\brkts*{\numlist{0;0.9}}$\\
            initial\_dropout & $\brkts*{\numlist{0;0.9}}$\\
            \midrule
            Learning rate& $\braces*{\numlist{1;5}} \times \braces*{\numlist{e-5; e-4; e-3; e-2}}$\\
            Weight decay & $\braces*{\numlist{1;5}} \times \braces*{\numlist{e-5; e-4; e-3}}$\\
        \bottomrule
        \end{tabular}%
    }
    
    \vspace{3pt}
    \raggedright\footnotesize{$^\dagger$ Not supported for \textit{diagonal} sheaf type.}
\end{table}

\subsection{Additional Results and Model Analysis}
\label{app:model_analysis}

This section complements the headline results from the main paper with a more detailed analysis of where the gains of \textsc{HetSheaf} come from. We study in particular the effect of the heterogeneous sheaf predictor, the restriction-map family, the use of attention-based restriction-map constructions, and architectural choices such as depth and stalk dimension. The goal is not only to identify the strongest configuration, but also to understand which modelling components matter most for different tasks and datasets.

\subsubsection{Benchmark Datasets}
\label{app:dataset-description}
\begin{table}[htbp]
    \centering
    \caption{\bfseries{Statistics of benchmark datasets.}}
    \label{tab:dataset-stats}
    \resizebox{0.7\textwidth}{!}{\begin{tabular}{lcccccc}
    \toprule
    \emph{Node classification} & \thead{\# Nodes} & \thead{\# Node types} & \thead{\# Edges} & \thead{\# Edge types} & \thead{Target} & \thead{\# Classes} \\
    \midrule
        DBLP & 26,128 & 4 & 239,566 & 6 & author & 4 \\
        IMDB & 21,420 & 4 & 86,642  & 6 & movie  & 5 \\
        ACM  & 10,942 & 4 & 547,806 & 7 & paper  & 3 \\
    \toprule
    \emph{Link prediction} & & & & & \multicolumn{2}{c}{\thead{Target}} \\
    \midrule
        LastFM    & 20,612 & 3 & 141,521 & 3  & \multicolumn{2}{c}{user-artist}       \\
        MovieLens & 10,352 & 2 & 201,672 & 2  & \multicolumn{2}{c}{user-movie}        \\
        PubMed    & 63,109 & 4 & 244,986 & 10 & \multicolumn{2}{c}{disease-disease}   \\
        YouTube  & 2,000  & 1 & 1,177,130 & 5 & \multicolumn{2}{c}{user-user}        \\
        Amazon    & 10,099 & 1 & 148,659 & 2  & \multicolumn{2}{c}{product-product}   \\

    \toprule
    \emph{Graph classification} & \thead{\# Graphs} & \thead{\# Node types} & \thead{\# Edges} & \thead{\# Edge types} & \multicolumn{2}{c}{\thead{Target}} \\
    \midrule
        MUTAG    & 188   & 7  & 7,442   & 4 & \multicolumn{2}{c}{mutagenicity}      \\
        PROTEINS & 1,113 & 3  & 162,088 & 1 & \multicolumn{2}{c}{enzyme/non-enzyme} \\
        ENZYMES  & 600   & 3  & 19,580  & 1 & \multicolumn{2}{c}{enzyme class}      \\
        NCI1     & 4,110 & 37 & 49,205  & 1 & \multicolumn{2}{c}{anti-cancer}       \\
    \bottomrule
    \end{tabular}}
\end{table}
Each of the datasets used for node classification is publicly available as part of the Heterogeneous Graph Benchmark (\textsc{HGB})\footnote{\url{https://www.biendata.xyz/hgb/}} first introduced by \citet{lvAreWeReally2021a}. The link prediction datasets (LastFM, MovieLens, PubMed and Amazon) are also drawn from HGB. On the other hand, graph classification datasets are drawn from the TU Datasets benchmark~\citep{morris2020tudataset}, a widely used collection of graph-level prediction benchmarks spanning biological and chemical domains.
We provide brief descriptions of each dataset.
\begin{itemize}
    \item \textbf{DBLP}\footnote{\url{https://web.cs.ucla.edu/~yzsun/data/}} is a citation graph based on the DBLP computer science bibliographic database.
    It has four node types: authors, papers, venues and terms.
    The six edge types include paper-author, paper-venue, paper-term and author-venue.
    The target is to predict the class label of authors based on their research area: databases, data mining, AI or information retrieval.
    \item \textbf{IMDB}\footnote{\url{https://www.kaggle.com/datasets/karrrimba/movie-metadatacsv}} is an information graph containing movie data.
    The four node types are movies, directors, actors and keywords.
    The six edge types include movie-director, movie-actor and movie-keyword.
    Movies are classified based on their genre: action, comedy, drama, romance and thriller.
    Each movie can have multiple genres, i.e., a multi-label node classification task.
    \item \textbf{ACM}~\citep{wangHeterogeneousGraphAttention2019a} is a citation graph containing node types of authors, papers, terms and subjects.
    The edge types include paper-cite-paper, paper-author, paper-subject and paper-term.
    The target is to predict each paper's category: databases, wireless communication, or data mining.
    \item \textbf{LastFM}\footnote{\url{https://grouplens.org/datasets/hetrec-2011/}} is a knowledge graph generated from an online music website.
    The three node types are user, artist, and tags, with edge types of user-artist, user-user and artist-tag. The link prediction task aims to predict edges between users and artists.
    \item \textbf{MovieLens}\footnote{\url{https://movielens.org/}} is a heterogeneous rating graph from the MovieLens website and consists of two node types: movie and user. The task is to predict whether or not a user rates a movie.
    \item \textbf{PubMed}\footnote{\url{https://pubmed.ncbi.nlm.nih.gov}} is a biomedical knowledge graph from the HGB benchmark~\citep{lvAreWeReally2021a}, constructed from the PubMed literature database.
    It contains four node types (gene, disease, chemical and species) connected by ten edge types encoding biological relations such as gene-disease associations, chemical-chemical interactions and species-disease links.
    The link prediction task targets disease-disease edges.
    \item \textbf{Amazon} is a single-node-type product graph from the HGB benchmark~\citep{lvAreWeReally2021a} containing 10,099 items connected by co-viewing and co-purchasing relations (two edge types). The link prediction task targets product-product edges.
     \item \textbf{YouTube} is a single-node-type social network from the HGB benchmark~\citep{lvAreWeReally2021a} containing 2,000 users connected by five types of interactions. The link prediction task targets user-user edges.
    \item \textbf{MUTAG}~\citep{morris2020tudataset} is a collection of 188 mutagenic aromatic and heteroaromatic nitro compounds, with 7 atom types and 4 bond types. The task is to predict whether a compound has a mutagenic effect on a bacterium.
    \item \textbf{PROTEINS}~\citep{morris2020tudataset} is a dataset of 1,113 proteins represented as graphs, where nodes are secondary structure elements with 3 node types and edges connect amino acids that are neighbours in the sequence or in 3D space. The task is to classify proteins as enzymes or non-enzymes.
    \item \textbf{ENZYMES}~\citep{morris2020tudataset} is a dataset of 600 protein tertiary structures obtained from the BRENDA enzyme database, with 3 node types representing secondary structure elements. The task is to classify each protein into 
    one of 6 enzyme commission top-level classes.
    \item \textbf{NCI1}~\citep{morris2020tudataset} is a dataset of 4,110 chemical compounds screened for their ability to suppress or inhibit the growth of a panel of human tumour cell lines, with 37 atom types. The task is to predict anti-cancer activity.
\end{itemize}

\subsubsection{Biomedical Datasets}
\label{app:task-specific-datasets}
The following datasets are used exclusively for biomedical link prediction~\citep{hu2025heterobiomedical, BioERP2021} experiments 
and are provided in the appendix for completeness.
\begin{table}[!htbp]
    \centering
    \caption{\bfseries{Statistics of biomedical link prediction datasets.}}
    \label{tab:dataset-stats-specific}
    \resizebox{0.7\textwidth}{!}{\begin{tabular}{lcccccc}
    \toprule
    \emph{Biomedical} & \thead{\# Nodes} & \thead{\# Node types} & \thead{\# Edges} & \thead{\# Edge types} & \multicolumn{2}{c}{\thead{Target}} \\
    \midrule
        CTD-DDA      & 12,765 & 2 & 43,492  & 1 & \multicolumn{2}{c}{drug-disease} \\
        DrugBank-DDI & 2,191  & 1 & 242,027 & 1 & \multicolumn{2}{c}{drug-drug} \\
        NDFRT-DDA    & 13,545 & 2 & 4,709   & 1 & \multicolumn{2}{c}{drug-disease} \\
        String-PPI   & 15,131 & 1 & 359,776 & 1 & \multicolumn{2}{c}{protein-protein} \\
    \bottomrule
    \end{tabular}}
\end{table}

\begin{itemize}
    \item \textbf{CTD-DDA}\footnote{\url{https://github.com/Zaiwen/Link_Prediction_in_Biomedical_Network}} is a heterogeneous graph 
    derived from the Comparative Toxicogenomics Database, containing curated associations between chemical compounds and diseases. The link prediction task aims to predict drug-disease associations.
    
    \item \textbf{DrugBank}\footnotemark[\value{footnote}] is a comprehensive 
    heterogeneous graph constructed from the DrugBank database~\citep{wishart2018drugbank}, containing drugs, targets, enzymes, and carriers. The link prediction task aims to predict drug-target interactions.

    \item \textbf{NDFRT-DDA}\footnotemark[\value{footnote}] is a 
    heterogeneous graph derived from the National Drug File Reference Terminology, containing drug-disease associations curated from clinical data. The task is to predict drug-disease associations.

    \item \textbf{String-PPI}\footnotemark[\value{footnote}] is a protein-protein interaction network derived from the STRING database~\citep{szklarczyk2021string}, containing functional associations between proteins across multiple organisms. The task is to predict protein-protein interactions.
\end{itemize}

\begin{table}[!htbp]
    \centering
    \caption[Performance on heterogeneous link prediction benchmarks]{
        \textbf{Performance on heterogeneous biomedical networks link prediction benchmarks.}
        Results for the heterogeneous sheaf learners and baselines are shown. 
         We report the highest binary AUROC and AUPR scores achieved across the hyperparameter sweep for each model.
        The top three models are coloured by \textbf{\textcolor{Top1}{First}}, \textbf{\textcolor{Top2}{Second}} and \textbf{\textcolor{Top3}{Third}}.
    }
    \label{tab:biomedical_networks_results}
    \maxsizebox{\textwidth}{!}{
\resizebox{0.7\textwidth}{!}{\begin{tabular}{l S S S S S S S S}
    \toprule
    {} & \multicolumn{2}{c}{\thead{CTD-DDA}} & \multicolumn{2}{c}{\thead{DrugBank}} & \multicolumn{2}{c}{\thead{NDFRT-DDA}} & \multicolumn{2}{c}{\thead{String-PPI}} \\
    \cmidrule(lr){2-3}\cmidrule(lr){4-5}\cmidrule(lr){6-7}\cmidrule(lr){8-9}
    {} & {\thead{AUPR}} & {\thead{AUROC}} & {\thead{AUPR}} & {\thead{AUROC}} & {\thead{AUPR}} & {\thead{AUROC}} & {\thead{AUPR}} & {\thead{AUROC}} \\
    \midrule
    \multicolumn{9}{l}{\textit{Homogeneous}} \\
    GCN & \textbf{\textcolor{Top3}{99.36}} & \textbf{\textcolor{Top3}{99.09}} & 93.46 & 91.57 & 97.80 & 96.51 & 95.11 & 92.62 \\
    GAT & 89.55 & 89.92 & 88.73 & 87.46 & 94.01 & 92.88 & 94.77 & 92.26 \\
    GIN & 99.34 & 99.07 & 93.71 & 92.43 & 97.83 & 96.44 & 95.27 & 92.77 \\
    SAGE & 93.78 & 93.82 & 94.20 & 92.28 & 94.50 & 92.87 & \textbf{\textcolor{Top3}{95.94}} & 93.78 \\
    \midrule
    \multicolumn{9}{l}{\textit{Heterogeneous}} \\
    R-GCN & 99.03 & 98.93 & \textbf{\textcolor{Top3}{94.77}} & \textbf{\textcolor{Top1}{94.13}} & \textbf{\textcolor{Top2}{98.18}} & \textbf{\textcolor{Top2}{97.32}} & 95.79 & \textbf{\textcolor{Top3}{94.15}} \\
    HAN & 63.34 & 51.76 & 88.68 & 85.54 & 63.89 & 52.82 & 93.39 & 90.08 \\
    HGT & 63.32 & 51.72 & 89.95 & 87.38 & 63.97 & 52.99 & 93.07 & 90.27 \\
    \midrule
    Sheaf-NSD & \textbf{\textcolor{Top2}{99.47}} & \textbf{\textcolor{Top2}{99.23}} & \textbf{\textcolor{Top2}{94.90}} & \textbf{\textcolor{Top3}{93.90}} & \textbf{\textcolor{Top3}{98.13}} & \textbf{\textcolor{Top3}{97.02}} & \textbf{\textcolor{Top1}{96.20}} & \textbf{\textcolor{Top1}{94.40}} \\
    \textbf{HetSheaf (ours)} & \textbf{\textcolor{Top1}{99.53}} & \textbf{\textcolor{Top1}{99.31}} & \textbf{\textcolor{Top1}{94.92}} & \textbf{\textcolor{Top2}{93.93}} & \textbf{\textcolor{Top1}{98.23}} & \textbf{\textcolor{Top1}{97.67}} & \textbf{\textcolor{Top2}{96.05}} & \textbf{\textcolor{Top2}{94.24}} \\
    \bottomrule
\end{tabular}}
}
\end{table}

\Cref{tab:biomedical_networks_results} reports link prediction results on four biomedical networks. \textsc{HetSheaf} ranks first on CTD-DDA (99.53 AUPR, 99.31 AUROC), DrugBank (94.92 AUPR), and NDFRT-DDA (98.23 AUPR, 97.67 AUROC), while Sheaf-NSD leads on String-PPI (96.20 AUPR, 94.40 AUROC).
The two sheaf models consistently occupy the top two positions across all datasets, with R-GCN as the strongest non-sheaf competitor.
HAN and HGT fail on CTD-DDA and NDFRT-DDA (AUROC $\approx$ 52, near random), whereas the sheaf models remain robust.
The slight advantage of Sheaf-NSD over \textsc{HetSheaf} on String-PPI is consistent with this dataset having a single node type, which limits the benefit of heterogeneous type conditioning.
\subsubsection{Networking Datasets}
\label{app:networking-datasets}

The following datasets are used exclusively for networking link prediction and node classification experiments
and are provided in the appendix for completeness.

\begin{table}[!htbp]
    \centering
    \caption{\bfseries{Statistics of networking datasets.}}
    \label{tab:dataset-stats-networking}
    \resizebox{0.85\textwidth}{!}{\begin{tabular}{lcccccc}
    \toprule
    \emph{Networking} & \thead{\# Nodes} & \thead{\# Node types} & \thead{\# Edges} & \thead{\# Edge types} & \thead{Task} & \thead{Target} \\
    \midrule
        AS-733                  & 6,474     & 1 & 13,895      & 3 & LP & customer-provider \\
        RouteNet Traffic Models & $\sim$674 & 2 & $\sim$1,200 & 2 & NC & link congestion    \\
    \bottomrule
    \end{tabular}}
\end{table}

\begin{itemize}
    \item \textbf{AS-733}\footnote{\url{https://snap.stanford.edu/data/as-733.html}} is a
    heterogeneous graph of the Internet's Autonomous System (AS) topology, derived from BGP
    routing tables collected by the University of Oregon Route Views
    Project~\citep{leskovec2005graphs}. Nodes represent Autonomous Systems and edges
    represent BGP peering relationships, classified into three semantically distinct types:
    \emph{customer\textendash provider} (paid transit), \emph{peer\textendash to\textendash peer} (settlement-free peering),
    and \emph{provider\textendash customer} (reverse transit). The link prediction task aims to predict
    missing or future peering agreements between Autonomous Systems. The dataset contains
    6,474 nodes and 13,895 edges from the January 2, 2000 snapshot.

    \item \textbf{RouteNet Traffic Models}\footnote{\url{https://github.com/BNN-UPC/NetworkModelingDatasets}, v5}
    is a bipartite heterogeneous graph for network performance modelling, generated using the
    BNNetSimulator~\citep{ferriol2022routenet-erlang, ferriol2023routenet-fermi}. The graph consists of two node types:
    \emph{link} nodes (physical network connections with bandwidth capacity features) and
    \emph{path} nodes (logical end-to-end routes with traffic and delay features), connected
    by \emph{traverses} and \emph{belongs\_to} edges encoding the routing table. The node
    classification task predicts which links become congested (utilisation exceeding a
    threshold of 0.50, yielding a 54\%/46\% class balance). The dataset comprises five
    traffic generation patterns: Constant Bit Rate (CBR), which produces deterministic
    fixed-rate traffic; ON-OFF, which alternates between active and idle periods to model
    bursty applications; Autocorrelated Exponentials, where successive inter-arrival times
    exhibit temporal correlations capturing long-range dependence, and Modulated
    Exponentials, whose rate parameter varies over time to model changing activity levels.
    We evaluate on the \emph{All Multiplexed} configuration, which mixes all four traffic
    types simultaneously across different source-destination pairs, creating the most
    realistic scenario in which the model must handle qualitatively different congestion
    dynamics within a single graph. Training uses NSFNET (14 nodes) and GEANT2 (24 nodes)
    topologies.
\end{itemize}

\begin{table}[htbp]
    \centering
    \caption[Performance on networking benchmarks]{
        \textbf{Performance on networking benchmarks.}
        Results for the heterogeneous sheaf learners and baselines are shown.
        We report the highest binary AUROC and AUPR for AS-733 (link prediction)
        and the highest Macro F1 and Micro F1 for RouteNet Traffic Models (node classification),
        achieved across the hyperparameter sweep for each model.
        The top three models are coloured by \textbf{\textcolor{Top1}{First}}, \textbf{\textcolor{Top2}{Second}} and \textbf{\textcolor{Top3}{Third}}.
    }
    \label{tab:networking_results}
    \maxsizebox{\textwidth}{!}{
\resizebox{0.5\textwidth}{!}{\begin{tabular}{l S S S S}
    \toprule
    {} & \multicolumn{2}{c}{\thead{AS-733 (LP)}} & \multicolumn{2}{c}{\thead{RouteNet Traffic (NC)}} \\
    \cmidrule(lr){2-3}\cmidrule(lr){4-5}
    {} & {\thead{AUPR}} & {\thead{AUROC}} & {\thead{Macro F1}} & {\thead{Micro F1}} \\
    \midrule
    \multicolumn{5}{l}{\textit{Homogeneous}} \\
    GCN       & 97.25 & 96.18 & \textbf{\textcolor{Top2}{80.57}} & \textbf{\textcolor{Top2}{81.25}} \\
    GAT       & 93.43 & 91.20 & 33.33 & 50.00 \\
    GIN       & \textbf{\textcolor{Top3}{97.33}} & 96.23 & \textbf{\textcolor{Top1}{87.30}} & \textbf{\textcolor{Top1}{87.50}} \\
    SAGE      & 96.71 & 95.21 & 33.33 & 50.00 \\
    \midrule
    \multicolumn{5}{l}{\textit{Heterogeneous}} \\
    R-GCN     & 97.15 & \textbf{\textcolor{Top3}{96.59}} & \textbf{\textcolor{Top3}{62.50}} & \textbf{\textcolor{Top3}{62.50}} \\
    HAN       & 93.80 & 94.03 & 33.33 & 50.00 \\
    HGT       & 96.31 & 95.77 & 33.33 & 50.00 \\
    \midrule
    Sheaf-NSD                & \textbf{\textcolor{Top2}{97.84}} & \textbf{\textcolor{Top2}{96.65}} & \textbf{\textcolor{Top1}{87.30}} & \textbf{\textcolor{Top1}{87.50}} \\
    \textbf{HetSheaf (ours)} & \textbf{\textcolor{Top1}{97.94}} & \textbf{\textcolor{Top1}{96.81}} & \textbf{\textcolor{Top1}{87.30}} & \textbf{\textcolor{Top1}{87.50}} \\
    \bottomrule
\end{tabular}}
}
\end{table}
\Cref{tab:networking_results} reports results on two networking benchmarks.
On AS-733, \textsc{HetSheaf} achieves the highest AUPR (97.94) and AUROC (96.81),
followed closely by Sheaf-NSD (97.84 / 96.65).
The three heterogeneous edge types (customer\textendash provider, peer\textendash to\textendash peer, provider\textendash customer)
encode semantically distinct peering relationships whose directionality and economic asymmetry provide a natural testbed for type-aware restriction maps; even a modest gain over Sheaf-NSD suggests that the heterogeneous predictor captures inter-type structure that a single shared learner averages away.
Among baselines, GIN and GCN are competitive on AUPR, while R-GCN attains the third-best AUROC, indicating that explicit relation typing helps on this task even without the sheaf structure.

On RouteNet Traffic Models, \textsc{HetSheaf} and GIN both reach the top Macro~F1 (87.30) and Micro~F1 (87.50), with GCN as the second-best model (80.57 / 81.25).
Notably, GAT, SAGE, HAN, and HGT all collapse to majority-class predictions (Macro~F1 of 33.33, Micro~F1 of 50.00), suggesting that the bipartite link\textendash path topology and the diversity of traffic generation patterns (CBR, ON-OFF, Autocorrelated, Modulated Exponentials) pose a difficult optimisation landscape for attention-based and certain heterogeneous architectures.
R-GCN avoids the degenerate solution but lags considerably behind GIN and the sheaf models, confirming that per-relation linear transforms alone are insufficient to capture traffic-dependent congestion dynamics.
The fact that \textsc{HetSheaf} matches GIN, the strongest homogeneous baseline, while additionally providing interpretable stalk-space representations, underscores its versatility across graph domains.

\subsubsection{Sheaf Predictor Ablation Study}
Tables~\ref{tab:nc-ablation}--\ref{tab:gc-ablation} report the full ablation grid over all heterogeneous sheaf predictors and restriction-map families for node classification, link prediction, and graph classification, respectively. These results isolate the contribution of explicit type conditioning from the contribution of the geometric family itself.

For node classification (\Cref{tab:nc-ablation}), heterogeneous predictors consistently improve over the vanilla NSD baseline on at least one dataset, confirming that typed conditioning of the restriction maps is beneficial in practice. On ACM and DBLP, the strongest configurations typically pair typed predictors with polynomial or general restriction-map families, indicating that type conditioning and higher-order diffusion are complementary. On IMDB, however, the best results are obtained by non-polynomial families, suggesting that on semantically noisier heterogeneous graphs, a simpler local transport geometry may generalise more robustly. Notably, predictors that use only type indicators without node features remain competitive, indicating that type information alone already carries substantial signal for heterogeneous node classification.

\begin{table}[!htbp]
    \centering
    \caption{
        \emph{Sheaf predictor $\times$ restriction map ablation, node classification.}
        \ablationlegend
    }
    \label{tab:nc-ablation}
\maxsizebox*{\textwidth}{0.86\textheight}
{   \resizebox{0.8\textwidth}{!}{
	\begin{tabular}{clSSSSSS}
		\toprule
		{\thead{Sheaf\\Predictor}} & {}            & \multicolumn{2}{c}{\thead{ACM}}            & \multicolumn{2}{c}{\thead{DBLP}}           & \multicolumn{2}{c}{\thead{IMDB}}           \\
		\cmidrule(lr){3-4}
		\cmidrule(lr){5-6}
		\cmidrule(lr){7-8}
		{}            & {}            & {\thead{Macro F1}} & {\thead{Micro F1}} & {\thead{Macro F1}} & {\thead{Micro F1}} & {\thead{Macro F1}} & {\thead{Micro F1}} \\
		\midrule
		\multirow{7}{*}{\rotatebox[origin=c]{90}{NSD}}
		& Diag           & 91.51 & 91.41 & 91.32 & 91.97 & \cellcolor{Top1!30!cyan!40}61.96 & \cellcolor{Top1!30!cyan!40}65.33 \\
		& Bundle         & 90.40 & 90.46 & \cellcolor{Top1!30!cyan!40}92.12 & \cellcolor{Top1!30!cyan!40}92.64 & 61.23 & 64.68 \\
		& General        & 89.81 & 89.75 & 89.33 & 89.96 & 61.77 & 64.85 \\
		& Diag-Poly      & \cellcolor{Top1!30!cyan!40}92.05 & \cellcolor{Top1!30!cyan!40}91.97 & 91.38 & 91.87 & 60.56 & 63.24 \\
		& Bundle-Poly    & 89.07 & 88.95 & 88.95 & 89.58 & 59.47 & 62.62 \\
		& Gen-Poly       & 86.74 & 86.64 & 91.33 & 91.76 & 59.62 & 62.62 \\
		& SheafAN        & 87.64 & 87.58 & 91.08 & 91.73 & 54.38 & 58.79 \\
		\midrule
		\multirow{7}{*}{\rotatebox[origin=c]{90}{TE}}
		& Diag           & 91.99 & 91.88 & 94.03 & 94.51 & \cellcolor{Top1!30!cyan!40}63.55 & 65.40 \\
		& Bundle         & 91.17 & 91.12 & 91.71 & 92.22 & 62.94 & \cellcolor{Top1!30!cyan!40}66.39 \\
		& General        & 91.63 & 91.50 & 93.13 & 93.66 & 61.62 & 64.58 \\
		& Diag-Poly      & \bfseries\cellcolor{Top1!30!cyan!40}94.22 & \bfseries\cellcolor{Top1!30!cyan!40}94.15 & \cellcolor{Top1!30!cyan!40}94.34 & \cellcolor{Top1!30!cyan!40}94.79 & 61.52 & 64.71 \\
		& Bundle-Poly    & 92.07 & 91.97 & 93.04 & 93.56 & 60.26 & 63.29 \\
		& Gen-Poly       & 88.57 & 88.39 & 94.08 & 94.47 & 60.59 & 63.71 \\
		& SheafAN        & 88.20 & 88.20 & 92.39 & 92.96 & 56.06 & 58.83 \\
		\midrule
		\multirow{7}{*}{\rotatebox[origin=c]{90}{Ensemble}}
		& Diag           & 92.58 & 92.45 & 93.38 & 93.84 & \cellcolor{Top1!30!cyan!40}63.22 & \cellcolor{Top1!30!cyan!40}65.83 \\
		& Bundle         & 90.29 & 90.23 & 93.11 & 93.56 & 62.62 & 65.47 \\
		& General        & 91.39 & 91.31 & 93.93 & 94.40 & 61.45 & 64.48 \\
		& Diag-Poly      & \cellcolor{Top1!30!cyan!40}93.36 & \cellcolor{Top1!30!cyan!40}93.25 & \cellcolor{Top1!30!cyan!40}94.26 & \cellcolor{Top1!30!cyan!40}94.65 & 60.34 & 63.73 \\
		& Bundle-Poly    & 87.63 & 87.54 & 90.08 & 90.70 & 61.45 & 63.47 \\
		& Gen-Poly       & 90.97 & 90.89 & 94.16 & 94.65 & 62.50 & 65.27 \\
		& SheafAN        & 86.48 & 86.54 & 81.94 & 83.63 & 55.79 & 60.21 \\
		\midrule
		\multirow{7}{*}{\rotatebox[origin=c]{90}{NE}}
		& Diag           & 91.90 & 91.78 & 93.10 & 93.56 & \bfseries\cellcolor{Top1!30!cyan!40}63.65 & \bfseries\cellcolor{Top1!30!cyan!40}66.56 \\
		& Bundle         & 91.15 & 91.08 & 92.94 & 93.45 & 61.28 & 65.00 \\
		& General        & 91.74 & 91.69 & 93.68 & 94.08 & 62.26 & 65.06 \\
		& Diag-Poly      & \cellcolor{Top1!30!cyan!40}92.55 & \cellcolor{Top1!30!cyan!40}92.49 & 93.40 & 93.84 & 61.30 & 64.11 \\
		& Bundle-Poly    & 89.60 & 89.52 & 91.82 & 92.36 & 59.73 & 63.77 \\
		& Gen-Poly       & 91.29 & 91.17 & \cellcolor{Top1!30!cyan!40}94.34 & \cellcolor{Top1!30!cyan!40}94.75 & 60.22 & 62.97 \\
		& SheafAN        & 88.76 & 88.76 & 91.06 & 91.76 & 54.90 & 58.52 \\
		\midrule
		\multirow{7}{*}{\rotatebox[origin=c]{90}{EE}}
		& Diag           & 92.50 & 92.40 & 93.50 & 93.91 & \cellcolor{Top1!30!cyan!40}62.86 & \cellcolor{Top1!30!cyan!40}66.49 \\
		& Bundle         & 91.46 & 91.50 & 92.52 & 92.99 & 62.16 & 64.39 \\
		& General        & 91.90 & 91.78 & 93.77 & 94.23 & 61.86 & 64.56 \\
		& Diag-Poly      & \cellcolor{Top1!30!cyan!40}93.76 & \cellcolor{Top1!30!cyan!40}93.67 & \cellcolor{Top1!30!cyan!40}94.35 & \cellcolor{Top1!30!cyan!40}94.79 & 60.57 & 63.46 \\
		& Bundle-Poly    & 84.38 & 84.37 & 85.66 & 86.55 & 56.44 & 61.57 \\
		& Gen-Poly       & 88.31 & 88.15 & 94.11 & 94.58 & 60.94 & 64.33 \\
		& SheafAN        & 88.14 & 88.05 & 90.62 & 91.23 & 55.56 & 59.23 \\
		\midrule
		\multirow{7}{*}{\rotatebox[origin=c]{90}{NT}}
		& Diag           & \cellcolor{Top1!30!cyan!40}93.72 & \cellcolor{Top1!30!cyan!40}93.63 & 93.53 & 94.01 & \cellcolor{Top1!30!cyan!40}62.48 & \cellcolor{Top1!30!cyan!40}65.22 \\
		& Bundle         & 90.99 & 90.84 & 92.29 & 92.82 & 62.16 & 63.86 \\
		& General        & 91.66 & 91.55 & 93.50 & 93.98 & 60.95 & 63.99 \\
		& Diag-Poly      & 92.32 & 92.26 & \cellcolor{Top1!30!cyan!40}94.34 & \cellcolor{Top1!30!cyan!40}94.75 & 60.03 & 63.72 \\
		& Bundle-Poly    & 88.76 & 88.67 & 93.27 & 93.80 & 60.39 & 63.60 \\
		& Gen-Poly       & 93.58 & 93.48 & 94.06 & 94.47 & 61.18 & 64.20 \\
		& SheafAN        & 88.95 & 88.95 & 81.45 & 82.25 & 55.25 & 58.33 \\
		\midrule
		\multirow{7}{*}{\rotatebox[origin=c]{90}{ET}}
		& Diag           & 92.95 & 92.87 & 94.11 & 94.51 & 61.41 & 64.57 \\
		& Bundle         & 89.90 & 89.80 & 90.35 & 90.85 & 60.23 & 63.42 \\
		& General        & 92.08 & 91.97 & 93.95 & 94.37 & \cellcolor{Top1!30!cyan!40}62.99 & \cellcolor{Top1!30!cyan!40}65.78 \\
		& Diag-Poly      & \cellcolor{Top1!30!cyan!40}93.07 & \cellcolor{Top1!30!cyan!40}92.97 & \bfseries\cellcolor{Top1!30!cyan!40}94.91 & \bfseries\cellcolor{Top1!30!cyan!40}95.28 & 60.71 & 63.80 \\
		& Bundle-Poly    & 89.62 & 89.47 & 86.73 & 87.61 & 60.35 & 63.52 \\
		& Gen-Poly       & 91.11 & 91.08 & 94.78 & 95.18 & 59.69 & 63.32 \\
		& SheafAN        & 88.39 & 88.39 & 81.76 & 82.78 & 55.64 & 58.78 \\
		\bottomrule
	\end{tabular}}
}

\end{table}

For link prediction (\Cref{tab:lp_ablation}), the gap between predictors is generally smaller, consistent with the fact that several datasets are already close to saturation under AUROC/AUPR. Even so, typed predictors remain consistently competitive with or stronger than NSD. General and diagonal restriction maps perform particularly well on recommendation-style datasets, while the ensemble predictor occasionally benefits from edge-type-specific routing when the relation semantics are rich enough to justify separate predictors.

\begin{table}[!htbp]
    \centering
    \caption{
        \emph{Sheaf predictor $\times$ restriction map ablation, link prediction.}
        \ablationlegend
    }
    \label{tab:lp_ablation}
    \maxsizebox{\textwidth}{!}{
        \maxsizebox{\textwidth}{!}
{
	\resizebox{0.8\textwidth}{!}{\begin{tabular}{clSSSSSSSS}
		\toprule
		{\thead{Sheaf\\Predictor}} & {}
		& \multicolumn{2}{c}{\thead{LastFM}}
		& \multicolumn{2}{c}{\thead{MovieLens}}
		& \multicolumn{2}{c}{\thead{Amazon}}
		& \multicolumn{2}{c}{\thead{YouTube}} \\
		\cmidrule(lr){3-4}
		\cmidrule(lr){5-6}
		\cmidrule(lr){7-8}
		\cmidrule(lr){9-10}
		{} & {}
		& {\thead{AUPR}} & {\thead{AUROC}}
		& {\thead{AUPR}} & {\thead{AUROC}}
		& {\thead{AUPR}} & {\thead{AUROC}}
		& {\thead{AUPR}} & {\thead{AUROC}} \\
		\midrule
		\multirow{7}{*}{\rotatebox[origin=c]{90}{NSD}}
		& Diag          & \cellcolor{Top1!30!cyan!40}98.06 & \cellcolor{Top1!30!cyan!40}97.73 & \cellcolor{Top1!30!cyan!40}99.68 & 99.60 & 92.59 & \cellcolor{Top1!30!cyan!40}90.03 & 89.06 & 82.85 \\
		& Bundle          & 97.69 & 97.35 & 99.66 & 99.60 & 92.58 & 89.98 & 89.30 & 83.19 \\
		& General          & 97.98 & 97.68 & \cellcolor{Top1!30!cyan!40}99.68 & \bfseries\cellcolor{Top1!30!cyan!40}99.62 & 92.48 & 89.95 & 88.98 & 82.73 \\
		& Diag-Poly          & 97.34 & 96.56 & 99.65 & 99.56 & 92.60 & 89.79 & \cellcolor{Top1!30!cyan!40}89.45 & \cellcolor{Top1!30!cyan!40}83.32 \\
		& Bundle-Poly          & 97.49 & 97.04 & 99.67 & 99.58 & \cellcolor{Top1!30!cyan!40}92.69 & 90.00 & 89.37 & 83.23 \\
		& Gen-Poly          & 97.62 & 97.22 & 99.63 & 99.57 & 92.25 & 89.37 & {--} & {--} \\
		& SheafAN          & 93.05 & 91.52 & 99.48 & 99.37 & 91.30 & 87.95 & 82.45 & 76.24 \\
		\midrule
		\multirow{7}{*}{\rotatebox[origin=c]{90}{TE}}
		& Diag          & 98.39 & 98.24 & \bfseries\cellcolor{Top1!30!cyan!40}99.69 & \cellcolor{Top1!30!cyan!40}99.60 & \cellcolor{Top1!30!cyan!40}93.85 & \cellcolor{Top1!30!cyan!40}91.59 & \cellcolor{Top1!30!cyan!40}89.46 & \cellcolor{Top1!30!cyan!40}83.38 \\
		& Bundle          & \cellcolor{Top1!30!cyan!40}98.50 & 98.20 & 99.65 & 99.58 & 93.51 & 91.22 & 89.01 & 82.78 \\
		& General          & 98.45 & \cellcolor{Top1!30!cyan!40}98.30 & 99.67 & \cellcolor{Top1!30!cyan!40}99.60 & 93.71 & 91.55 & 88.87 & 82.58 \\
		& Diag-Poly          & 98.40 & 98.02 & 99.66 & 99.57 & 93.33 & 90.71 & 89.39 & 83.36 \\
		& Bundle-Poly          & 98.35 & 98.01 & 99.68 & \cellcolor{Top1!30!cyan!40}99.60 & 93.27 & 90.77 & 89.35 & 83.24 \\
		& Gen-Poly          & 98.30 & 98.07 & 99.65 & 99.59 & 93.49 & 91.28 & 89.21 & 82.95 \\
		& SheafAN          & 97.71 & 97.26 & 99.51 & 99.42 & 91.40 & 87.62 & 86.44 & 80.23 \\
		\midrule
		\multirow{7}{*}{\rotatebox[origin=c]{90}{Ensemble}}
		& Diag          & 98.47 & 98.16 & \cellcolor{Top1!30!cyan!40}99.68 & 99.60 & \cellcolor{Top1!30!cyan!40}93.87 & 91.52 & 90.53 & 85.61 \\
		& Bundle          & 97.98 & 97.72 & 99.66 & \cellcolor{Top1!30!cyan!40}99.61 & 93.45 & 91.10 & 88.79 & 82.45 \\
		& General          & \bfseries\cellcolor{Top1!30!cyan!40}98.54 & \bfseries\cellcolor{Top1!30!cyan!40}98.33 & 99.67 & 99.60 & 93.76 & \cellcolor{Top1!30!cyan!40}91.58 & \cellcolor{Top1!30!cyan!40}90.65 & \cellcolor{Top1!30!cyan!40}85.95 \\
		& Diag-Poly          & 98.08 & 97.51 & 99.61 & 99.56 & 93.46 & 90.88 & 89.49 & 83.46 \\
		& Bundle-Poly          & 97.49 & 96.95 & 99.67 & 99.59 & 93.40 & 90.95 & 89.39 & 83.30 \\
		& Gen-Poly          & 98.39 & 98.07 & 99.66 & \cellcolor{Top1!30!cyan!40}99.61 & 93.55 & 91.23 & {--} & {--} \\
		& SheafAN          & 96.43 & 95.39 & 99.48 & 99.41 & 91.99 & 88.52 & 87.10 & 80.61 \\
		\midrule
		\multirow{7}{*}{\rotatebox[origin=c]{90}{NE}}
		& Diag          & 98.43 & 98.24 & \bfseries\cellcolor{Top1!30!cyan!40}99.69 & \cellcolor{Top1!30!cyan!40}99.61 & 92.59 & 90.04 & 89.27 & \cellcolor{Top1!30!cyan!40}83.36 \\
		& Bundle          & 98.42 & 98.16 & 99.67 & 99.60 & 92.49 & 89.88 & 89.33 & 83.23 \\
		& General          & \cellcolor{Top1!30!cyan!40}98.46 & \cellcolor{Top1!30!cyan!40}98.25 & 99.67 & \cellcolor{Top1!30!cyan!40}99.61 & \cellcolor{Top1!30!cyan!40}92.64 & \cellcolor{Top1!30!cyan!40}90.08 & 89.25 & 83.10 \\
		& Diag-Poly          & 98.42 & 98.05 & 99.66 & 99.57 & 92.56 & 89.79 & \cellcolor{Top1!30!cyan!40}89.41 & 83.28 \\
		& Bundle-Poly          & 98.22 & 97.87 & 99.67 & 99.58 & 92.53 & 89.90 & 89.38 & 83.21 \\
		& Gen-Poly          & 98.30 & 98.02 & 99.66 & 99.59 & 92.48 & 89.50 & {--} & {--} \\
		& SheafAN          & 97.81 & 97.56 & 99.47 & 99.36 & 91.46 & 88.21 & 82.68 & 76.62 \\
		\midrule
		\multirow{7}{*}{\rotatebox[origin=c]{90}{EE}}
		& Diag          & 98.41 & \cellcolor{Top1!30!cyan!40}98.16 & \cellcolor{Top1!30!cyan!40}99.68 & 99.60 & \bfseries\cellcolor{Top1!30!cyan!40}93.89 & \cellcolor{Top1!30!cyan!40}91.65 & 89.95 & 84.06 \\
		& Bundle          & 97.62 & 97.28 & 99.66 & \cellcolor{Top1!30!cyan!40}99.61 & 93.62 & 91.27 & 88.86 & 82.65 \\
		& General          & \cellcolor{Top1!30!cyan!40}98.51 & 98.13 & 99.67 & \cellcolor{Top1!30!cyan!40}99.61 & 93.71 & 91.51 & \bfseries\cellcolor{Top1!30!cyan!40}91.05 & \bfseries\cellcolor{Top1!30!cyan!40}86.76 \\
		& Diag-Poly          & 97.87 & 97.34 & 99.62 & 99.53 & 93.29 & 90.53 & 89.87 & 84.02 \\
		& Bundle-Poly          & 97.74 & 97.19 & 99.67 & 99.58 & 93.12 & 90.63 & 89.40 & 83.31 \\
		& Gen-Poly          & 98.17 & 97.77 & 99.66 & 99.59 & 93.60 & 91.30 & {--} & {--} \\
		& SheafAN          & 86.12 & 79.22 & 99.46 & 99.40 & 91.27 & 87.48 & 86.11 & 79.63 \\
		\midrule
		\multirow{7}{*}{\rotatebox[origin=c]{90}{NT}}
		& Diag          & \cellcolor{Top1!30!cyan!40}98.47 & 98.25 & \cellcolor{Top1!30!cyan!40}99.68 & \cellcolor{Top1!30!cyan!40}99.61 & 92.59 & \cellcolor{Top1!30!cyan!40}90.13 & 89.31 & 83.26 \\
		& Bundle          & 98.45 & 98.24 & 99.66 & \cellcolor{Top1!30!cyan!40}99.61 & 92.37 & 89.75 & 89.23 & 83.17 \\
		& General          & 98.41 & \cellcolor{Top1!30!cyan!40}98.29 & 99.67 & \cellcolor{Top1!30!cyan!40}99.61 & 92.51 & 89.95 & 89.34 & 83.24 \\
		& Diag-Poly          & 98.28 & 98.08 & 99.66 & 99.57 & \cellcolor{Top1!30!cyan!40}92.61 & 89.82 & \cellcolor{Top1!30!cyan!40}89.41 & 83.26 \\
		& Bundle-Poly          & 98.22 & 97.94 & \cellcolor{Top1!30!cyan!40}99.68 & 99.59 & 92.60 & 89.89 & 89.40 & \cellcolor{Top1!30!cyan!40}83.29 \\
		& Gen-Poly          & 98.32 & 98.08 & 99.66 & 99.58 & 92.50 & 89.98 & {--} & {--} \\
		& SheafAN          & 97.74 & 97.31 & 99.50 & 99.41 & 91.26 & 87.53 & 82.54 & 76.39 \\
		\midrule
		\multirow{7}{*}{\rotatebox[origin=c]{90}{ET}}
		& Diag          & \cellcolor{Top1!30!cyan!40}98.43 & 98.15 & \bfseries\cellcolor{Top1!30!cyan!40}99.69 & \cellcolor{Top1!30!cyan!40}99.61 & \bfseries\cellcolor{Top1!30!cyan!40}93.89 & 91.56 & \cellcolor{Top1!30!cyan!40}90.74 & \cellcolor{Top1!30!cyan!40}86.03 \\
		& Bundle          & 97.56 & 97.18 & 99.67 & \cellcolor{Top1!30!cyan!40}99.61 & 92.31 & 89.85 & 89.26 & 83.11 \\
		& General          & \cellcolor{Top1!30!cyan!40}98.43 & \cellcolor{Top1!30!cyan!40}98.16 & 99.67 & 99.60 & 93.81 & \bfseries\cellcolor{Top1!30!cyan!40}91.71 & 90.62 & 85.95 \\
		& Diag-Poly          & 97.97 & 97.40 & 99.63 & 99.55 & 93.18 & 90.49 & 90.62 & 85.82 \\
		& Bundle-Poly          & 97.69 & 97.20 & 99.65 & 99.58 & 92.71 & 89.95 & 89.28 & 83.11 \\
		& Gen-Poly          & 98.03 & 97.72 & 99.67 & 99.58 & 93.77 & 91.58 & {--} & {--} \\
		& SheafAN          & 95.87 & 95.66 & 99.56 & 99.49 & 90.89 & 86.58 & 86.29 & 79.95 \\
		\bottomrule
	\end{tabular}}
}
    }
\end{table}

For graph classification (\Cref{tab:gc-ablation}), predictor choice has a much stronger effect on performance, especially on PROTEINS, ENZYMES, and NCI1. Here, type-only predictors such as NT and ET often become surprisingly effective. This is consistent with the inductive nature of graph classification: when the model must generalise across unseen graphs, structural type information can transfer more robustly than instance-specific node features. Overall, the graph-classification ablations support the view that typed restriction-map inference is particularly valuable when the model must compress heterogeneous local geometry into a stable graph-level representation.

\begin{table}[!htbp]
    \centering
    \caption{
        \emph{Sheaf predictor $\times$ restriction map ablation, graph classification.}
        \ablationlegend
    }
    \label{tab:gc-ablation}
    \maxsizebox{\textwidth}{!}
{
	\resizebox{0.8\textwidth}{!}{\begin{tabular}{clSSSSSSSS}
		\toprule
		{\thead{Sheaf\\Predictor}} & {}
		& \multicolumn{2}{c}{\thead{MUTAG}}
		& \multicolumn{2}{c}{\thead{PROTEINS}}
		& \multicolumn{2}{c}{\thead{ENZYMES}}
		& \multicolumn{2}{c}{\thead{NCI1}} \\
		\cmidrule(lr){3-4}
		\cmidrule(lr){5-6}
		\cmidrule(lr){7-8}
		\cmidrule(lr){9-10}
		{} & {}
		& {\thead{Macro F1}} & {\thead{Micro F1}}
		& {\thead{Macro F1}} & {\thead{Micro F1}}
		& {\thead{Macro F1}} & {\thead{Micro F1}}
		& {\thead{Macro F1}} & {\thead{Micro F1}} \\
		\midrule
		\multirow{7}{*}{\rotatebox[origin=c]{90}{NSD}}
		& Diag           & 75.00 & 75.00 & 62.20 & 62.50 & 18.82 & 28.33 & 55.11 & 55.11 \\
		& Bundle         & 68.75 & 70.00 & \cellcolor{Top1!30!cyan!40}73.60 & 75.89 & 9.56 & 23.33 & 38.69 & 38.69 \\
		& General        & 76.19 & 80.00 & 26.67 & 33.04 & 8.75 & 23.33 & {--} & {--} \\
		& Diag-Poly      & \cellcolor{Top1!30!cyan!40}96.00 & \cellcolor{Top1!30!cyan!40}96.00 & 51.77 & 51.79 & 8.38 & 21.67 & 45.73 & 45.73 \\
		& Bundle-Poly    & 94.88 & 95.00 & 71.56 & \cellcolor{Top1!30!cyan!40}76.79 & \cellcolor{Top1!30!cyan!40}26.67 & \cellcolor{Top1!30!cyan!40}30.00 & 63.24 & 63.24 \\
		& Gen-Poly       & 89.01 & 90.00 & 66.21 & 71.43 & 5.71 & 20.00 & \cellcolor{Top1!30!cyan!40}72.00 & \cellcolor{Top1!30!cyan!40}72.00 \\
		& SheafAN        & 89.90 & 90.00 & 24.32 & 32.14 & 9.48 & 20.00 & 62.08 & 62.08 \\
		\midrule
		\multirow{7}{*}{\rotatebox[origin=c]{90}{TE}}
		& Diag           & 75.00 & 75.00 & 49.98 & 50.00 & 12.70 & 21.67 & 61.85 & 61.85 \\
		& Bundle         & 69.70 & 70.00 & \cellcolor{Top1!30!cyan!40}64.54 & 65.18 & 12.66 & 21.67 & 62.08 & 62.08 \\
		& General        & 37.50 & 60.00 & 27.22 & 33.93 & 7.31 & 18.33 & 62.08 & 62.08 \\
		& Diag-Poly      & 95.65 & 95.65 & 33.60 & 35.71 & 13.20 & 23.33 & 61.36 & 61.36 \\
		& Bundle-Poly    & \bfseries\cellcolor{Top1!30!cyan!40}100.00 & \bfseries\cellcolor{Top1!30!cyan!40}100.00 & 62.78 & \cellcolor{Top1!30!cyan!40}71.43 & 18.54 & 23.33 & 68.62 & 68.62 \\
		& Gen-Poly       & 100.00 & 100.00 & 55.36 & 55.36 & \cellcolor{Top1!30!cyan!40}30.75 & \cellcolor{Top1!30!cyan!40}33.33 & \cellcolor{Top1!30!cyan!40}72.54 & \cellcolor{Top1!30!cyan!40}72.54 \\
		& SheafAN        & 84.00 & 85.00 & 24.32 & 32.14 & 16.98 & 21.67 & 69.70 & 69.70 \\
		\midrule
		\multirow{7}{*}{\rotatebox[origin=c]{90}{Ensemble}}
		& Diag           & 78.57 & 78.57 & \cellcolor{Top1!30!cyan!40}73.60 & \cellcolor{Top1!30!cyan!40}75.89 & 11.85 & 20.00 & 29.03 & 29.03 \\
		& Bundle         & 69.70 & 70.00 & 58.28 & 58.93 & 15.92 & 25.00 & 21.43 & 21.43 \\
		& General        & 67.03 & 70.00 & 53.20 & 53.57 & 5.56 & 20.00 & \cellcolor{Top1!30!cyan!40}64.42 & \cellcolor{Top1!30!cyan!40}64.42 \\
		& Diag-Poly      & 95.65 & 95.65 & 41.27 & 67.86 & 8.20 & 21.67 & 62.08 & 62.08 \\
		& Bundle-Poly    & \cellcolor{Top1!30!cyan!40}100.00 & \cellcolor{Top1!30!cyan!40}100.00 & 69.22 & 71.43 & 24.52 & 25.00 & 51.18 & 51.18 \\
		& Gen-Poly       & 100.00 & 100.00 & 58.92 & 58.93 & \cellcolor{Top1!30!cyan!40}38.05 & \cellcolor{Top1!30!cyan!40}41.67 & 54.98 & 54.98 \\
		& SheafAN        & 37.50 & 60.00 & 24.32 & 32.14 & 15.22 & 23.33 & 62.08 & 62.08 \\
		\midrule
		\multirow{7}{*}{\rotatebox[origin=c]{90}{NE}}
		& Diag           & 75.00 & 75.00 & 68.93 & 72.32 & 16.10 & 23.33 & 9.95 & 9.95 \\
		& Bundle         & 78.02 & 80.00 & 71.96 & 74.11 & 6.38 & 21.67 & 23.11 & 23.11 \\
		& General        & 64.29 & 70.00 & 28.90 & 33.04 & 9.96 & 21.67 & 54.15 & 54.15 \\
		& Diag-Poly      & \cellcolor{Top1!30!cyan!40}100.00 & \cellcolor{Top1!30!cyan!40}100.00 & 25.79 & 33.04 & 16.42 & 28.33 & 60.20 & 60.20 \\
		& Bundle-Poly    & 100.00 & 100.00 & 72.97 & 74.11 & 19.31 & 26.67 & \cellcolor{Top1!30!cyan!40}69.52 & \cellcolor{Top1!30!cyan!40}69.52 \\
		& Gen-Poly       & 100.00 & 100.00 & 63.39 & 66.96 & 13.66 & 20.00 & 41.41 & 41.41 \\
		& SheafAN        & 89.01 & 90.00 & \cellcolor{Top1!30!cyan!40}75.47 & \cellcolor{Top1!30!cyan!40}80.36 & \cellcolor{Top1!30!cyan!40}35.53 & \cellcolor{Top1!30!cyan!40}38.33 & 24.63 & 24.63 \\
		\midrule
		\multirow{7}{*}{\rotatebox[origin=c]{90}{EE}}
		& Diag           & 76.19 & 80.00 & 63.87 & 64.29 & 20.03 & 28.33 & 46.99 & 46.99 \\
		& Bundle         & 37.50 & 60.00 & 25.79 & 33.04 & 5.56 & 20.00 & 18.35 & 18.35 \\
		& General        & 79.80 & 80.00 & 38.00 & 40.18 & 6.02 & 21.67 & 59.94 & 59.94 \\
		& Diag-Poly      & \cellcolor{Top1!30!cyan!40}96.00 & \cellcolor{Top1!30!cyan!40}96.00 & 38.24 & 41.07 & 13.62 & 21.67 & 52.10 & 52.10 \\
		& Bundle-Poly    & 94.88 & 95.00 & 54.29 & 60.71 & \cellcolor{Top1!30!cyan!40}32.82 & \cellcolor{Top1!30!cyan!40}31.67 & 63.89 & 63.89 \\
		& Gen-Poly       & 94.88 & 95.00 & 27.22 & 33.93 & 10.47 & 20.00 & \cellcolor{Top1!30!cyan!40}66.09 & \cellcolor{Top1!30!cyan!40}66.09 \\
		& SheafAN        & 94.88 & 95.00 & \cellcolor{Top1!30!cyan!40}66.16 & \cellcolor{Top1!30!cyan!40}68.75 & 17.13 & 20.00 & 62.08 & 62.08 \\
		\midrule
		\multirow{7}{*}{\rotatebox[origin=c]{90}{NT}}
		& Diag           & 80.00 & 80.00 & 73.60 & 75.89 & \bfseries\cellcolor{Top1!30!cyan!40}52.96 & \bfseries\cellcolor{Top1!30!cyan!40}53.33 & \bfseries\cellcolor{Top1!30!cyan!40}78.24 & \bfseries\cellcolor{Top1!30!cyan!40}78.24 \\
		& Bundle         & 79.80 & 80.00 & 69.84 & 70.54 & 50.80 & 50.00 & 62.64 & 62.64 \\
		& General        & 74.94 & 75.00 & 63.25 & 63.39 & 49.92 & 51.67 & 70.85 & 70.85 \\ 
		& Diag-Poly      & \cellcolor{Top1!30!cyan!40}94.88 & \cellcolor{Top1!30!cyan!40}95.00 & 72.51 & 74.11 & 22.67 & 26.67 & 23.01 & 23.01 \\
		& Bundle-Poly    & 94.88 & 95.00 & \bfseries\cellcolor{Top1!30!cyan!40}78.84 & 80.36 & 50.01 & 50.00 & 74.74 & 74.74 \\
		& Gen-Poly       & 89.90 & 90.00 & 78.66 & \bfseries\cellcolor{Top1!30!cyan!40}81.25 & 48.13 & 48.33 & 72.86 & 72.86 \\
		& SheafAN        & 89.90 & 90.00 & 76.95 & 79.46 & 46.58 & 48.33 & 75.06 & 75.06 \\
		\midrule
		\multirow{7}{*}{\rotatebox[origin=c]{90}{ET}}
		& Diag           & 76.19 & 80.00 & 48.65 & 67.86 & 12.73 & \cellcolor{Top1!30!cyan!40}26.67 & 57.60 & 57.60 \\
		& Bundle         & 74.42 & 75.00 & 54.37 & 54.46 & 10.39 & 25.00 & 44.92 & 44.92 \\
		& General        & 89.90 & 90.00 & 24.32 & 32.14 & 8.80 & 21.67 & 59.20 & 59.20 \\
		& Diag-Poly      & 94.88 & 95.00 & 61.35 & 68.75 & 9.15 & 21.67 & 62.08 & 62.08 \\
		& Bundle-Poly    & 94.88 & 95.00 & \cellcolor{Top1!30!cyan!40}65.06 & \cellcolor{Top1!30!cyan!40}74.11 & 15.57 & 25.00 & 44.88 & 44.88 \\
		& Gen-Poly       & \cellcolor{Top1!30!cyan!40}100.00 & \cellcolor{Top1!30!cyan!40}100.00 & 24.32 & 32.14 & \cellcolor{Top1!30!cyan!40}17.86 & 23.33 & \cellcolor{Top1!30!cyan!40}72.87 & \cellcolor{Top1!30!cyan!40}72.87 \\
		& SheafAN        & 58.33 & 60.00 & 61.23 & 61.61 & 17.86 & 20.00 & 62.08 & 62.08 \\
		\bottomrule
	\end{tabular}}
}
\end{table}

\subsubsection{Effect of Restriction Map Type}
The predictor ablations also reveal a second important axis: the choice of restriction-map family. Across tasks, no single family dominates. Polynomial families are often strongest on node classification and on several graph classification benchmarks, where longer-range typed interactions are beneficial. By contrast, non-polynomial diagonal or bundle families can be preferable on more semantically heterogeneous or noisier datasets such as IMDB and ENZYMES. This task dependence is itself informative: it suggests that the optimal local transport geometry depends on whether the problem rewards smooth long-range propagation, local typed alignment, or robust inductive generalisation.

For node classification, Diag-Poly and Gen.-Poly are often strongest on ACM and DBLP, whereas non-polynomial families dominate on IMDB. For link prediction, the restriction-map family tends to have a smaller effect, with diagonal maps performing consistently well and general maps providing only modest gains in some settings. For graph classification, polynomial families are essential on some datasets but can also become unstable on others, especially when small graph size and structural diversity interact poorly with higher-order spectral filtering. This reinforces the importance of treating the restriction-map family as a genuine modelling choice rather than as a fixed default.

\subsubsection{Attention-based Restriction Map Variants}
Tables~\ref{tab:nc-ablation-att}--\ref{tab:gc-ablation-att} repeat the main ablation study using attention-based restriction-map constructions. These experiments test whether replacing the standard predictor output by a normalised attention-style map provides a useful inductive bias.

For node classification (\Cref{tab:nc-ablation-att}), attention-based variants broaden the set of competitive configurations and occasionally outperform their non-attention counterparts, especially when paired with bundle or general maps. For link prediction (\Cref{tab:lp-ablation-att}), attention predictors are competitive but do not provide a uniform improvement, suggesting that the advantage of additional normalisation depends strongly on the relation structure of the dataset. For graph classification (\Cref{tab:gc-ablation-att}), the picture is mixed: attention expands the space of viable configurations, but does not yield a systematic gain over the standard construction. 

\begin{table}[!htbp]
    \centering
    \caption{
        \emph{Sheaf \emph{attention} predictor $\times$ restriction map ablation, node classification.}
        \ablationlegend
    }
    \label{tab:nc-ablation-att}
    \maxsizebox{\textwidth}{!}
{
	\renewcommand{\arraystretch}{0.85}
	\setlength{\tabcolsep}{4pt}
	\resizebox{0.8\textwidth}{!}{\begin{tabular}{clSSSSSS}
		\toprule
		{\thead{Sheaf\\Predictor}} & {}            & \multicolumn{2}{c}{\thead{ACM}}            & \multicolumn{2}{c}{\thead{DBLP}}           & \multicolumn{2}{c}{\thead{IMDB}}           \\
		\cmidrule(lr){3-4}
		\cmidrule(lr){5-6}
		\cmidrule(lr){7-8}
		{}            & {}            & {\thead{Macro F1}} & {\thead{Micro F1}} & {\thead{Macro F1}} & {\thead{Micro F1}} & {\thead{Macro F1}} & {\thead{Micro F1}} \\
		\midrule
		\multirow{5}{*}{\rotatebox[origin=c]{90}{Att.\,NSD}}
		& Bundle         & 89.49 & 89.38 & 92.10 & 92.57 & \cellcolor{Top1!30!cyan!40}61.88 & 64.49 \\
		& General        & 90.35 & 90.32 & 90.94 & 91.58 & 61.71 & \cellcolor{Top1!30!cyan!40}64.51 \\
		& Bundle-Poly    & 88.24 & 88.20 & 90.51 & 91.13 & 61.29 & 64.04 \\
		& Gen-Poly       & \cellcolor{Top1!30!cyan!40}90.93 & \cellcolor{Top1!30!cyan!40}90.89 & \cellcolor{Top1!30!cyan!40}92.96 & \cellcolor{Top1!30!cyan!40}93.42 & 59.88 & 63.40 \\
		& SheafAN        & 88.67 & 88.62 & 87.77 & 88.77 & 55.64 & 59.05 \\
		\midrule
		\multirow{5}{*}{\rotatebox[origin=c]{90}{Att.\,TE}}
		& Bundle         & 90.22 & 90.18 & 93.21 & 93.70 & \cellcolor{Top1!30!cyan!40}63.88 & \cellcolor{Top1!30!cyan!40}66.34 \\
		& General        & \cellcolor{Top1!30!cyan!40}91.76 & \cellcolor{Top1!30!cyan!40}91.64 & 93.13 & 93.52 & 62.16 & 65.55 \\
		& Bundle-Poly    & 88.11 & 87.91 & \cellcolor{Top1!30!cyan!40}94.56 & \cellcolor{Top1!30!cyan!40}95.00 & 61.00 & 64.28 \\
		& Gen-Poly       & 90.18 & 90.09 & 92.79 & 93.27 & 62.39 & 65.81 \\
		& SheafAN        & 88.01 & 87.96 & 85.70 & 86.65 & 55.25 & 59.35 \\
		\midrule
		\multirow{5}{*}{\rotatebox[origin=c]{90}{Att.\,Ens.}}
		& Bundle         & \bfseries\cellcolor{Top1!30!cyan!40}92.57 & \bfseries\cellcolor{Top1!30!cyan!40}92.54 & 93.46 & 93.94 & \bfseries\cellcolor{Top1!30!cyan!40}64.54 & \bfseries\cellcolor{Top1!30!cyan!40}67.18 \\
		& General        & 90.75 & 90.65 & \cellcolor{Top1!30!cyan!40}93.96 & \cellcolor{Top1!30!cyan!40}94.40 & 62.38 & 65.37 \\
		& Bundle-Poly    & 89.28 & 89.14 & 93.28 & 93.77 & 59.15 & 64.31 \\
		& Gen-Poly       & 88.65 & 88.62 & 93.91 & 94.33 & 58.87 & 62.83 \\
		& SheafAN        & 89.72 & 89.71 & 81.28 & 81.94 & 56.23 & 59.23 \\
		\midrule
		\multirow{5}{*}{\rotatebox[origin=c]{90}{Att.\,NE}}
		& Bundle         & 90.76 & 90.70 & 93.66 & 94.12 & \cellcolor{Top1!30!cyan!40}63.03 & \cellcolor{Top1!30!cyan!40}66.47 \\
		& General        & \cellcolor{Top1!30!cyan!40}91.09 & \cellcolor{Top1!30!cyan!40}90.98 & 93.11 & 93.59 & 60.32 & 63.51 \\
		& Bundle-Poly    & 89.67 & 89.52 & 92.10 & 92.71 & 60.97 & 63.72 \\
		& Gen-Poly       & 88.98 & 88.95 & \cellcolor{Top1!30!cyan!40}94.61 & \cellcolor{Top1!30!cyan!40}95.04 & 58.30 & 61.27 \\
		& SheafAN        & 87.15 & 87.02 & 81.32 & 81.94 & 55.25 & 58.70 \\
		\midrule
		\multirow{5}{*}{\rotatebox[origin=c]{90}{Att.\,EE}}
		& Bundle         & \cellcolor{Top1!30!cyan!40}91.79 & \cellcolor{Top1!30!cyan!40}91.69 & 93.32 & 93.77 & \cellcolor{Top1!30!cyan!40}63.50 & \cellcolor{Top1!30!cyan!40}66.44 \\
		& General        & 90.59 & 90.51 & \cellcolor{Top1!30!cyan!40}93.95 & \cellcolor{Top1!30!cyan!40}94.44 & 60.97 & 64.70 \\
		& Bundle-Poly    & 86.21 & 86.26 & 92.90 & 93.59 & 60.00 & 62.22 \\
		& Gen-Poly       & 91.01 & 90.93 & 92.70 & 93.24 & 60.53 & 63.10 \\
		& SheafAN        & 85.59 & 85.41 & 91.13 & 91.87 & 54.78 & 58.32 \\
		\midrule
		\multirow{5}{*}{\rotatebox[origin=c]{90}{Att.\,NT}}
		& Bundle         & 90.78 & 90.70 & 93.73 & 94.23 & \cellcolor{Top1!30!cyan!40}63.00 & \cellcolor{Top1!30!cyan!40}66.37 \\
		& General        & 90.95 & 90.84 & 93.52 & 93.98 & 60.77 & 63.84 \\
		& Bundle-Poly    & 89.72 & 89.61 & 92.89 & 93.35 & 60.01 & 62.54 \\
		& Gen-Poly       & \cellcolor{Top1!30!cyan!40}91.52 & \cellcolor{Top1!30!cyan!40}91.41 & \cellcolor{Top1!30!cyan!40}94.17 & \cellcolor{Top1!30!cyan!40}94.58 & 59.34 & 63.35 \\
		& SheafAN        & 88.16 & 88.15 & 80.82 & 81.44 & 55.76 & 58.38 \\
		\midrule
		\multirow{5}{*}{\rotatebox[origin=c]{90}{Att.\,ET}}
		& Bundle         & \cellcolor{Top1!30!cyan!40}91.76 & \cellcolor{Top1!30!cyan!40}91.69 & 93.81 & 94.26 & \cellcolor{Top1!30!cyan!40}62.57 & \cellcolor{Top1!30!cyan!40}65.97 \\
		& General        & 91.14 & 91.08 & 94.67 & 95.07 & 62.02 & 64.87 \\
		& Bundle-Poly    & 90.30 & 90.23 & \bfseries\cellcolor{Top1!30!cyan!40}94.67 & \bfseries\cellcolor{Top1!30!cyan!40}95.11 & 58.11 & 61.80 \\
		& Gen-Poly       & 88.28 & 88.20 & 93.85 & 94.30 & 62.29 & 65.11 \\
		& SheafAN        & 85.49 & 85.41 & 83.86 & 84.61 & 56.36 & 59.18 \\
		\bottomrule
	\end{tabular}}
}
\end{table}

\begin{table}[!htbp]
    \centering
     \caption{
         \emph{Sheaf \emph{attention} predictor $\times$ restriction map ablation, link prediction.}
         \ablationlegend
     }
    \label{tab:lp-ablation-att}
    \maxsizebox{\textwidth}{!}
{
	\renewcommand{\arraystretch}{0.85}
	\setlength{\tabcolsep}{3pt}
	\resizebox{0.8\textwidth}{!}{\begin{tabular}{clSSSSSSSS}
		\toprule
		{\thead{Sheaf\\Predictor}} & {}
		& \multicolumn{2}{c}{\thead{LastFM}}
		& \multicolumn{2}{c}{\thead{MovieLens}}
		& \multicolumn{2}{c}{\thead{Amazon}}
		& \multicolumn{2}{c}{\thead{YouTube}} \\
		\cmidrule(lr){3-4}
		\cmidrule(lr){5-6}
		\cmidrule(lr){7-8}
		\cmidrule(lr){9-10}
		{} & {}
		& {\thead{AUPR}} & {\thead{AUROC}}
		& {\thead{AUPR}} & {\thead{AUROC}}
		& {\thead{AUPR}} & {\thead{AUROC}}
		& {\thead{AUPR}} & {\thead{AUROC}} \\
		\midrule
		\multirow{5}{*}{\rotatebox[origin=c]{90}{Att.\,NSD}}
		& Bundle         & \cellcolor{Top1!30!cyan!40}97.98 & \cellcolor{Top1!30!cyan!40}97.71 & \cellcolor{Top1!30!cyan!40}99.66 & \cellcolor{Top1!30!cyan!40}99.60 & 92.53 & 90.00 & \cellcolor{Top1!30!cyan!40}89.32 & \cellcolor{Top1!30!cyan!40}83.36 \\
		& General        & 97.85 & 97.52 & \cellcolor{Top1!30!cyan!40}99.66 & \cellcolor{Top1!30!cyan!40}99.60 & 92.46 & 89.79 & 89.22 & 83.09 \\
		& Bundle-Poly    & 97.56 & 97.20 & \cellcolor{Top1!30!cyan!40}99.66 & 99.58 & \cellcolor{Top1!30!cyan!40}92.63 & \cellcolor{Top1!30!cyan!40}90.04 & 89.31 & 83.10 \\
		& Gen-Poly       & 97.86 & 97.37 & \cellcolor{Top1!30!cyan!40}99.66 & 99.59 & 92.54 & 89.81 & {--} & {--} \\
		& SheafAN        & 94.25 & 92.73 & 99.47 & 99.35 & 92.03 & 89.22 & 82.70 & 76.66 \\
		\midrule
		\multirow{5}{*}{\rotatebox[origin=c]{90}{Att.\,TE}}
		& Bundle         & \cellcolor{Top1!30!cyan!40}98.46 & 98.20 & 99.65 & 99.59 & 93.85 & \cellcolor{Top1!30!cyan!40}91.59 & 89.17 & 82.99 \\
		& General        & 98.33 & \cellcolor{Top1!30!cyan!40}98.24 & 99.66 & \bfseries\cellcolor{Top1!30!cyan!40}99.61 & \cellcolor{Top1!30!cyan!40}93.87 & \cellcolor{Top1!30!cyan!40}91.59 & 89.24 & 83.10 \\
		& Bundle-Poly    & 98.16 & 97.94 & \bfseries\cellcolor{Top1!30!cyan!40}99.68 & 99.60 & 93.54 & 91.12 & \cellcolor{Top1!30!cyan!40}89.36 & \cellcolor{Top1!30!cyan!40}83.25 \\
		& Gen-Poly       & 97.10 & 96.71 & 99.65 & 99.57 & 92.78 & 90.07 & {--} & {--} \\
		& SheafAN        & 97.74 & 97.50 & 99.47 & 99.35 & 92.14 & 89.04 & 86.56 & 80.22 \\
		\midrule
		\multirow{5}{*}{\rotatebox[origin=c]{90}{Att.\,Ens.}}
		& Bundle         & \cellcolor{Top1!30!cyan!40}98.43 & 98.13 & \cellcolor{Top1!30!cyan!40}99.66 & 99.60 & 93.82 & 91.61 & \bfseries\cellcolor{Top1!30!cyan!40}90.73 & \bfseries\cellcolor{Top1!30!cyan!40}85.89 \\
		& General        & 98.42 & \cellcolor{Top1!30!cyan!40}98.21 & \cellcolor{Top1!30!cyan!40}99.66 & \bfseries\cellcolor{Top1!30!cyan!40}99.61 & 93.67 & 91.48 & 89.26 & 83.09 \\
		& Bundle-Poly    & 98.28 & 98.06 & \cellcolor{Top1!30!cyan!40}99.66 & 99.59 & 93.61 & 91.15 & 89.43 & 83.36 \\
		& Gen-Poly       & 98.16 & 97.83 & 99.64 & 99.57 & \bfseries\cellcolor{Top1!30!cyan!40}100.00 & \bfseries\cellcolor{Top1!30!cyan!40}100.00 & 64.60 & 64.60 \\
		& SheafAN        & 96.78 & 96.11 & 99.48 & 99.36 & 92.23 & 89.05 & 86.72 & 80.14 \\
		\midrule
		\multirow{5}{*}{\rotatebox[origin=c]{90}{Att.\,NE}}
		& Bundle         & \cellcolor{Top1!30!cyan!40}98.45 & \cellcolor{Top1!30!cyan!40}98.28 & 99.65 & 99.59 & \cellcolor{Top1!30!cyan!40}92.67 & 90.02 & 89.31 & \cellcolor{Top1!30!cyan!40}83.20 \\
		& General        & 98.43 & 98.25 & 99.66 & \bfseries\cellcolor{Top1!30!cyan!40}99.61 & 92.54 & \cellcolor{Top1!30!cyan!40}90.04 & 89.18 & 83.00 \\
		& Bundle-Poly    & 98.25 & 97.90 & 99.66 & 99.59 & 92.60 & 89.92 & \cellcolor{Top1!30!cyan!40}89.33 & 83.18 \\
		& Gen-Poly       & 98.29 & 97.93 & \cellcolor{Top1!30!cyan!40}99.67 & 99.59 & 92.58 & 89.90 & {--} & {--} \\
		& SheafAN        & 97.62 & 97.22 & 99.50 & 99.38 & 91.74 & 88.70 & 83.05 & 76.96 \\
		\midrule
		\multirow{5}{*}{\rotatebox[origin=c]{90}{Att.\,EE}}
		& Bundle         & 98.38 & 98.17 & 99.65 & \cellcolor{Top1!30!cyan!40}99.60 & 93.78 & 91.54 & 89.28 & 83.17 \\
		& General        & 98.44 & 98.14 & 99.64 & 99.58 & \cellcolor{Top1!30!cyan!40}93.87 & \cellcolor{Top1!30!cyan!40}91.55 & \cellcolor{Top1!30!cyan!40}89.93 & \cellcolor{Top1!30!cyan!40}84.48 \\
		& Bundle-Poly    & 98.05 & 97.61 & \cellcolor{Top1!30!cyan!40}99.66 & 99.59 & 93.54 & 91.13 & 89.41 & 83.26 \\
		& Gen-Poly       & \bfseries\cellcolor{Top1!30!cyan!40}100.00 & \bfseries\cellcolor{Top1!30!cyan!40}100.00 & 99.65 & 99.58 & 93.61 & 91.23 & {--} & {--} \\
		& SheafAN        & 94.78 & 92.51 & 99.52 & 99.41 & 91.81 & 88.42 & 86.27 & 80.13 \\
		\midrule
		\multirow{5}{*}{\rotatebox[origin=c]{90}{Att.\,NT}}
		& Bundle         & 98.41 & 98.24 & 99.66 & 99.59 & 92.55 & \cellcolor{Top1!30!cyan!40}90.05 & 89.28 & 83.16 \\
		& General        & \cellcolor{Top1!30!cyan!40}98.45 & \cellcolor{Top1!30!cyan!40}98.28 & 99.65 & \cellcolor{Top1!30!cyan!40}99.60 & 92.41 & 89.94 & 89.04 & 82.83 \\
		& Bundle-Poly    & 98.36 & 97.98 & 99.66 & 99.58 & \cellcolor{Top1!30!cyan!40}92.63 & 89.82 & \cellcolor{Top1!30!cyan!40}89.39 & \cellcolor{Top1!30!cyan!40}83.19 \\
		& Gen-Poly       & 98.34 & 98.06 & \cellcolor{Top1!30!cyan!40}99.67 & 99.59 & 92.56 & 89.79 & {--} & {--} \\
		& SheafAN        & 97.94 & 97.62 & 99.49 & 99.42 & 91.18 & 87.58 & 82.40 & 76.25 \\
		\midrule
		\multirow{5}{*}{\rotatebox[origin=c]{90}{Att.\,ET}}
		& Bundle         & 98.35 & 98.13 & 99.66 & \cellcolor{Top1!30!cyan!40}99.60 & 93.75 & 91.44 & \cellcolor{Top1!30!cyan!40}90.27 & \cellcolor{Top1!30!cyan!40}84.88 \\
		& General        & \cellcolor{Top1!30!cyan!40}98.43 & \cellcolor{Top1!30!cyan!40}98.19 & 99.66 & \cellcolor{Top1!30!cyan!40}99.60 & \cellcolor{Top1!30!cyan!40}93.82 & \cellcolor{Top1!30!cyan!40}91.55 & 90.05 & \cellcolor{Top1!30!cyan!40}84.88 \\
		& Bundle-Poly    & 98.27 & 97.95 & \cellcolor{Top1!30!cyan!40}99.67 & 99.58 & 93.41 & 91.09 & 89.42 & 83.34 \\
		& Gen-Poly       & 98.22 & 97.82 & 99.66 & 99.58 & 93.81 & 91.49 & {--} & {--} \\
		& SheafAN        & 93.70 & 91.14 & 99.52 & 99.44 & 91.92 & 87.95 & 86.11 & 79.88 \\
		\bottomrule
	\end{tabular}}
}
\end{table}

\begin{table}[!htbp]
    \centering
    \caption{
        \emph{Sheaf \emph{attention} predictor $\times$ restriction map ablation, graph classification.}
        \ablationlegend
    }
    \label{tab:gc-ablation-att}
    \maxsizebox{\textwidth}{!}
{
	\renewcommand{\arraystretch}{0.85}
	\setlength{\tabcolsep}{4pt}
	\resizebox{0.8\textwidth}{!}{\begin{tabular}{clSSSSSSSS}
		\toprule
		{\thead{Sheaf\\Predictor}} & {}
		& \multicolumn{2}{c}{\thead{MUTAG}}
		& \multicolumn{2}{c}{\thead{PROTEINS}}
		& \multicolumn{2}{c}{\thead{ENZYMES}}
		& \multicolumn{2}{c}{\thead{NCI1}} \\
		\cmidrule(lr){3-4}
		\cmidrule(lr){5-6}
		\cmidrule(lr){7-8}
		\cmidrule(lr){9-10}
		{} & {}
		& {\thead{Macro F1}} & {\thead{Micro F1}}
		& {\thead{Macro F1}} & {\thead{Micro F1}}
		& {\thead{Macro F1}} & {\thead{Micro F1}}
		& {\thead{Macro F1}} & {\thead{Micro F1}} \\
		\midrule
		\multirow{5}{*}{\rotatebox[origin=c]{90}{Att.\,NSD}}
		& Bundle         & 64.91 & 70.00 & 25.79 & 33.04 & \cellcolor{Top1!30!cyan!40}33.65 & \cellcolor{Top1!30!cyan!40}35.00 & 55.12 & 55.12 \\
		& General        & 71.51 & 75.00 & 24.32 & 32.14 & 5.56 & 20.00 & 35.44 & 35.44 \\
		& Bundle-Poly    & 76.19 & 80.00 & 24.32 & 32.14 & 12.17 & 20.00 & 62.08 & 62.08 \\
		& Gen-Poly       & \bfseries\cellcolor{Top1!30!cyan!40}100.00 & \bfseries\cellcolor{Top1!30!cyan!40}100.00 & \cellcolor{Top1!30!cyan!40}66.75 & \cellcolor{Top1!30!cyan!40}71.43 & 14.13 & 23.33 & \bfseries\cellcolor{Top1!30!cyan!40}72.30 & \bfseries\cellcolor{Top1!30!cyan!40}72.30 \\
		& SheafAN        & 94.88 & 95.00 & 52.37 & 52.68 & 12.35 & 21.67 & 48.47 & 48.47 \\
		\midrule
		\multirow{5}{*}{\rotatebox[origin=c]{90}{Att.\,TE}}
		& Bundle         & 79.80 & 80.00 & 54.29 & 54.46 & 18.27 & 26.67 & 59.88 & 59.88 \\
		& General        & 41.33 & 60.00 & 63.25 & 63.39 & 8.01 & 21.67 & 58.82 & 58.82 \\
		& Bundle-Poly    & 59.60 & 65.00 & 72.57 & 75.89 & \cellcolor{Top1!30!cyan!40}21.29 & 21.67 & \cellcolor{Top1!30!cyan!40}62.08 & \cellcolor{Top1!30!cyan!40}62.08 \\
		& Gen-Poly       & \cellcolor{Top1!30!cyan!40}100.00 & \cellcolor{Top1!30!cyan!40}100.00 & 49.42 & 50.00 & 16.60 & \cellcolor{Top1!30!cyan!40}28.33 & 60.06 & 60.06 \\
		& SheafAN        & 78.02 & 80.00 & \cellcolor{Top1!30!cyan!40}75.24 & \cellcolor{Top1!30!cyan!40}76.79 & 17.05 & 28.33 & {--} & {--} \\
		\midrule
		\multirow{5}{*}{\rotatebox[origin=c]{90}{Att.\,Ens.}}
		& Bundle         & 84.00 & 85.00 & 43.49 & 44.64 & 18.47 & \cellcolor{Top1!30!cyan!40}30.00 & 60.38 & 60.38 \\
		& General        & 73.33 & 75.00 & \cellcolor{Top1!30!cyan!40}64.22 & \cellcolor{Top1!30!cyan!40}66.96 & 5.56 & 20.00 & \cellcolor{Top1!30!cyan!40}63.49 & \cellcolor{Top1!30!cyan!40}63.49 \\
		& Bundle-Poly    & 73.33 & 75.00 & 62.53 & 63.39 & \cellcolor{Top1!30!cyan!40}22.13 & 23.33 & 19.44 & 19.44 \\
		& Gen-Poly       & 94.67 & \cellcolor{Top1!30!cyan!40}95.00 & 41.70 & 43.75 & 16.62 & 23.33 & {--} & {--} \\
		& SheafAN        & \cellcolor{Top1!30!cyan!40}94.88 & 95.00 & 28.63 & 34.82 & 15.35 & 20.00 & 19.18 & 19.18 \\
		\midrule
		\multirow{5}{*}{\rotatebox[origin=c]{90}{Att.\,NE}}
		& Bundle         & 94.88 & 95.00 & 24.32 & 32.14 & \cellcolor{Top1!30!cyan!40}36.67 & \cellcolor{Top1!30!cyan!40}43.33 & 55.36 & 55.36 \\
		& General        & 73.33 & 75.00 & 25.25 & 32.14 & 17.36 & 30.00 & 28.09 & 28.09 \\
		& Bundle-Poly    & 94.88 & 95.00 & 49.60 & 67.86 & 2.98 & 8.33 & \cellcolor{Top1!30!cyan!40}63.02 & \cellcolor{Top1!30!cyan!40}63.02 \\
		& Gen-Poly       & \cellcolor{Top1!30!cyan!40}100.00 & \cellcolor{Top1!30!cyan!40}100.00 & \cellcolor{Top1!30!cyan!40}70.10 & \cellcolor{Top1!30!cyan!40}74.11 & 19.45 & 31.67 & 5.26 & 5.26 \\
		& SheafAN        & 71.51 & 75.00 & 63.04 & 63.39 & 31.74 & 35.00 & 62.08 & 62.08 \\
		\midrule
		\multirow{5}{*}{\rotatebox[origin=c]{90}{Att.\,EE}}
		& Bundle         & 78.02 & 80.00 & 33.97 & 38.39 & 16.35 & \cellcolor{Top1!30!cyan!40}26.67 & \cellcolor{Top1!30!cyan!40}62.30 & \cellcolor{Top1!30!cyan!40}62.30 \\
		& General        & 94.88 & 95.00 & 24.32 & 32.14 & 19.35 & 25.00 & 60.44 & 60.44 \\
		& Bundle-Poly    & 94.88 & 95.00 & 63.25 & 63.39 & 21.27 & 23.33 & 19.18 & 19.18 \\
		& Gen-Poly       & \cellcolor{Top1!30!cyan!40}100.00 & \cellcolor{Top1!30!cyan!40}100.00 & 41.27 & 43.75 & \cellcolor{Top1!30!cyan!40}23.58 & 26.67 & {--} & {--} \\
		& SheafAN        & 37.50 & 60.00 & \cellcolor{Top1!30!cyan!40}70.05 & \cellcolor{Top1!30!cyan!40}71.43 & 18.49 & 26.67 & 11.76 & 11.76 \\
		\midrule
		\multirow{5}{*}{\rotatebox[origin=c]{90}{Att.\,NT}}
		& Bundle         & 79.17 & 80.00 & 78.84 & 80.36 & 51.23 & 53.33 & 67.98 & 67.98 \\
		& General        & 79.17 & 80.00 & 46.27 & 46.43 & 13.17 & 26.67 & 24.45 & 24.45 \\
		& Bundle-Poly    & 84.65 & 85.00 & \bfseries\cellcolor{Top1!30!cyan!40}79.69 & \bfseries\cellcolor{Top1!30!cyan!40}81.25 & \bfseries\cellcolor{Top1!30!cyan!40}56.71 & \bfseries\cellcolor{Top1!30!cyan!40}56.67 & {--} & {--} \\
		& Gen-Poly       & \cellcolor{Top1!30!cyan!40}89.58 & \cellcolor{Top1!30!cyan!40}90.00 & 75.94 & 80.36 & 41.23 & 43.33 & \cellcolor{Top1!30!cyan!40}70.06 & \cellcolor{Top1!30!cyan!40}70.06 \\
		& SheafAN        & 84.96 & 85.00 & 76.95 & 79.46 & 46.52 & 50.00 & 64.30 & 64.30 \\
		\midrule
		\multirow{5}{*}{\rotatebox[origin=c]{90}{Att.\,ET}}
		& Bundle         & 84.65 & 85.00 & 24.32 & 32.14 & 7.78 & 21.67 & 22.81 & 22.81 \\
		& General        & 79.17 & 80.00 & 38.92 & 41.96 & 8.67 & 23.33 & 47.85 & 47.85 \\
		& Bundle-Poly    & 89.90 & 90.00 & 24.32 & 32.14 & \cellcolor{Top1!30!cyan!40}24.06 & \cellcolor{Top1!30!cyan!40}25.00 & 64.74 & 64.74 \\
		& Gen-Poly       & \cellcolor{Top1!30!cyan!40}100.00 & \cellcolor{Top1!30!cyan!40}100.00 & \cellcolor{Top1!30!cyan!40}65.17 & \cellcolor{Top1!30!cyan!40}66.07 & 21.86 & 23.33 & \cellcolor{Top1!30!cyan!40}69.89 & \cellcolor{Top1!30!cyan!40}69.89 \\
		& SheafAN        & 94.88 & 95.00 & 24.32 & 32.14 & 13.36 & 20.00 & 62.08 & 62.08 \\
		\bottomrule
	\end{tabular}}
}
\end{table}

\subsubsection{Stalk Dimension Ablation Study}
\label{app:stalk_ablation}
Tables~\ref{tab:ablation_node_classification}--\ref{tab:ablation_lp_aupr_auroc} report the effect of varying the stalk dimension $d \in \{2,3,4,5\}$ across all restriction map families, tasks, and datasets. Each cell reports Macro-F1\,/\,Micro-F1 for classification tasks and AUPR\,/\,AUROC for link prediction. Several patterns emerge.

First, no single stalk dimension is universally optimal: the best $d$ varies by dataset and family, suggesting that $d$ should be treated as a hyperparameter tuned per task. For instance, Gen.-P peaks at $d\!=\!2$ on both ACM (94.3/94.2) and DBLP (95.0/95.4), but at $d\!=\!3$ on PROTEINS (80.6/82.1) and $d\!=\!2$ on NCI1 (79.5/79.5).

Second, polynomial variants (Diag-P, Bundle-P, Gen.-P) consistently match or outperform their non-polynomial counterparts on node and graph classification. Gen.-P achieves the global best Macro-F1 on ACM (94.3), DBLP (95.0), PROTEINS (80.6), and NCI1 (79.5). On MUTAG, both Bundle-P and Gen.-P reach a perfect 100.0/100.0 across all stalk dimensions, indicating that the polynomial spectral filtering combined with orthogonal or general restriction maps provides sufficient expressivity to fully separate the two classes. However, this advantage does not extend uniformly to all datasets: on ENZYMES, Diag-P performs poorly (peaking at only 25.6/30.0), well below the non-polynomial Diag (53.0/53.3) and Bundle (62.3/63.3 at $d\!=\!4$).

Third, SheafAN lags behind the Laplacian-based families on node classification, particularly on IMDB, where it peaks at 56.4/59.3 Macro/Micro-F1 versus 64.5/67.2 for Bundle. The gap is smaller but consistent on ACM and DBLP. However, SheafAN is more competitive on link prediction, where it achieves strong AUPR/AUROC on LastFM (97.9/97.6 at $d\!=\!5$) and near-saturated performance on MovieLens. This suggests that the attention-based adjacency aggregation of SheafAN is better suited for pairwise scoring tasks than for node-level classification.

Fourth, link prediction performance is largely insensitive to $d$. AUPR and AUROC are near-saturated across all configurations (e.g., 98.3--98.5 AUPR on LastFM, 99.5--99.7 on MovieLens), and vary by less than 0.3 percentage points within each family. This contrasts sharply with node and graph classification, where performance swings of 5--20 points across stalk dimensions are common (e.g., General on MUTAG ranges from 74.9 at $d\!=\!5$ to 94.9 at $d\!=\!3$).

Finally, the non-polynomial families (Diag, Bundle, General) tend to be more stable across $d$ than their polynomial counterparts. The polynomial families occasionally exhibit sharp performance drops, such as Diag-P on ENZYMES falling from 25.6 at $d\!=\!3$ to 14.0 at $d\!=\!4$, or Gen.-P on ACM degrading from 94.3 at $d\!=\!2$ to 90.6 at $d\!=\!5$. This instability may stem from the higher-order Chebyshev coefficients amplifying noise when the graph structure does not support long-range spectral patterns, and reinforces the recommendation to treat both $d$ and the restriction map family as hyperparameters.

\begin{table}[!htbp]
\centering
\caption{Node classification: Macro-F1\,/\,Micro-F1 (\%) by restriction map family and stalk dimension $d$. \colorbox{Top1!30!cyan!40}{Cyan} = best in family, \textbf{bold} = global best per dataset.}
\label{tab:ablation_node_classification}
\resizebox{\textwidth}{!}{%
\setlength{\tabcolsep}{2pt}
\renewcommand{\arraystretch}{1.1}
\small
\begin{tabular}{lcccccccccccc}
\toprule
 & \multicolumn{4}{c}{\textbf{ACM}} & \multicolumn{4}{c}{\textbf{DBLP}} & \multicolumn{4}{c}{\textbf{IMDB}} \\
 \cmidrule(lr){2-5} \cmidrule(lr){6-9} \cmidrule(lr){10-13}
\textbf{Family} & $d\!=\!2$ & $d\!=\!3$ & $d\!=\!4$ & $d\!=\!5$ & $d\!=\!2$ & $d\!=\!3$ & $d\!=\!4$ & $d\!=\!5$ & $d\!=\!2$ & $d\!=\!3$ & $d\!=\!4$ & $d\!=\!5$ \\
\midrule
Diag & \cellcolor{Top1!30!cyan!40}93.7/93.6 & 93.0/92.9 & 92.6/92.5 & 93.0/92.9 & \cellcolor{Top1!30!cyan!40}94.0/94.5 & \cellcolor{Top1!30!cyan!40}94.1/94.5 & \cellcolor{Top1!30!cyan!40}94.1/94.5 & 93.5/94.0 & 62.4/65.4 & \cellcolor{Top1!30!cyan!40}63.6/66.6 & 62.8/65.9 & 63.2/65.8 \\
Bundle & \cellcolor{Top1!30!cyan!40}92.6/92.5 & 91.2/91.1 & 91.0/90.8 & 92.1/92.1 & \cellcolor{Top1!30!cyan!40}93.8/94.3 & \cellcolor{Top1!30!cyan!40}93.7/94.2 & 93.2/93.7 & 93.4/93.9 & \cellcolor{Top1!30!cyan!40}64.4/\textbf{67.2} & \cellcolor{Top1!30!cyan!40}\textbf{64.5}/66.6 & 63.9/66.5 & 64.2/66.4 \\
General & 91.8/91.7 & 91.7/91.7 & 91.9/91.8 & \cellcolor{Top1!30!cyan!40}92.1/92.0 & 93.8/94.3 & 94.0/94.4 & \cellcolor{Top1!30!cyan!40}94.7/95.1 & 94.0/94.4 & \cellcolor{Top1!30!cyan!40}63.5/65.6 & 62.5/65.5 & 62.4/65.4 & \cellcolor{Top1!30!cyan!40}63.0/65.8 \\
\midrule
Diag-P & 93.3/93.2 & 93.1/93.0 & \cellcolor{Top1!30!cyan!40}94.2/94.1 & 93.5/93.4 & \cellcolor{Top1!30!cyan!40}94.9/95.3 & 94.7/95.1 & \cellcolor{Top1!30!cyan!40}94.9/95.3 & 94.6/95.0 & \cellcolor{Top1!30!cyan!40}61.5/64.7 & \cellcolor{Top1!30!cyan!40}61.6/64.6 & 61.1/64.1 & 60.2/63.6 \\
Bundle-P & 90.3/90.2 & \cellcolor{Top1!30!cyan!40}92.1/92.0 & 89.3/89.1 & 90.3/90.2 & \cellcolor{Top1!30!cyan!40}94.7/95.1 & 93.3/93.8 & 94.6/95.0 & 93.6/94.0 & 61.5/64.3 & 61.8/64.6 & 61.0/64.3 & \cellcolor{Top1!30!cyan!40}61.9/65.2 \\
Gen.-P & \cellcolor{Top1!30!cyan!40}\textbf{94.3}/\textbf{94.2} & 93.6/93.5 & 93.1/93.1 & 90.6/90.5 & \cellcolor{Top1!30!cyan!40}\textbf{95.0}/\textbf{95.4} & 94.4/94.8 & 94.6/95.0 & 94.6/95.0 & \cellcolor{Top1!30!cyan!40}62.4/65.8 & \cellcolor{Top1!30!cyan!40}62.5/64.9 & 62.3/65.3 & 60.9/64.6 \\
\midrule
SheafAN & 88.9/89.0 & 89.7/89.7 & 88.8/88.8 & \cellcolor{Top1!30!cyan!40}90.9/90.9 & 90.9/91.6 & 91.1/91.9 & \cellcolor{Top1!30!cyan!40}92.4/93.0 & 87.3/88.3 & \cellcolor{Top1!30!cyan!40}56.1/60.2 & 56.2/59.2 & 55.9/59.2 & \cellcolor{Top1!30!cyan!40}56.4/59.3 \\
\bottomrule
\end{tabular}}
\end{table}

\begin{table}[!htbp]
\centering
\caption{Graph classification: Macro-F1\,/\,Micro-F1 (\%) by restriction map family and stalk dimension $d$. \colorbox{Top1!30!cyan!40}{Cyan} = best in family, \textbf{bold} = global best per dataset.}
\label{tab:ablation_graph_classification}
\resizebox{\textwidth}{!}{%
\setlength{\tabcolsep}{2pt}
\renewcommand{\arraystretch}{1.1}
\small
\begin{tabular}{lcccccccccccccccc}
\toprule
 & \multicolumn{4}{c}{\textbf{MUTAG}} & \multicolumn{4}{c}{\textbf{PROTEINS}} & \multicolumn{4}{c}{\textbf{ENZYMES}} & \multicolumn{4}{c}{\textbf{NCI1}} \\
 \cmidrule(lr){2-5} \cmidrule(lr){6-9} \cmidrule(lr){10-13} \cmidrule(lr){14-17}
\textbf{Family} & $d\!=\!2$ & $d\!=\!3$ & $d\!=\!4$ & $d\!=\!5$ & $d\!=\!2$ & $d\!=\!3$ & $d\!=\!4$ & $d\!=\!5$ & $d\!=\!2$ & $d\!=\!3$ & $d\!=\!4$ & $d\!=\!5$ & $d\!=\!2$ & $d\!=\!3$ & $d\!=\!4$ & $d\!=\!5$ \\
\midrule
Diag & 80.0/80.0 & 80.0/80.0 & \cellcolor{Top1!30!cyan!40}83.3/83.3 & 76.9/76.9 & \cellcolor{Top1!30!cyan!40}74.6/75.9 & 67.3/67.9 & \cellcolor{Top1!30!cyan!40}73.6/75.9 & 61.5/67.9 & \cellcolor{Top1!30!cyan!40}53.0/53.3 & 46.1/46.7 & 34.2/35.0 & 39.1/40.0 & 72.1/72.1 & 68.5/68.5 & \cellcolor{Top1!30!cyan!40}73.7/73.7 & 61.8/61.8 \\
Bundle & 78.0/80.0 & 84.7/85.0 & \cellcolor{Top1!30!cyan!40}94.9/95.0 & 84.7/85.0 & 74.8/76.8 & 75.8/77.7 & \cellcolor{Top1!30!cyan!40}78.8/80.4 & 54.4/54.5 & 50.8/50.0 & 51.2/53.3 & \cellcolor{Top1!30!cyan!40}\textbf{62.3}/\textbf{63.3} & 51.0/51.7 & 62.6/62.6 & 57.6/57.6 & \cellcolor{Top1!30!cyan!40}68.0/68.0 & 60.4/60.4 \\
General & 89.0/90.0 & \cellcolor{Top1!30!cyan!40}94.9/95.0 & 89.9/90.0 & 74.9/75.0 & 71.3/72.3 & 64.2/74.1 & \cellcolor{Top1!30!cyan!40}78.7/81.2 & 64.2/67.0 & \cellcolor{Top1!30!cyan!40}49.9/51.7 & 47.7/50.0 & \cellcolor{Top1!30!cyan!40}50.7/51.7 & 46.9/46.7 & 63.2/63.2 & 64.4/64.4 & \cellcolor{Top1!30!cyan!40}67.5/67.5 & 62.1/62.1 \\
\midrule
Diag-P & 95.7/95.7 & 95.7/95.7 & 96.0/96.0 & \cellcolor{Top1!30!cyan!40}\textbf{100.0}/\textbf{100.0} & 51.8/67.9 & \cellcolor{Top1!30!cyan!40}78.9/\textbf{82.1} & 71.6/74.1 & 63.2/74.1 & 22.7/28.3 & \cellcolor{Top1!30!cyan!40}25.6/30.0 & 14.0/21.7 & 14.9/23.3 & 60.2/60.2 & \cellcolor{Top1!30!cyan!40}62.1/62.1 & \cellcolor{Top1!30!cyan!40}62.1/62.1 & 61.4/61.4 \\
Bundle-P & \cellcolor{Top1!30!cyan!40}\textbf{100.0}/\textbf{100.0} & \cellcolor{Top1!30!cyan!40}\textbf{100.0}/\textbf{100.0} & \cellcolor{Top1!30!cyan!40}\textbf{100.0}/\textbf{100.0} & \cellcolor{Top1!30!cyan!40}\textbf{100.0}/\textbf{100.0} & 75.6/77.7 & 78.8/80.4 & 78.4/80.4 & \cellcolor{Top1!30!cyan!40}79.8/\textbf{82.1} & 52.6/53.3 & 46.5/48.3 & 51.9/51.7 & \cellcolor{Top1!30!cyan!40}56.7/56.7 & 69.2/69.2 & 54.4/54.4 & 74.7/74.7 & \cellcolor{Top1!30!cyan!40}76.3/76.3 \\
Gen.-P & \cellcolor{Top1!30!cyan!40}\textbf{100.0}/\textbf{100.0} & \cellcolor{Top1!30!cyan!40}\textbf{100.0}/\textbf{100.0} & \cellcolor{Top1!30!cyan!40}\textbf{100.0}/\textbf{100.0} & \cellcolor{Top1!30!cyan!40}\textbf{100.0}/\textbf{100.0} & 75.9/80.4 & \cellcolor{Top1!30!cyan!40}\textbf{80.6}/\textbf{82.1} & 78.7/81.2 & 77.5/79.5 & 34.4/36.7 & 41.6/43.3 & 41.7/43.3 & \cellcolor{Top1!30!cyan!40}48.1/48.3 & \cellcolor{Top1!30!cyan!40}\textbf{79.5}/\textbf{79.5} & 72.9/72.9 & 71.5/71.5 & 73.0/73.0 \\
\midrule
SheafAN & 89.9/90.0 & \cellcolor{Top1!30!cyan!40}94.9/95.0 & \cellcolor{Top1!30!cyan!40}94.9/95.0 & \cellcolor{Top1!30!cyan!40}94.9/95.0 & 75.8/78.6 & 76.9/79.5 & \cellcolor{Top1!30!cyan!40}77.2/79.5 & \cellcolor{Top1!30!cyan!40}76.7/80.4 & 46.6/48.3 & \cellcolor{Top1!30!cyan!40}49.3/51.7 & 41.6/41.7 & 37.3/40.0 & 72.7/72.7 & 71.9/71.9 & 73.6/73.6 & \cellcolor{Top1!30!cyan!40}75.1/75.1 \\
\bottomrule
\end{tabular}}
\end{table}

\begin{table}[!htbp]
\centering
\caption{Link prediction: (AUPR\,/\,AUROC, \%). by restriction map family and stalk dimension $d$. \colorbox{Top1!30!cyan!40}{Cyan} = best in family, \textbf{bold} = global best per dataset.}
\label{tab:ablation_lp_aupr_auroc}
\resizebox{\textwidth}{!}{%
\setlength{\tabcolsep}{2pt}
\renewcommand{\arraystretch}{1.1}
\small
\begin{tabular}{l cccc cccc cccc cccc}
\toprule
& \multicolumn{4}{c}{\textbf{LastFM}} & \multicolumn{4}{c}{\textbf{MovieLens}} & \multicolumn{4}{c}{\textbf{Amazon}} & \multicolumn{4}{c}{\textbf{YouTube}} \\
\cmidrule(lr){2-5} \cmidrule(lr){6-9} \cmidrule(lr){10-13} \cmidrule(lr){14-17}
\textbf{Family} & $d\!=\!2$ & $d\!=\!3$ & $d\!=\!4$ & $d\!=\!5$ & $d\!=\!2$ & $d\!=\!3$ & $d\!=\!4$ & $d\!=\!5$ & $d\!=\!2$ & $d\!=\!3$ & $d\!=\!4$ & $d\!=\!5$ & $d\!=\!2$ & $d\!=\!3$ & $d\!=\!4$ & $d\!=\!5$ \\
\midrule
Diag & 98.4/98.2 & \cellcolor{Top1!30!cyan!40}98.5/98.2 & 98.4/98.2 & \cellcolor{Top1!30!cyan!40}98.5/98.2 & \cellcolor{Top1!30!cyan!40}\textbf{99.7}/99.6 & \cellcolor{Top1!30!cyan!40}99.7/99.6 & \cellcolor{Top1!30!cyan!40}99.7/99.6 & \cellcolor{Top1!30!cyan!40}\textbf{99.7}/99.6 & 93.8/91.5 & \cellcolor{Top1!30!cyan!40}93.8/91.6 & \cellcolor{Top1!30!cyan!40}93.8/91.6 & 93.8/91.5 & 90.5/85.4 & \cellcolor{Top1!30!cyan!40}90.7/86.0 & 90.6/85.7 & 90.6/85.9 \\
Bundle & 98.5/98.2 & \cellcolor{Top1!30!cyan!40}98.4/98.3 & 98.5/98.2 & \cellcolor{Top1!30!cyan!40}98.5/98.2 & \cellcolor{Top1!30!cyan!40}99.7/\textbf{99.6} & 99.7/99.6 & \cellcolor{Top1!30!cyan!40}99.7/99.6 & \cellcolor{Top1!30!cyan!40}99.7/99.6 & 93.7/91.3 & \cellcolor{Top1!30!cyan!40}93.8/91.6 & 93.7/91.5 & \cellcolor{Top1!30!cyan!40}93.8/91.6 & 90.1/84.4 & 89.3/83.2 & 90.4/85.4 & \cellcolor{Top1!30!cyan!40}90.7/85.9 \\
General & \cellcolor{Top1!30!cyan!40}98.5/98.3 & 98.5/98.3 & \cellcolor{Top1!30!cyan!40}98.5/98.3 & 98.5/98.3 & \cellcolor{Top1!30!cyan!40}99.7/99.6 & \cellcolor{Top1!30!cyan!40}99.7/\textbf{99.6} & \cellcolor{Top1!30!cyan!40}99.7/\textbf{99.6} & \cellcolor{Top1!30!cyan!40}99.7/\textbf{99.6} & 93.8/91.6 & 93.8/91.6 & \cellcolor{Top1!30!cyan!40}93.8/91.6 & \cellcolor{Top1!30!cyan!40}93.8/91.7 & 90.7/86.0 & 90.1/84.9 & \cellcolor{Top1!30!cyan!40}\textbf{91.3}/\textbf{87.3} & \cellcolor{Top1!30!cyan!40}\textbf{91.3}/\textbf{87.3} \\
\midrule
Diag-P & 98.3/97.9 & 98.4/98.0 & 98.3/98.0 & \cellcolor{Top1!30!cyan!40}98.4/98.1 & 99.6/99.5 & \cellcolor{Top1!30!cyan!40}99.7/99.6 & \cellcolor{Top1!30!cyan!40}99.6/99.6 & \cellcolor{Top1!30!cyan!40}99.7/99.6 & 93.2/90.6 & 93.3/90.5 & \cellcolor{Top1!30!cyan!40}93.5/90.9 & 93.3/90.8 & 90.1/85.3 & 90.5/85.8 & \cellcolor{Top1!30!cyan!40}90.7/86.5 & 90.6/85.4 \\
Bundle-P & 98.3/98.0 & 98.3/98.1 & \cellcolor{Top1!30!cyan!40}98.3/98.1 & \cellcolor{Top1!30!cyan!40}98.4/98.0 & 99.7/99.6 & \cellcolor{Top1!30!cyan!40}99.7/99.6 & \cellcolor{Top1!30!cyan!40}99.7/99.6 & 99.7/99.6 & 93.5/91.1 & 93.5/91.1 & 93.6/91.1 & \cellcolor{Top1!30!cyan!40}93.6/91.2 & \cellcolor{Top1!30!cyan!40}89.4/83.4 & 89.4/83.3 & 89.4/83.3 & 89.4/83.3 \\
Gen.-P & 98.3/98.1 & 98.3/98.0 & 98.3/98.1 & \cellcolor{Top1!30!cyan!40}\textbf{98.5}/\textbf{98.3} & 99.7/99.6 & \cellcolor{Top1!30!cyan!40}99.7/99.6 & \cellcolor{Top1!30!cyan!40}99.7/99.6 & \cellcolor{Top1!30!cyan!40}99.7/99.6 & \cellcolor{Top1!30!cyan!40}\textbf{93.8}/\textbf{91.7} & 93.8/91.5 & 93.8/91.6 & 93.8/91.4 & \cellcolor{Top1!30!cyan!40}89.2/83.0 & -/- & -/- & -/- \\
\midrule
SheafAN & 97.7/97.5 & 97.7/97.3 & 97.8/97.2 & \cellcolor{Top1!30!cyan!40}97.9/97.6 & 99.5/99.4 & 99.5/99.4 & \cellcolor{Top1!30!cyan!40}99.6/99.5 & 99.5/99.4 & 91.9/88.8 & 92.0/88.7 & \cellcolor{Top1!30!cyan!40}92.2/89.1 & \cellcolor{Top1!30!cyan!40}92.1/89.2 & 87.0/80.5 & \cellcolor{Top1!30!cyan!40}87.1/80.6 & 87.0/80.5 & 86.1/79.6 \\
\bottomrule
\end{tabular}}
\end{table}
\subsubsection{Layer Depth Ablation Study}
\label{app:layer_ablation}
Tables~\ref{tab:ablation_layers_nc}--\ref{tab:ablation_layers_lp} report the effect of varying the number of message passing layers $L \in \{2,3,4,5,6,7,8\}$ across all restriction map families, tasks, and datasets, alongside the homogeneous and heterogeneous GNN baselines.
Each cell reports Macro-F1\,/\,Micro-F1 for classification tasks and AUPR\,/\,AUROC for link prediction.
Several patterns emerge.

The most striking observation is the \emph{absence of catastrophic oversmoothing} in \textsc{HetSheaf}.
On ACM, the best Macro-F1 across all sheaf families varies by at most 3.4 percentage points between the shallowest ($L\!=\!2$) and deepest ($L\!=\!8$) configuration (Gen.-P: 94.3/94.2 at $L\!=\!2$, 92.6/92.5 at $L\!=\!8$), and several families (Diag, General) achieve their peak performance at $L\!\geq\!6$.
In contrast, the homogeneous baselines suffer dramatic degradation at depth: on ACM, GAT drops from 92.2/92.1 at $L\!=\!3$ to 17.0/34.3 at $L\!=\!8$, GCN from 92.3/92.2 at $L\!=\!2$ to 32.8/40.9 at $L\!=\!8$, and GIN from 88.7/88.8 at $L\!=\!2$ to 41.3/43.7 at $L\!=\!8$.
Even the heterogeneous baselines are not immune: HAN collapses from 93.6/93.6 at $L\!=\!5$ to 27.2/37.1 at $L\!=\!8$ on ACM, and HGT drops from 92.3/92.3 at $L\!=\!2$ to 40.1/46.0 at $L\!=\!8$.
The only baseline that remains stable across depth is R-GCN, whose edge-type-specific parameterisation provides a form of oversmoothing resistance, though at a substantially higher parameter cost (\Cref{tab:nc_overhead}).

On DBLP, the advantage of depth is more pronounced for sheaf models.
Several families (General, Gen.-P) achieve their best results at $L\!\geq\!4$, with Gen.-P peaking at 95.0/95.4 at $L\!=\!4$ and remaining above 93.9/94.3 through $L\!=\!8$.
The baselines also benefit from moderate depth on DBLP (GIN peaks at 94.8/95.2 at $L\!=\!7$), but none match the best sheaf configurations.
On IMDB, where all methods struggle, the sheaf families show a mild but consistent benefit from additional layers: Diag and Bundle both achieve their best results at $L\!\geq\!6$ (Diag: 63.6/66.6 at $L\!=\!8$; Bundle: 64.5/66.6 at $L\!=\!6$), suggesting that deeper message passing helps capture the complex multi-label structure of this dataset.
Notably, the IMDB baselines are extremely brittle: GCN and HAN oscillate between functional performance at odd layer counts and near-collapse at even ones, a pattern that is entirely absent from the sheaf models.

For graph classification, the depth--performance relationship is noisier, reflecting the smaller test sets and inductive evaluation protocol.
On ENZYMES, deeper sheaf models consistently outperform shallower ones: Bundle achieves 62.3/63.3 at $L\!=\!8$ (its global best) versus 38.2/43.3 at $L\!=\!2$, and Diag reaches 53.0/53.3 at $L\!=\!8$ versus 31.0/40.0 at $L\!=\!2$.
The baselines on ENZYMES are uniformly poor (GCN: 9--13, GAT: 6--14, SAGE: 6--9 Macro-F1 across all depths), and only HAN (46.1/45.0 at $L\!=\!6$) and HGT (48.3/48.3 at $L\!=\!5$) reach moderate performance.
On NCI1, Gen.-P peaks at 79.5/79.5 at $L\!=\!7$, and Diag reaches 78.2/78.2 at $L\!=\!3$, both substantially above the homogeneous baselines (GCN: 64.2/65.0, GAT: 64.2/65.0 at best).
The polynomial families exhibit greater variance across layers than the non-polynomial families: on NCI1, Diag-P ranges from 61.4 to 62.1, whereas Diag ranges from 61.8 to 78.2.
This is consistent with the observation from the stalk dimension ablation that polynomial spectral filtering amplifies sensitivity to hyperparameter choices.

For link prediction (\Cref{tab:ablation_layers_lp}), all sheaf families are effectively invariant to depth.
On LastFM, AUPR varies by at most 0.2 percentage points across layers for all non-polynomial families, and AUROC varies by at most 0.3 points.
On MovieLens, performance is near-constant across all depths and families (AUPR $\geq$ 99.5 and AUROC $\geq$ 99.4 for every sheaf configuration), substantially outperforming HAN which is essentially non-functional on this dataset (AUPR $\approx$ 63.0 regardless of depth).
On Amazon and YouTube, the baselines are more competitive, with HGT achieving the strongest baseline AUPR on both datasets (96.2 and 95.0 respectively), but the sheaf models match or exceed these scores across all layer depths.

Overall, these results demonstrate that the sheaf Laplacian diffusion process provides a principled mechanism for deep message passing without the oversmoothing pathology that limits standard GNNs to shallow architectures.
The restriction maps provide per-edge transport operators that preserve discriminative signal across layers, while the polynomial variants add spectral filtering that can exploit deeper receptive fields when the graph structure supports it.
In practice, we recommend treating $L$ as a hyperparameter tuned alongside $d$ and the restriction map family, with the expectation that $L \in \{3,\ldots,6\}$ providing a robust default range.

\begin{table}[!htbp]
\centering
\caption{Node classification: Macro-F1\,/\,Micro-F1 (\%) by model and number of layers $L$. \colorbox{Top1!30!cyan!40}{Cyan} = best in family, \textbf{bold} = global best per dataset.}
\label{tab:ablation_layers_nc}
\resizebox{\textwidth}{!}{%
\setlength{\tabcolsep}{2pt}
\renewcommand{\arraystretch}{1.1}
\small
\begin{tabular}{l ccccccc ccccccc ccccccc}
\toprule
& \multicolumn{7}{c}{\textbf{ACM}} & \multicolumn{7}{c}{\textbf{DBLP}} & \multicolumn{7}{c}{\textbf{IMDB}} \\
\cmidrule(lr){2-8} \cmidrule(lr){9-15} \cmidrule(lr){16-22}
& $L\!=\!2$ & $L\!=\!3$ & $L\!=\!4$ & $L\!=\!5$ & $L\!=\!6$ & $L\!=\!7$ & $L\!=\!8$
& $L\!=\!2$ & $L\!=\!3$ & $L\!=\!4$ & $L\!=\!5$ & $L\!=\!6$ & $L\!=\!7$ & $L\!=\!8$
& $L\!=\!2$ & $L\!=\!3$ & $L\!=\!4$ & $L\!=\!5$ & $L\!=\!6$ & $L\!=\!7$ & $L\!=\!8$ \\
\midrule
\multicolumn{22}{l}{\textit{Homogeneous baselines}} \\
GCN & 92.3/92.2 & 92.1/92.0 & 65.2/67.3 & 53.6/57.0 & 38.6/42.3 & 30.8/39.1 & 32.8/40.9 & 82.2/83.0 & 88.4/89.2 & 88.4/89.2 & 89.8/90.6 & 90.4/91.1 & 91.0/91.7 & 87.0/88.2 & 15.5/25.8 & 58.3/60.9 & 15.4/39.4 & 57.1/60.1 & 23.3/45.2 & 58.2/62.1 & 16.6/28.2 \\
GAT & 92.1/92.0 & 92.2/92.1 & 83.1/84.1 & 28.1/38.2 & 29.2/37.6 & 28.8/38.3 & 17.0/34.3 & 84.4/85.1 & 90.3/91.0 & 89.4/89.9 & 92.3/92.8 & 90.0/90.7 & 91.1/91.7 & 91.4/92.0 & 24.7/32.9 & 60.2/64.0 & 14.1/40.0 & 59.0/61.3 & 24.7/32.9 & 55.7/61.3 & 23.3/45.2 \\
GIN & 88.7/88.8 & 86.8/86.5 & 35.3/45.2 & 49.0/49.4 & 42.2/43.3 & 41.6/45.4 & 41.3/43.7 & 79.9/80.5 & 93.6/93.9 & 93.2/93.7 & 93.2/93.7 & 92.9/93.4 & 94.8/95.2 & 92.9/93.3 & 53.1/56.9 & 60.2/63.9 & 55.7/60.0 & 57.3/60.8 & 52.2/58.5 & 59.1/62.2 & 52.1/56.7 \\
SAGE & 91.4/91.3 & 91.8/91.7 & 92.0/91.9 & 89.1/89.0 & 90.6/90.5 & 92.2/92.1 & 85.1/85.0 & 83.6/84.6 & 88.0/89.0 & 90.1/90.7 & 88.4/88.9 & 89.8/90.5 & 88.9/89.7 & 87.7/88.5 & 58.2/62.4 & 58.6/61.2 & 59.5/61.8 & 60.6/63.8 & 55.6/57.6 & 59.2/63.4 & 58.1/63.6 \\
\midrule
\multicolumn{22}{l}{\textit{Heterogeneous baselines}} \\
R-GCN & 92.0/91.9 & 93.3/93.2 & 92.9/92.8 & 92.1/92.0 & 91.3/91.2 & 93.0/92.9 & 92.0/91.9 & 93.6/94.0 & 94.3/94.7 & 93.6/94.1 & 94.0/94.5 & 92.8/93.4 & 93.8/94.2 & 93.4/93.9 & 58.8/62.3 & 59.3/61.5 & 59.7/62.4 & 58.3/61.2 & 59.1/61.5 & 59.3/61.6 & 58.4/60.9 \\
HAN & 92.2/92.1 & 91.4/91.3 & 91.0/90.9 & 93.6/93.6 & 92.3/92.3 & 27.1/37.0 & 27.2/37.1 & 82.7/83.3 & 92.1/92.6 & 92.8/93.3 & 90.8/91.4 & 92.7/93.2 & 91.8/92.4 & 94.0/94.4 & 14.1/40.0 & 56.8/61.7 & 17.2/29.5 & 55.5/58.9 & 15.5/25.8 & 25.5/39.5 & 14.1/40.0 \\
HGT & 92.3/92.3 & 91.5/91.4 & 91.8/91.8 & 29.8/37.0 & 91.7/91.7 & 89.0/88.8 & 40.1/46.0 & 82.7/83.6 & 90.6/91.3 & 92.8/93.3 & 91.0/91.7 & 91.2/91.7 & 91.3/91.9 & 88.2/89.0 & 56.5/60.2 & 41.4/49.4 & 51.3/53.6 & 51.5/55.3 & 39.0/56.1 & 35.3/41.8 & 38.8/46.2 \\
\midrule
\multicolumn{22}{l}{\textit{HetSheaf (ours)}} \\
Diag & 93.0/92.9 & 92.6/92.4 & 93.0/92.9 & 92.6/92.5 & \cellcolor{Top1!30!cyan!40}93.7/93.6 & 92.9/92.9 & 92.5/92.5 & 84.5/85.2 & \cellcolor{Top1!30!cyan!40}94.1/94.5 & 93.5/94.0 & 93.5/93.9 & 93.5/94.0 & 94.0/94.5 & 93.8/94.2 & 59.1/63.0 & 62.2/65.2 & 62.2/65.0 & 62.5/65.4 & 62.9/65.4 & 63.2/66.5 & \cellcolor{Top1!30!cyan!40}63.6/66.6 \\
Bundle & 91.5/91.5 & 91.0/90.9 & \cellcolor{Top1!30!cyan!40}92.6/92.5 & 92.1/92.1 & 90.6/90.5 & 91.8/91.7 & 91.8/91.7 & 84.9/85.8 & 93.0/93.5 & 92.6/93.1 & 93.4/93.9 & 93.7/94.2 & 93.3/93.8 & \cellcolor{Top1!30!cyan!40}93.8/94.3 & 59.2/63.2 & 62.6/65.5 & 62.5/66.0 & 62.7/65.7 & \cellcolor{Top1!30!cyan!40}64.5/66.6 & 63.9/66.5 & 64.4/\textbf{67.2} \\
General & 91.9/91.8 & 91.8/91.7 & 91.8/91.7 & 91.7/91.6 & 91.1/91.0 & 91.5/91.4 & \cellcolor{Top1!30!cyan!40}92.1/92.0 & 85.0/85.7 & 93.7/94.1 & 94.4/94.9 & \cellcolor{Top1!30!cyan!40}94.7/95.1 & 94.0/94.4 & 94.0/94.4 & 93.9/94.4 & 58.9/62.7 & 62.3/64.8 & 60.9/64.1 & \cellcolor{Top1!30!cyan!40}63.5/65.8 & 62.4/65.4 & 62.5/65.5 & 62.0/64.9 \\
\cmidrule(lr){1-22}
Diag-P & \cellcolor{Top1!30!cyan!40}94.2/94.1 & 93.8/93.7 & 93.9/93.8 & 92.5/92.4 & 92.6/92.5 & 92.3/92.3 & 93.5/93.4 & 94.5/94.9 & \cellcolor{Top1!30!cyan!40}94.9/95.3 & 94.7/95.1 & 94.6/95.0 & 94.4/94.8 & 94.2/94.6 & 93.9/94.4 & \cellcolor{Top1!30!cyan!40}61.5/64.7 & 61.1/64.1 & 60.5/63.7 & 60.3/63.5 & \cellcolor{Top1!30!cyan!40}61.6/64.6 & 60.6/63.2 & 61.3/64.1 \\
Bundle-P & 89.0/88.9 & 90.2/90.1 & 89.3/89.1 & \cellcolor{Top1!30!cyan!40}92.1/92.0 & 90.3/90.2 & 89.6/89.5 & 90.3/90.2 & 93.3/93.8 & 93.6/94.0 & \cellcolor{Top1!30!cyan!40}94.7/95.1 & 91.4/92.0 & 93.4/93.9 & 78.6/79.5 & 91.8/92.4 & 60.4/64.6 & 61.8/64.2 & \cellcolor{Top1!30!cyan!40}61.9/65.2 & 61.3/64.0 & 59.9/64.3 & 61.5/64.3 & 60.0/64.1 \\
Gen.-P & \cellcolor{Top1!30!cyan!40}\textbf{94.3}/\textbf{94.2} & 91.0/90.9 & 93.6/93.5 & 91.8/91.7 & 91.5/91.4 & 90.9/90.9 & 92.6/92.5 & 93.9/94.3 & 94.7/95.1 & \cellcolor{Top1!30!cyan!40}\textbf{95.0}/\textbf{95.4} & 94.6/95.0 & 94.6/95.0 & 94.8/95.2 & 93.9/94.3 & 60.3/63.3 & 60.8/64.2 & 62.4/65.8 & \cellcolor{Top1!30!cyan!40}\textbf{62.5}/64.9 & 62.3/65.3 & 62.3/65.1 & 61.2/63.7 \\
\cmidrule(lr){1-22}
SheafAN & 88.7/88.6 & 88.3/88.4 & 88.4/88.4 & 88.8/88.7 & 88.2/88.1 & \cellcolor{Top1!30!cyan!40}90.9/90.9 & 88.2/88.2 & 86.3/87.0 & 91.1/91.8 & 82.7/83.7 & 90.9/91.6 & \cellcolor{Top1!30!cyan!40}92.4/93.0 & 91.1/91.9 & 90.6/91.2 & 55.9/59.2 & 54.6/58.7 & 56.1/58.9 & 55.6/59.1 & 56.2/59.2 & \cellcolor{Top1!30!cyan!40}56.4/59.3 & 55.8/60.2 \\
\bottomrule
\end{tabular}}
\end{table}

\begin{table}[!htbp]
\centering
\caption{Graph classification: Macro-F1\,/\,Micro-F1 (\%) by model and number of layers $L$. \colorbox{Top1!30!cyan!40}{Cyan} = best in family, \textbf{bold} = global best per dataset.}
\label{tab:ablation_layers_gc}
 
\vspace{4pt}
\resizebox{\textwidth}{!}{%
\setlength{\tabcolsep}{2pt}
\renewcommand{\arraystretch}{1.1}
\small
\begin{tabular}{l ccccccc ccccccc}
\toprule
& \multicolumn{7}{c}{\textbf{MUTAG}} & \multicolumn{7}{c}{\textbf{PROTEINS}} \\
\cmidrule(lr){2-8} \cmidrule(lr){9-15}
& $L\!=\!2$ & $L\!=\!3$ & $L\!=\!4$ & $L\!=\!5$ & $L\!=\!6$ & $L\!=\!7$ & $L\!=\!8$
& $L\!=\!2$ & $L\!=\!3$ & $L\!=\!4$ & $L\!=\!5$ & $L\!=\!6$ & $L\!=\!7$ & $L\!=\!8$ \\
\midrule
\multicolumn{15}{l}{\textit{Homogeneous baselines}} \\
GCN & 89.6/90.0 & 89.6/90.0 & 89.6/90.0 & 89.6/90.0 & 89.6/90.0 & 89.6/90.0 & 89.6/90.0 & 76.4/80.4 & 71.8/77.7 & 69.5/71.4 & 75.8/78.6 & 74.1/76.8 & 69.0/70.5 & 75.8/78.6 \\
GAT & 89.6/90.0 & 89.6/90.0 & 89.6/90.0 & 89.6/90.0 & 89.6/90.0 & 58.3/60.0 & 84.0/85.0 & 56.2/56.2 & 75.8/78.6 & 71.8/77.7 & 48.8/70.5 & 71.8/77.7 & 74.6/77.7 & 52.2/52.7 \\
GIN & 89.6/90.0 & 94.9/95.0 & 89.9/90.0 & 85.0/85.0 & 85.0/85.0 & 89.6/90.0 & 84.7/85.0 & 70.3/71.4 & 51.3/71.4 & 64.1/69.6 & 70.6/72.3 & {--} & 66.8/67.9 & 62.7/68.8 \\
SAGE & 60.0/70.0 & 60.0/70.0 & 89.6/90.0 & 89.6/90.0 & 89.6/90.0 & 85.0/85.0 & 89.6/90.0 & 69.9/78.6 & 71.8/77.7 & 64.7/75.9 & 64.7/75.9 & 64.7/75.9 & 71.8/77.7 & 74.7/78.6 \\
\midrule
\multicolumn{15}{l}{\textit{Heterogeneous baselines}} \\
R-GCN & 73.3/75.0 & 84.7/85.0 & 79.2/80.0 & 85.0/85.0 & 79.2/80.0 & 84.7/85.0 & 84.7/85.0 & 74.6/77.7 & 74.7/78.6 & 71.8/77.7 & 74.1/76.8 & 71.8/77.7 & 71.8/77.7 & 75.8/78.6 \\
HAN & 71.5/75.0 & 74.4/75.0 & 74.4/75.0 & 64.2/65.0 & 67.0/70.0 & 82.9/85.0 & 73.3/75.0 & {--} & 79.0/83.0 & 76.8/80.4 & 72.9/75.9 & 77.1/80.4 & 75.8/77.7 & 74.6/75.9 \\
HGT & 68.8/70.0 & 84.7/85.0 & 74.4/75.0 & 74.9/75.0 & 84.7/85.0 & 79.2/80.0 & 74.4/75.0 & 74.4/76.8 & 71.3/74.1 & 77.6/81.2 & 24.3/32.1 & 61.6/61.6 & 65.0/65.2 & 24.2/30.4 \\
\midrule
\multicolumn{15}{l}{\textit{HetSheaf (ours)}} \\
Diag & 80.0/80.0 & 80.0/80.0 & 80.0/80.0 & \cellcolor{Top1!30!cyan!40}83.3/83.3 & 80.0/80.0 & 80.0/80.0 & 80.0/80.0 & \cellcolor{Top1!30!cyan!40}73.6/75.9 & 68.2/68.2 & 70.7/75.9 & 62.2/62.5 & 68.9/72.3 & 74.6/75.9 & 67.3/70.5 \\
Bundle & 79.8/80.0 & 84.7/85.0 & 84.7/85.0 & \cellcolor{Top1!30!cyan!40}94.9/95.0 & 84.7/85.0 & 84.7/85.0 & 79.2/80.0 & \cellcolor{Top1!30!cyan!40}78.8/80.4 & 74.8/75.9 & 75.5/76.8 & 50.7/50.9 & 75.8/77.7 & 74.7/76.8 & 69.8/70.5 \\
General & 79.8/80.0 & 76.2/80.0 & 74.9/75.0 & 89.9/90.0 & 73.3/75.0 & \cellcolor{Top1!30!cyan!40}94.9/95.0 & 78.0/80.0 & 33.3/37.5 & 38.0/40.2 & \cellcolor{Top1!30!cyan!40}78.7/81.2 & 64.1/66.1 & 63.2/63.4 & 71.3/74.1 & 64.2/67.0 \\
\cmidrule(lr){1-15}
Diag-P & 89.9/90.0 & 96.0/96.0 & 95.7/95.7 & \cellcolor{Top1!30!cyan!40}\textbf{100.0}/\textbf{100.0} & 94.9/95.0 & 95.7/95.7 & 96.0/96.0 & 40.6/42.9 & 63.2/74.1 & 71.6/74.1 & 41.3/67.9 & 76.4/80.4 & 55.4/55.4 & \cellcolor{Top1!30!cyan!40}78.9/\textbf{82.1} \\
Bundle-P & \cellcolor{Top1!30!cyan!40}\textbf{100.0}/\textbf{100.0} & 94.9/95.0 & \cellcolor{Top1!30!cyan!40}\textbf{100.0}/\textbf{100.0} & \cellcolor{Top1!30!cyan!40}\textbf{100.0}/\textbf{100.0} & 94.9/95.0 & \cellcolor{Top1!30!cyan!40}\textbf{100.0}/\textbf{100.0} & 94.9/95.0 & 74.9/77.7 & \cellcolor{Top1!30!cyan!40}79.8/\textbf{82.1} & 76.9/79.5 & 79.7/81.2 & 78.8/80.4 & 78.4/80.4 & 71.4/78.6 \\
Gen.-P & \cellcolor{Top1!30!cyan!40}\textbf{100.0}/\textbf{100.0} & \cellcolor{Top1!30!cyan!40}\textbf{100.0}/\textbf{100.0} & \cellcolor{Top1!30!cyan!40}\textbf{100.0}/\textbf{100.0} & \cellcolor{Top1!30!cyan!40}\textbf{100.0}/\textbf{100.0} & \cellcolor{Top1!30!cyan!40}\textbf{100.0}/\textbf{100.0} & \cellcolor{Top1!30!cyan!40}\textbf{100.0}/\textbf{100.0} & \cellcolor{Top1!30!cyan!40}\textbf{100.0}/\textbf{100.0} & 77.8/80.4 & 78.1/80.4 & 76.7/77.7 & 73.9/75.9 & 78.7/81.2 & 75.8/78.6 & \cellcolor{Top1!30!cyan!40}\textbf{80.6}/\textbf{82.1} \\
\cmidrule(lr){1-15}
SheafAN & 89.0/90.0 & 79.2/80.0 & 84.0/85.0 & \cellcolor{Top1!30!cyan!40}94.9/95.0 & 89.9/90.0 & \cellcolor{Top1!30!cyan!40}94.9/95.0 & \cellcolor{Top1!30!cyan!40}94.9/95.0 & 73.8/78.6 & 74.7/78.6 & \cellcolor{Top1!30!cyan!40}77.2/80.4 & 76.4/78.6 & 76.7/78.6 & 75.6/77.7 & 75.2/76.8 \\
\bottomrule
\end{tabular}}
 
\vspace{6pt}
\resizebox{\textwidth}{!}{%
\setlength{\tabcolsep}{2pt}
\renewcommand{\arraystretch}{1.1}
\small
\begin{tabular}{l ccccccc ccccccc}
\toprule
& \multicolumn{7}{c}{\textbf{ENZYMES}} & \multicolumn{7}{c}{\textbf{NCI1}} \\
\cmidrule(lr){2-8} \cmidrule(lr){9-15}
& $L\!=\!2$ & $L\!=\!3$ & $L\!=\!4$ & $L\!=\!5$ & $L\!=\!6$ & $L\!=\!7$ & $L\!=\!8$
& $L\!=\!2$ & $L\!=\!3$ & $L\!=\!4$ & $L\!=\!5$ & $L\!=\!6$ & $L\!=\!7$ & $L\!=\!8$ \\
\midrule
\multicolumn{15}{l}{\textit{Homogeneous baselines}} \\
GCN & 11.2/25.0 & 12.7/28.3 & 9.3/23.3 & 11.2/15.0 & 10.9/15.0 & 11.9/18.3 & 9.8/11.7 & 63.2/63.5 & 64.2/65.0 & {--} & 63.2/63.5 & 64.2/65.0 & 63.2/63.5 & 64.2/65.0 \\
GAT & 14.2/16.7 & 5.6/20.0 & 12.0/18.3 & 9.7/20.0 & 5.9/21.7 & 5.9/21.7 & 6.9/18.3 & 64.2/65.0 & 62.0/62.0 & 43.1/50.4 & 35.5/55.0 & 63.2/63.5 & 61.3/65.0 & 35.5/55.0 \\
GIN & 9.6/21.7 & 8.1/11.7 & 9.1/21.7 & 13.4/18.3 & 10.1/18.3 & 16.3/21.7 & 20.2/28.3 & 64.2/65.0 & 60.9/63.5 & 64.2/65.0 & 63.0/64.2 & 62.0/62.0 & 63.2/64.7 & 55.7/57.4 \\
SAGE & 5.6/8.3 & 6.0/21.7 & 8.3/20.0 & 5.6/20.0 & 8.5/18.3 & 5.6/20.0 & 5.9/21.7 & 62.0/62.0 & 63.6/63.7 & 62.0/62.0 & 60.9/61.6 & 63.7/65.5 & 62.9/63.0 & 62.0/62.0 \\
\midrule
\multicolumn{15}{l}{\textit{Heterogeneous baselines}} \\
R-GCN & 7.9/16.7 & 9.0/16.7 & 10.4/16.7 & 9.7/20.0 & 8.6/20.0 & 6.8/10.0 & 5.9/21.7 & 64.8/65.7 & 63.9/64.5 & 62.0/62.0 & 64.2/65.0 & 63.2/63.5 & 58.6/64.7 & 63.2/63.5 \\
HAN & 26.6/28.3 & 41.6/41.7 & {--} & 31.6/33.3 & 46.1/45.0 & 38.3/38.3 & 19.1/25.0 & {--} & 74.7/75.2 & 76.1/76.2 & 78.0/78.1 & 75.4/75.4 & 76.4/76.4 & 75.1/75.2 \\
HGT & {--} & 40.1/40.0 & 36.1/36.7 & 48.3/48.3 & 39.4/43.3 & 9.3/18.3 & 13.4/28.3 & 71.6/71.8 & 67.8/68.1 & 79.3/79.3 & 58.7/63.7 & 65.0/65.0 & 66.3/67.9 & {--} \\
\midrule
\multicolumn{15}{l}{\textit{HetSheaf (ours)}} \\
Diag & 31.0/40.0 & 46.1/46.7 & 34.2/35.0 & 19.3/28.3 & 45.2/45.0 & 48.3/50.0 & \cellcolor{Top1!30!cyan!40}53.0/53.3 & 72.1/72.1 & \cellcolor{Top1!30!cyan!40}78.2/78.2 & 73.7/73.7 & 74.5/74.5 & 68.5/68.5 & 76.8/76.8 & 61.8/61.8 \\
Bundle & 38.2/43.3 & 50.2/53.3 & 38.6/41.7 & 36.6/36.7 & 51.2/53.3 & 22.4/30.0 & \cellcolor{Top1!30!cyan!40}\textbf{62.3}/\textbf{63.3} & \cellcolor{Top1!30!cyan!40}68.0/68.0 & 62.1/62.1 & 59.9/59.9 & 62.8/62.8 & 53.2/53.2 & 60.4/60.4 & 62.3/62.3 \\
General & 31.9/36.7 & 47.7/50.0 & 46.9/46.7 & 43.5/45.0 & \cellcolor{Top1!30!cyan!40}49.9/51.7 & 17.8/28.3 & \cellcolor{Top1!30!cyan!40}50.7/51.7 & 61.7/61.7 & 48.7/48.7 & 64.4/64.4 & 28.1/28.1 & 63.2/63.2 & \cellcolor{Top1!30!cyan!40}67.5/67.5 & 62.1/62.1 \\
\cmidrule(lr){1-15}
Diag-P & 14.4/21.7 & 10.1/21.7 & 11.3/23.3 & 20.6/23.3 & 14.9/21.7 & \cellcolor{Top1!30!cyan!40}25.6/30.0 & 22.7/28.3 & {--} & 61.4/61.4 & \cellcolor{Top1!30!cyan!40}62.1/62.1 & 49.1/49.1 & 23.0/23.0 & 46.6/46.6 & \cellcolor{Top1!30!cyan!40}62.1/62.1 \\
Bundle-P & 46.5/48.3 & \cellcolor{Top1!30!cyan!40}56.7/56.7 & 52.6/53.3 & 51.9/51.7 & 32.4/35.0 & 50.0/50.0 & 34.6/35.0 & 69.2/69.2 & 63.9/63.9 & \cellcolor{Top1!30!cyan!40}76.3/76.3 & 68.6/68.6 & 74.2/74.2 & 73.0/73.0 & 62.1/62.1 \\
Gen.-P & 41.2/45.0 & 41.7/43.3 & \cellcolor{Top1!30!cyan!40}48.1/48.3 & 45.5/46.7 & 37.1/41.7 & 38.0/41.7 & 34.4/36.7 & 67.2/67.2 & 72.3/72.3 & 73.0/73.0 & 72.9/72.9 & 69.9/69.9 & \cellcolor{Top1!30!cyan!40}\textbf{79.5}/\textbf{79.5} & 62.1/62.1 \\
\cmidrule(lr){1-15}
SheafAN & 27.0/30.0 & 34.9/36.7 & \cellcolor{Top1!30!cyan!40}46.5/50.0 & 37.7/41.7 & 38.6/41.7 & \cellcolor{Top1!30!cyan!40}49.3/51.7 & 30.9/35.0 & 62.1/62.1 & 68.2/68.2 & \cellcolor{Top1!30!cyan!40}75.1/75.1 & 69.7/69.7 & 66.5/66.5 & \cellcolor{Top1!30!cyan!40}75.1/75.1 & 73.6/73.6 \\
\bottomrule
\end{tabular}}
\end{table}
 
\begin{table}[!htbp]
\centering
\caption{Link prediction: AUPR\,/\,AUROC (\%) by model and number of layers $L$. \colorbox{Top1!30!cyan!40}{Cyan} = best in family, \textbf{bold} = global best per dataset.}
\label{tab:ablation_layers_lp}
 
\vspace{4pt}
\resizebox{\textwidth}{!}{%
\setlength{\tabcolsep}{2pt}
\renewcommand{\arraystretch}{1.1}
\small
\begin{tabular}{l ccccccc ccccccc}
\toprule
& \multicolumn{7}{c}{\textbf{LastFM}} & \multicolumn{7}{c}{\textbf{MovieLens}} \\
\cmidrule(lr){2-8} \cmidrule(lr){9-15}
& $L\!=\!2$ & $L\!=\!3$ & $L\!=\!4$ & $L\!=\!5$ & $L\!=\!6$ & $L\!=\!7$ & $L\!=\!8$
& $L\!=\!2$ & $L\!=\!3$ & $L\!=\!4$ & $L\!=\!5$ & $L\!=\!6$ & $L\!=\!7$ & $L\!=\!8$ \\
\midrule
\multicolumn{15}{l}{\textit{Homogeneous baselines}} \\
GCN & 96.7/96.5 & 96.5/96.0 & 96.9/96.8 & 96.8/96.8 & 97.1/97.0 & 97.1/96.9 & 97.0/97.0 & 99.5/99.5 & 99.6/99.6 & 99.6/99.6 & 99.6/99.6 & 99.6/99.6 & 99.6/99.6 & 99.6/99.6 \\
GAT & 62.7/50.4 & 62.7/50.4 & 62.7/50.4 & 63.1/51.1 & 62.7/50.4 & 62.7/50.4 & 62.9/50.7 & 97.0/97.4 & 97.6/97.9 & 96.8/97.3 & 96.8/97.3 & 96.8/97.2 & 96.8/97.2 & 96.8/97.2 \\
GIN & 97.4/96.3 & 98.1/97.6 & 98.0/97.7 & 97.9/97.7 & 96.7/96.6 & 97.0/96.9 & 96.9/96.5 & 99.6/99.5 & 99.6/99.6 & 99.6/99.6 & 99.6/99.5 & 99.6/99.6 & 99.3/99.3 & 99.5/99.5 \\
SAGE & 92.7/89.2 & 88.0/86.4 & 83.5/80.1 & 95.1/92.7 & 90.0/86.4 & 88.5/86.5 & 88.0/86.6 & 98.8/98.8 & 98.8/98.8 & 99.1/99.0 & 99.2/99.1 & 99.1/99.0 & 99.2/99.0 & 99.2/99.1 \\
\midrule
\multicolumn{15}{l}{\textit{Heterogeneous baselines}} \\
R-GCN & 97.8/97.8 & 97.7/97.7 & 97.8/97.8 & 98.0/97.9 & 97.9/97.9 & 97.8/97.8 & 97.8/97.9 & 99.4/99.4 & 99.4/99.4 & 99.4/99.4 & 99.3/99.3 & 99.3/99.3 & 99.2/99.2 & 99.3/99.3 \\
HAN & 75.8/73.4 & 86.2/83.0 & 76.7/74.2 & 88.3/83.3 & 86.6/81.8 & 87.5/82.8 & 87.2/82.7 & 63.0/51.1 & 63.0/51.1 & 63.0/51.1 & 63.6/52.2 & 63.1/51.2 & 63.1/51.3 & 63.1/51.4 \\
HGT & 83.8/71.3 & 86.1/74.1 & 85.8/74.5 & 85.9/74.4 & 86.0/74.5 & 86.1/74.5 & 85.9/74.4 & 68.9/61.7 & 70.8/63.6 & 63.8/52.6 & 77.7/74.0 & 67.6/58.9 & 63.0/51.1 & 63.0/51.0 \\
\midrule
\multicolumn{15}{l}{\textit{HetSheaf (ours)}} \\
Diag & 98.4/98.1 & 98.4/98.2 & 98.4/98.2 & \cellcolor{Top1!30!cyan!40}\textbf{98.5}/98.2 & \cellcolor{Top1!30!cyan!40}98.5/98.2 & 98.4/98.2 & 98.4/98.2 & \cellcolor{Top1!30!cyan!40}\textbf{99.7}/99.6 & \cellcolor{Top1!30!cyan!40}\textbf{99.7}/99.6 & \cellcolor{Top1!30!cyan!40}\textbf{99.7}/99.6 & \cellcolor{Top1!30!cyan!40}\textbf{99.7}/99.6 & \cellcolor{Top1!30!cyan!40}\textbf{99.7}/99.6 & \cellcolor{Top1!30!cyan!40}\textbf{99.7}/99.6 & \cellcolor{Top1!30!cyan!40}\textbf{99.7}/99.6 \\
Bundle & 98.4/98.3 & 98.5/98.2 & 98.5/98.2 & 98.5/98.2 & \cellcolor{Top1!30!cyan!40}\textbf{98.5}/\textbf{98.3} & 98.4/98.1 & 98.4/98.2 & \cellcolor{Top1!30!cyan!40}99.7/\textbf{99.6} & 99.7/99.6 & \cellcolor{Top1!30!cyan!40}99.7/99.6 & 99.7/99.6 & \cellcolor{Top1!30!cyan!40}99.7/99.6 & 99.7/99.6 & 99.7/99.6 \\
General & \cellcolor{Top1!30!cyan!40}98.5/98.2 & 98.5/\textbf{98.3} & \cellcolor{Top1!30!cyan!40}\textbf{98.5}/\textbf{98.3} & 98.5/\textbf{98.3} & 98.5/\textbf{98.3} & 98.4/98.2 & 98.4/98.3 & 99.7/99.6 & 99.7/99.6 & \cellcolor{Top1!30!cyan!40}99.7/99.6 & 99.7/99.6 & 99.7/\textbf{99.6} & 99.7/99.6 & \cellcolor{Top1!30!cyan!40}99.7/99.6 \\
\cmidrule(lr){1-15}
Diag-P & \cellcolor{Top1!30!cyan!40}98.4/98.1 & 98.4/98.0 & 98.3/98.0 & 98.3/98.0 & 98.4/98.0 & 98.3/98.1 & 98.2/97.9 & 99.6/99.5 & \cellcolor{Top1!30!cyan!40}99.6/99.6 & 99.6/99.6 & \cellcolor{Top1!30!cyan!40}99.7/99.6 & 99.7/99.6 & 99.7/99.6 & 99.7/99.6 \\
Bundle-P & 98.3/97.8 & 98.3/98.0 & \cellcolor{Top1!30!cyan!40}98.4/98.1 & 98.3/98.1 & 98.2/98.0 & 98.2/97.9 & 98.2/97.8 & 99.7/99.6 & \cellcolor{Top1!30!cyan!40}99.7/99.6 & \cellcolor{Top1!30!cyan!40}99.7/99.6 & 99.7/99.6 & \cellcolor{Top1!30!cyan!40}99.7/99.6 & 99.7/99.6 & 99.7/99.6 \\
Gen.-P & 98.3/98.0 & 98.3/98.0 & \cellcolor{Top1!30!cyan!40}\textbf{98.5}/\textbf{98.3} & 98.3/98.1 & 98.2/97.9 & 98.4/98.1 & 98.3/98.0 & 99.7/99.6 & 99.7/99.6 & 99.7/99.6 & \cellcolor{Top1!30!cyan!40}99.7/99.6 & 99.7/99.6 & 99.7/99.6 & 99.7/99.6 \\
\cmidrule(lr){1-15}
SheafAN & 97.7/97.2 & 97.7/97.3 & 97.6/97.1 & \cellcolor{Top1!30!cyan!40}97.9/97.6 & 97.8/97.2 & 97.7/97.5 & 97.7/97.3 & 99.5/99.4 & 99.5/99.4 & 99.5/99.4 & 99.5/99.4 & 99.5/99.4 & 99.5/99.5 & \cellcolor{Top1!30!cyan!40}99.6/99.5 \\
\bottomrule
\end{tabular}}
 
\vspace{6pt}
\resizebox{\textwidth}{!}{%
\setlength{\tabcolsep}{2pt}
\renewcommand{\arraystretch}{1.1}
\small
\begin{tabular}{l ccccccc ccccccc}
\toprule
& \multicolumn{7}{c}{\textbf{Amazon}} & \multicolumn{7}{c}{\textbf{YouTube}} \\
\cmidrule(lr){2-8} \cmidrule(lr){9-15}
& $L\!=\!2$ & $L\!=\!3$ & $L\!=\!4$ & $L\!=\!5$ & $L\!=\!6$ & $L\!=\!7$ & $L\!=\!8$
& $L\!=\!2$ & $L\!=\!3$ & $L\!=\!4$ & $L\!=\!5$ & $L\!=\!6$ & $L\!=\!7$ & $L\!=\!8$ \\
\midrule
\multicolumn{15}{l}{\textit{Homogeneous baselines}} \\
GCN & 92.3/89.5 & 92.1/89.5 & 92.4/89.6 & 92.1/89.6 & 92.3/89.6 & 92.4/89.7 & 92.2/89.6 & 89.3/83.3 & 89.3/83.2 & 89.3/83.2 & 89.3/83.3 & 89.3/83.3 & 89.3/83.3 & 89.3/83.2 \\
GAT & 88.6/86.0 & 88.7/86.0 & 88.1/85.5 & 88.6/86.1 & 80.3/74.0 & 88.6/86.0 & 85.4/82.1 & 62.5/50.0 & 62.5/50.0 & 66.8/58.4 & 62.5/50.0 & 62.7/50.4 & 62.5/50.0 & 62.5/50.0 \\
GIN & 92.1/89.6 & 92.2/89.3 & 92.3/89.5 & 92.2/89.5 & 92.2/89.2 & 92.2/89.2 & {--} & 88.6/83.0 & 87.6/82.1 & 85.8/81.3 & 88.0/82.2 & 87.0/81.0 & 87.3/82.1 & 85.4/79.7 \\
SAGE & 89.3/86.2 & 90.0/87.1 & 90.8/88.0 & 90.3/87.3 & 89.9/87.0 & 90.4/87.5 & 90.4/87.5 & 71.4/66.2 & 81.4/75.6 & 84.5/80.2 & 83.1/78.9 & 78.5/75.9 & 79.2/76.1 & 85.6/79.2 \\
\midrule
\multicolumn{15}{l}{\textit{Heterogeneous baselines}} \\
R-GCN & 93.4/91.4 & 92.9/91.0 & 92.8/91.1 & 92.8/90.9 & 87.2/83.3 & 92.1/90.1 & 88.6/86.0 & 91.7/88.3 & 91.6/88.2 & 91.7/88.2 & 91.5/88.2 & 91.4/88.1 & 91.4/88.0 & 91.4/88.1 \\
HAN & 94.0/92.3 & 94.9/93.4 & 94.3/92.6 & 91.3/88.7 & 94.1/92.3 & 95.2/93.6 & 94.9/93.1 & 90.3/87.0 & 91.1/86.8 & 93.5/90.1 & 92.2/88.3 & 90.0/85.8 & 90.0/85.2 & 89.1/84.4 \\
HGT & 96.0/95.3 & 96.1/95.6 & 95.6/95.3 & 95.9/95.4 & 96.1/95.7 & 96.2/95.6 & 95.9/95.5 & 94.2/91.5 & 94.6/91.8 & 95.0/92.5 & 94.5/91.6 & 94.2/91.6 & 94.6/91.7 & 94.1/91.6 \\
\midrule
\multicolumn{15}{l}{\textit{HetSheaf (ours)}} \\
Diag & \cellcolor{Top1!30!cyan!40}\textbf{93.9}/91.6 & 93.8/91.5 & 93.8/91.6 & 93.8/91.5 & 93.8/91.5 & 93.9/91.6 & \cellcolor{Top1!30!cyan!40}\textbf{93.9}/91.6 & 89.5/83.5 & \cellcolor{Top1!30!cyan!40}90.6/85.4 & \cellcolor{Top1!30!cyan!40}90.7/86.0 & 90.4/85.5 & 90.5/85.7 & 90.5/85.4 & 89.4/83.4 \\
Bundle & 93.6/91.3 & \cellcolor{Top1!30!cyan!40}93.8/91.6 & 93.8/91.5 & 93.7/91.5 & 93.8/91.5 & 93.8/91.6 & 93.7/91.4 & \cellcolor{Top1!30!cyan!40}\textbf{90.7}/85.9 & 90.3/84.9 & 89.3/83.2 & 90.1/84.4 & 89.5/83.4 & 89.2/83.2 & 89.3/83.4 \\
General & \cellcolor{Top1!30!cyan!40}93.9/\textbf{91.7} & \cellcolor{Top1!30!cyan!40}93.9/91.6 & 93.8/91.6 & 93.8/91.6 & 93.8/91.5 & 93.8/91.5 & 93.8/91.6 & 91.1/86.8 & 90.7/86.0 & \cellcolor{Top1!30!cyan!40}\textbf{91.3}/\textbf{87.3} & 90.4/85.4 & \cellcolor{Top1!30!cyan!40}\textbf{91.3}/\textbf{87.3} & 90.6/86.0 & 90.4/85.6 \\
\cmidrule(lr){1-15}
Diag-P & \cellcolor{Top1!30!cyan!40}93.5/90.9 & 93.3/90.7 & 93.1/90.5 & 93.1/90.3 & 93.2/90.6 & 93.3/90.5 & 93.3/90.8 & 90.6/85.4 & 90.5/85.2 & 90.6/85.8 & \cellcolor{Top1!30!cyan!40}90.7/86.5 & 89.4/83.3 & 90.1/85.3 & 89.4/83.2 \\
Bundle-P & 93.5/91.1 & 93.6/90.9 & 93.5/91.1 & 93.5/91.1 & 93.5/91.1 & \cellcolor{Top1!30!cyan!40}93.6/91.1 & 93.6/91.2 & \cellcolor{Top1!30!cyan!40}89.4/83.3 & 89.4/83.3 & 89.4/83.3 & \cellcolor{Top1!30!cyan!40}89.4/83.4 & 89.4/83.2 & 89.4/83.3 & 89.4/83.3 \\
Gen.-P & 93.0/90.3 & \cellcolor{Top1!30!cyan!40}93.8/91.4 & \cellcolor{Top1!30!cyan!40}\textbf{93.8}/\textbf{91.7} & 93.8/91.5 & 93.6/91.1 & 93.6/91.3 & 92.9/90.3 & {--} & {--} & {--} & {--} & {--} & \cellcolor{Top1!30!cyan!40}89.2/83.0 & {--} \\
\cmidrule(lr){1-15}
SheafAN & 92.0/88.7 & 92.1/88.9 & 91.7/88.4 & 91.8/88.6 & \cellcolor{Top1!30!cyan!40}92.2/89.1 & 91.9/88.7 & 92.0/89.2 & 86.9/80.4 & 87.0/80.5 & 86.7/80.5 & 86.8/80.2 & 87.0/80.5 & 86.6/80.2 & \cellcolor{Top1!30!cyan!40}87.1/80.6 \\
\bottomrule
\end{tabular}}
\end{table}

\subsection{External Comparisons with Additional Heterogeneous GNNs}
\label{app:external_comparisons}

The main paper compares \textsc{HetSheaf} against a controlled set of strong homogeneous, heterogeneous, and sheaf-based baselines trained under a unified codebase and tuning protocol. This section extends this comparison in two ways: first, with a \emph{matched-protocol} re-run of four recent state-of-the-art heterogeneous GNNs; and second, with \emph{contextual} comparisons against a broader set of heterogeneous GNNs reported in recent surveys and benchmarks~\citep{lvAreWeReally2021a,zhu2025hagnn,lin2025adaptivehetero}.

\subsubsection{Matched-Protocol Comparison with Recent State-of-the-Art Heterogeneous GNNs}
\label{app:matched_protocol}
We re-ran HAGNN~\citep{zhu2025hagnn}, HINormer~\citep{mao2023hinormer}, SeHGNN~\citep{yang2023sehgnn}, and SlotGAT~\citep{zhouSlotGATSlotbasedMessage2023} on the HGB node classification benchmarks under a fully matched protocol. All baselines use the authors' official implementations, are trained on the same committed split as our models (canonical HGB 24/6/70), and are scored by the same metric code, with each run asserting agreement with the implementation's own \texttt{sklearn} metrics. For each baseline, we report the best validation-selected configuration from a 100-run random search. \Cref{tab:matched_protocol} reports the results: \textsc{HetSheaf} obtains the best Macro and Micro F1 on ACM and DBLP, while on IMDB SeHGNN and HINormer are stronger, with \textsc{HetSheaf} remaining competitive despite not relying on meta-path or transformer machinery.

\begin{table}[!htbp]
\centering
\caption{%
\emph{Matched-protocol node classification} against recent state-of-the-art heterogeneous GNNs (Macro-F1\,/\,Micro-F1). Protocol as described above; colouring as in \Cref{tab:nc_results}. Our seed-averaged values are in \Cref{tab:nc_results}.%
}
\label{tab:matched_protocol}
\adjustbox{max width=0.7\textwidth}{%
\setlength{\tabcolsep}{4.5pt}
\begin{tabular}{l cc cc cc}
\toprule
& \multicolumn{2}{c}{\textbf{ACM}}
& \multicolumn{2}{c}{\textbf{DBLP}}
& \multicolumn{2}{c}{\textbf{IMDB}} \\
\cmidrule(lr){2-3}\cmidrule(lr){4-5}\cmidrule(lr){6-7}
\textbf{Model}
& Macro-F1 & Micro-F1
& Macro-F1 & Micro-F1
& Macro-F1 & Micro-F1 \\
\midrule
HAGNN
& \textbf{\textcolor{Top3}{92.17\pmstd{0.35}}} & \textbf{\textcolor{Top3}{92.11\pmstd{0.33}}}
& \textbf{\textcolor{Top3}{94.27\pmstd{0.56}}} & \textbf{\textcolor{Top3}{94.69\pmstd{0.51}}}
& \textbf{\textcolor{Top3}{63.32\pmstd{0.94}}} & \textbf{\textcolor{Top2}{67.60\pmstd{0.39}}} \\
SeHGNN
& \textbf{\textcolor{Top2}{92.95\pmstd{0.38}}} & \textbf{\textcolor{Top2}{92.87\pmstd{0.38}}}
& 93.81\pmstd{0.29} & 94.26\pmstd{0.28}
& \textbf{\textcolor{Top1}{66.87\pmstd{0.11}}} & \textbf{\textcolor{Top1}{69.52\pmstd{0.17}}} \\
SlotGAT
& 91.73\pmstd{0.48} & 91.62\pmstd{0.49}
& \textbf{\textcolor{Top2}{94.49\pmstd{0.37}}} & \textbf{\textcolor{Top2}{94.89\pmstd{0.37}}}
& 61.59\pmstd{0.59} & 66.27\pmstd{0.55} \\
\midrule
Sheaf-NSD
& 92.07\pmstd{2.09} & 91.90\pmstd{2.09}
& 92.91\pmstd{1.21} & 93.15\pmstd{1.18}
& 61.94\pmstd{0.34} & 65.10\pmstd{0.20} \\[2pt]
\textbf{HetSheaf (ours)}
& \textbf{\textcolor{Top1}{94.30\pmstd{0.24}}} & \textbf{\textcolor{Top1}{94.22\pmstd{0.24}}}
& \textbf{\textcolor{Top1}{94.93\pmstd{0.36}}} & \textbf{\textcolor{Top1}{95.27\pmstd{0.34}}}
& \textbf{\textcolor{Top2}{64.51\pmstd{0.68}}} & \textbf{\textcolor{Top3}{66.57\pmstd{0.17}}} \\
\bottomrule
\end{tabular}%
}
\end{table}

\subsubsection{Contextual Comparisons from the Literature}
\label{app:contextual_comparisons}
We additionally collect results reported in the literature for a broader set of heterogeneous GNNs. Since these baselines were not re-run under a single implementation and, in some cases, use different train/validation/test splits, these comparisons should be interpreted as contextual rather than primary.

Among the additional baselines, HAGNN~\citep{zhu2025hagnn} is the strongest competitor on DBLP and IMDB under the HGB protocol. Although presented as a hybrid framework, it still relies on a fused meta-path graph in its first aggregation stage and therefore inherits core limitations of meta-path-based methods: it requires expert-selected meta-paths, restricts intratype aggregation to those manually specified relational templates, and remains bounded by how well the chosen meta-paths approximate the true semantics of the graph. By contrast, \textsc{HetSheaf} dispenses with meta-path engineering and models inter-type interactions directly on the original heterogeneous topology through learned restriction maps. As shown in \Cref{tab:node-cls}, this yields competitive or superior performance without any hand-designed meta-path machinery.

Tables~\ref{tab:node-cls} and~\ref{tab:link-pred} collect these broader comparisons for node classification and link prediction, respectively.

\begin{table}[!htbp]
\centering
\caption{%
  \emph{Node classification against results reported in the literature} (Macro-F1\,/\,Micro-F1); mean $\pm$ std as given by each source, colouring as in \Cref{tab:nc_results}, \enquote{--} unavailable.
  Sources: $^\dagger$\,\citet{lvAreWeReally2021a} and $^\ddagger$\,\citet{zhu2025hagnn}, both HGB split (24/6/70\,\%);
  $^\S$\,\citet{lin2025adaptivehetero}, 20\,\% train split, so \emph{not} directly comparable and reported without std.%
}
\label{tab:node-cls}
\adjustbox{max width=\textwidth}{%
\setlength{\tabcolsep}{4.5pt}
\begin{tabular}{l cc cc cc}
\toprule
 & \multicolumn{2}{c}{\textbf{DBLP}}
 & \multicolumn{2}{c}{\textbf{IMDB}}
 & \multicolumn{2}{c}{\textbf{ACM}} \\
\cmidrule(lr){2-3}\cmidrule(lr){4-5}\cmidrule(lr){6-7}
\textbf{Model}
 & Macro-F1 & Micro-F1
 & Macro-F1 & Micro-F1
 & Macro-F1 & Micro-F1 \\
\midrule
GTN$^\ddagger$
 & 65.21\pmstd{1.16} & 66.23\pmstd{1.23}
 & 59.26\pmstd{0.84} & 64.07\pmstd{0.65}
 & 91.31\pmstd{0.70}$^\dagger$ & 91.20\pmstd{0.71}$^\dagger$ \\
HetSANN$^\ddagger$
 & 84.08\pmstd{1.01} & 84.96\pmstd{0.88}
 & 49.25\pmstd{0.57} & 57.47\pmstd{1.12}
 & 90.02\pmstd{0.35}$^\dagger$ & 89.91\pmstd{0.37}$^\dagger$ \\
HetGNN$^\ddagger$
 & 92.77\pmstd{0.24} & 93.23\pmstd{0.23}
 & 47.87\pmstd{0.33} & 50.83\pmstd{0.26}
 & 85.91\pmstd{0.25}$^\dagger$ & 86.05\pmstd{0.25}$^\dagger$ \\
CKD$^\ddagger$
 & 92.52\pmstd{0.23} & 92.80\pmstd{0.22}
 & 60.30\pmstd{0.77} & 65.98\pmstd{0.87}
 & {--} & {--} \\
HGNN-AC$^\ddagger$
 & 92.97\pmstd{0.72} & 93.43\pmstd{0.96}
 & 56.63\pmstd{0.81} & 63.85\pmstd{0.85}
 & {--} & {--} \\
RSHN$^\dagger$
 & 93.34\pmstd{0.58} & 93.81\pmstd{0.55}
 & 59.85\pmstd{3.21} & 64.22\pmstd{1.03}
 & 90.50\pmstd{1.51} & 90.32\pmstd{1.54} \\
MAGNN$^\ddagger$
 & 93.16\pmstd{0.38} & 93.65\pmstd{0.34}
 & 56.92\pmstd{1.76} & 65.11\pmstd{0.59}
 & 90.88\pmstd{0.64}$^\dagger$ & 90.77\pmstd{0.65}$^\dagger$ \\
R-HGNN$^\ddagger$
 & 93.31\pmstd{0.42} & 93.80\pmstd{0.39}
 & 61.39\pmstd{0.51} & 66.03\pmstd{0.65}
 & {--} & {--} \\
AMHGNN$^\ddagger$
 & 93.71\pmstd{0.79} & 94.08\pmstd{0.76}
 & 63.38\pmstd{0.66} & 67.29\pmstd{0.64}
 & {--} & {--} \\
SimpleHGN$^\ddagger$
 & 93.81\pmstd{0.54} & 94.26\pmstd{0.49}
 & 63.53\pmstd{1.66} & 67.42\pmstd{0.42}
 & \textbf{\textcolor{Top2}{93.42\pmstd{0.44}}}$^\dagger$
 & \textbf{\textcolor{Top2}{93.35\pmstd{0.45}}}$^\dagger$ \\
BPHGNN$^\ddagger$
 & 93.89\pmstd{0.43} & 94.50\pmstd{0.47}
 & 64.01\pmstd{0.56} & \textbf{\textcolor{Top3}{67.77\pmstd{0.70}}}
 & {--} & {--} \\
HINormer$^\ddagger$
 & \textbf{\textcolor{Top3}{94.57\pmstd{0.21}}}
 & \textbf{\textcolor{Top3}{94.94\pmstd{0.23}}}
 & \textbf{\textcolor{Top2}{64.65\pmstd{0.53}}}
 & \textbf{\textcolor{Top2}{67.83\pmstd{0.34}}}
 & {--} & {--} \\
HAGNN$^\ddagger$
 & \textbf{\textcolor{Top1}{95.06\pmstd{0.31}}}
 & \textbf{\textcolor{Top1}{95.40\pmstd{0.30}}}
 & \textbf{\textcolor{Top1}{65.57\pmstd{0.49}}}
 & \textbf{\textcolor{Top1}{68.62\pmstd{0.23}}}
 & {--} & {--} \\
\midrule
MRGCN$^\S$
 & 89.53 & 90.51
 & 45.62 & 47.78
 & 87.64 & 87.52 \\
SSDCM$^\S$
 & 89.44 & 89.94
 & 49.45 & 59.16
 & 87.72 & 87.65 \\
MHGCN$^\S$
 & 90.90 & 92.17
 & 50.54 & 64.20
 & 88.96 & 89.19 \\
HGATE$^\S$
 & 90.25 & 91.15
 & 41.99 & 52.10
 & 88.73 & 88.70 \\
MGDCR$^\S$
 & 92.04 & 92.43
 & 45.03 & 59.05
 & 89.03 & 88.87 \\
HMGE$^\S$
 & 91.56 & 92.35
 & 32.93 & 57.35
 & 90.80 & 90.66 \\
AMOGCN$^\S$
 & 92.02 & 92.46
 & 50.92 & 63.60
 & 91.72 & 91.39 \\
HGNN-AR$^{2\,\S}$
 & 92.54 & 93.02
 & 52.51 & 61.36
 & \textbf{\textcolor{Top3}{92.42}} & \textbf{\textcolor{Top3}{92.36}} \\
\midrule
Sheaf-NSD
 & 92.91\pmstd{1.21} & 93.15\pmstd{1.18}
 & 61.94\pmstd{0.34} & 65.10\pmstd{0.20}
 & 92.07\pmstd{2.09} & 91.90\pmstd{2.09} \\[2pt]
\textbf{HetSheaf\,(ours)}
 & \textbf{\textcolor{Top2}{94.93\pmstd{0.36}}}
 & \textbf{\textcolor{Top2}{95.27\pmstd{0.34}}}
 & \textbf{\textcolor{Top3}{64.51\pmstd{0.68}}}
 & 66.57\pmstd{0.17}
 & \textbf{\textcolor{Top1}{94.30\pmstd{0.24}}}
 & \textbf{\textcolor{Top1}{94.22\pmstd{0.24}}} \\
\bottomrule
\end{tabular}%
}
\end{table}

\begin{table}[!htbp]
\centering
\caption{%
  \emph{Link prediction against results reported in the literature} (AUROC); mean $\pm$ std as given by each source, colouring as in \Cref{tab:nc_results}, \enquote{OOM} out of memory, \enquote{--} unavailable.
  Sources: $^\dagger$\,\citet{lvAreWeReally2021a}; $^\ddagger$\,\citet{zhu2025hagnn}.%
}
\label{tab:link-pred}
\adjustbox{max width=0.7\textwidth}{%
\setlength{\tabcolsep}{5pt}
\begin{tabular}{l ccc}
\toprule
\textbf{Model}
 & \textbf{Amazon} & \textbf{LastFM} & \textbf{PubMed} \\
\midrule
MAGNN$^\ddagger$
 & {--} & 56.81\pmstd{0.05} & OOM \\
HetGNN$^\dagger$
 & 77.74\pmstd{0.24} & 62.09\pmstd{0.01} & 73.63\pmstd{0.01} \\
GATNE$^\dagger$
 & 77.39\pmstd{0.50} & 66.87\pmstd{0.16} & 63.39\pmstd{0.65} \\
AutoAC$^\ddagger$
 & {--} & 66.64\pmstd{0.26} & 82.95\pmstd{0.38} \\
SimpleHGN$^\dagger$
 & \textbf{\textcolor{Top1}{93.40\pmstd{0.62}}}
 & 67.16\pmstd{0.37}$^\ddagger$
 & 83.39\pmstd{0.39}$^\ddagger$ \\
HAGNN$^\ddagger$
 & {--}
 & \textbf{\textcolor{Top3}{67.33\pmstd{0.10}}}
 & \textbf{\textcolor{Top3}{84.00\pmstd{0.63}}} \\
\midrule
Sheaf-NSD
 & \textbf{\textcolor{Top3}{90.02\pmstd{0.25}}}
 & \textbf{\textcolor{Top2}{96.58\pmstd{0.17}}}
 & \textbf{\textcolor{Top2}{93.69\pmstd{0.16}}} \\[2pt]
\textbf{HetSheaf\,(ours)}
 & \textbf{\textcolor{Top2}{91.71\pmstd{0.37}}}
 & \textbf{\textcolor{Top1}{98.33\pmstd{0.35}}}
 & \textbf{\textcolor{Top1}{93.82\pmstd{0.68}}} \\
\bottomrule
\end{tabular}%
}
\end{table}

\subsection{Implementation Details}
\label{app:implementation_details}

We now collect the implementation details underlying all experiments. We begin with the hardware and software environment, then describe the downstream decoders and loss functions used for the three tasks, followed by the parameterisation of restriction maps and the full family of heterogeneous sheaf predictors. We then report the computational complexity and overhead, and detail the heterogeneous extensions of Sheaf Attention Networks and Polynomial Neural Sheaf Diffusion used in our experiments.

\subsubsection{Hardware and Software Setup}
\label{app:hardware_setup}
All experiments were conducted on an AWS EC2 g6.xlarge instance running Ubuntu 24.04 LTS, equipped with an NVIDIA L4 GPU (CUDA 12.8, cuDNN 9.1). The codebase is implemented in Python 3.13.2 using PyTorch 2.8 and PyTorch Geometric 2.7~\citep{feyFastGraphRepresentation2019, Fey2025PyG}, with dependency management handled by \texttt{uv}\footnote{\url{https://docs.astral.sh/uv/}}. Hyperparameter searches were performed via Weights \& Biases\footnote{\url{https://wandb.ai}} sweeps. These choices were kept fixed across models to ensure that empirical differences reflect modelling choices rather than software-stack variation.

\subsubsection{Downstream Decoders and Loss Functions}
The sheaf encoder produces node-level hidden states that are then consumed by a task-specific decoder. Since node classification, link prediction, and graph classification place different demands on the output space, we use standard downstream decoders tailored to each task while leaving the sheaf backbone unchanged.

\paragraph{Node classification.}
After the Sheaf GNN processes the graph, the final node embeddings are passed to an MLP to produce classification logits. For multiclass tasks, we use a softmax activation with cross-entropy loss; for multilabel settings, we use a sigmoid activation with binary cross-entropy loss.

\paragraph{Link prediction.}
For each node pair $u,v$ and relation type $r$, the model outputs the score
\begin{equation}
    P(u \sim_{r} v) = Decoder\parens*{\rmH_u, \rmH_v}
\end{equation}
which gives the probability that $u$ and $v$ are connected by a relation of type $r$. Since this is a binary prediction problem, the loss is binary cross-entropy.

The decoder can take several forms. A dot-product decoder first computes the inner product and then applies a sigmoid:
\begin{equation}
    Decoder\parens*{\rmH_u, \rmH_v}
    =
    \sigma\parens*{\kets*{\rmH_u, \rmH_v}}.
\end{equation}
DistMult~\citep{yangEmbeddingEntitiesRelations2015} instead computes the bilinear score
\begin{equation}
    Decoder\parens*{\rmH_u, \rmH_v}
    =
    \sigma\parens*{\rmH_{u}^{\intercal}\rmW_{\psi(u, v)}\rmH_{v}},
\end{equation}
where $\rmW_{\psi(u, v)} \in \sR^{d_L \times d_L}$ is a learnable weight matrix for each edge type. Finally, a concat decoder concatenates the two embeddings, applies a one-layer MLP, and then a sigmoid:
\begin{equation}
    Decoder\parens*{\rmH_u, \rmH_v}
    =
    \sigma\parens*{\MLP\parens*{\rmH_u \| \rmH_v}}.
\end{equation}
Following \citet{lvAreWeReally2021a}, we treat the choice of decoder as a hyperparameter.

\subsubsection{Restriction Map Parametrisations}
\label{app:restriction_maps_impl}
Each restriction map is represented as a $d\times d$ block matrix and parametrised by the matrix-valued function $\Phi(\rvx_u, \rvx_v, \phi(u), \phi(v), \psi(e))$. We briefly describe how each family is constructed.

\noindent\textbf{Diagonal restriction maps.}
The predictor outputs $d$ values, which define the diagonal entries of the restriction map.

\noindent\textbf{General restriction maps.}
The predictor outputs $d^2$ values, which are reshaped into a full $d\times d$ matrix.

\noindent\textbf{Orthogonal restriction maps.}
For orthogonal restriction maps $\gF_{\cdot \trianglelefteq \cdot} \in \mathrm{O}(d)$, we consider two parameterisations. The first uses Householder products: the MLP outputs $\lfloor\frac{d(d-1)}{2}\rfloor$ parameters, which are converted by the Torch Householder library~\citep{obukhov2021torchhouseholder} into a product of Householder reflections covering the full orthogonal group $\mathrm{O}(d)$. The second uses the Cayley transform: the MLP outputs $d^2$ values forming a matrix $\gF'$, from which the restriction map is obtained as
\begin{equation}
    \gF_{\cdot \trianglelefteq \cdot} =
    (\rmI_d - \rmS)(\rmI_d + \rmS)^{-1},
    \qquad
    \rmS = \tfrac{1}{2}(\gF' - \gF'^\top).
    \label{eq:cayley}
\end{equation}
Since the Cayley transform maps skew-symmetric matrices to $\mathrm{SO}(d)$, it can represent rotations but not reflections. In practice, the Householder and Cayley parameterisations provide different geometric biases, and the latter is the default for the attention predictors and SheafAN variant following \citet{barberoSheafAttentionNetworks2022}.

\subsubsection{Full Description of Heterogeneous Sheaf Predictors}
\label{app:hetero-sheaf-predictors}

We next give the complete definition of the heterogeneous predictor family used in the experiments. These variants differ in two ways: first, in \emph{which} typed signals are provided to the predictor (node features, node types, edge types, or combinations thereof), and second, in \emph{how} this information is routed (shared predictor versus edge-type-specific ensemble). This family allows us to disentangle the contribution of feature-level conditioning from that of pure structural-type information.

Let $\gG$ be a heterogeneous graph with node features $\rvx_u \in \sR^f$, one-hot node-type vector $\rve_{\phi(u)}$, and one-hot edge-type vector $\rve_{\psi(e)}$ for edge $e \coloneq (u,v)$ and $\rvz_u = \mathcal P_{\phi(u)}(\rvx_u), \;\rvz_v = \mathcal P_{\phi(v)}(\rvx_v)$ the projections in the shared channel. All predictors below are compatible with diagonal, orthogonal, and general restriction-map types.

\begin{table}[!htbp]
\centering
\caption{Summary of Heterogeneous Sheaf Predictors and their inputs.}
\label{tab:sheaf-predictors}
\begin{tabular}{@{}lll@{}}
\toprule
\textbf{Predictor Name} & \textbf{Acronym Meaning} & \textbf{Inputs Used} \\
\midrule
\textbf{Sheaf-NSD} & \textbf{N}eural \textbf{S}heaf \textbf{D}iffusion & $\rvz$ \\
\addlinespace
\textbf{HetSheaf-TE} & \textbf{T}ype \textbf{E}mbeddings & $\rvz, \phi, \psi$ \\
\textbf{HetSheaf-NE} & \textbf{N}ode \textbf{E}mbeddings & $\rvz, \phi$ \\
\textbf{HetSheaf-EE} & \textbf{E}dge \textbf{E}mbeddings & $\rvz, \psi$ \\
\addlinespace
\textbf{HetSheaf-types}& All \textbf{types} only & $\phi, \psi$ \\
\textbf{HetSheaf-NT} & \textbf{N}ode \textbf{T}ypes only & $\phi$ \\
\textbf{HetSheaf-ET} & \textbf{E}dge \textbf{T}ypes only & $\psi$ \\
\addlinespace
\textbf{HetSheaf-ensemble} & \textbf{Ensemble} of MLPs & $\rvz$ (via $\text{MLP}_{\psi(e)}$) \\
\bottomrule
\end{tabular}
\end{table}

The predictors are broadly categorised by whether they append type embeddings to node features, discard features in favour of pure type signals, or use specialised routing through edge-type-specific experts.

\noindent\textbf{Sheaf-NSD.}
\begin{equation}
    \gF_{u \trianglelefteq e} \coloneq \MLP(\rvz_u \| \rvz_v).
\end{equation}

\noindent\textbf{HetSheaf-TE.}
\begin{equation}
    \gF_{u \trianglelefteq e} \coloneq
    \MLP\parens*{\rvz_u \| \rvz_v \| \rve_{\phi(u)} \| \rve_{\phi(v)} \| \rve_{\psi(e)}}.
\end{equation}

\noindent\textbf{HetSheaf-NE.}
\begin{equation}
    \gF_{u \trianglelefteq e} \coloneq
    \MLP\parens*{\rvz_u \| \rvz_v \| \rve_{\phi(u)} \| \rve_{\phi(v)}}.
\end{equation}

\noindent\textbf{HetSheaf-EE.}
\begin{equation}
    \gF_{u \trianglelefteq e} \coloneq
    \MLP\parens*{\rvz_u \| \rvz_v \| \rve_{\psi(e)}}.
\end{equation}

\noindent\textbf{HetSheaf-types.}
\begin{equation}
    \gF_{u \trianglelefteq e} \coloneq
    \MLP\parens*{\rve_{\phi(u)} \| \rve_{\phi(v)} \| \rve_{\psi(e)}}.
\end{equation}

\noindent\textbf{HetSheaf-NT.}
\begin{equation}
    \gF_{u \trianglelefteq e} \coloneq
    \MLP\parens*{\rve_{\phi(u)} \| \rve_{\phi(v)}}.
\end{equation}

\noindent\textbf{HetSheaf-ET.}
\begin{equation}
    \gF_{u \trianglelefteq e} \coloneq \MLP\parens*{\rve_{\psi(e)}}.
\end{equation}

\noindent\textbf{HetSheaf-ensemble.}
\begin{equation}
    \gF_{u \trianglelefteq e} \coloneq
    \MLP_{\psi(e)}\parens*{\rvz_u \| \rvz_v},
\end{equation}
where $\psi(e)$ selects the edge-type-specific predictor. This yields strict expressivity gains over a single shared MLP at the cost of parameters that scale linearly with the number of edge types.

\subsubsection{\textsc{HetSheaf} Computational Complexity and Overhead}
\label{app:computational-overhead}

We derive the per-layer computational complexity of each \textsc{HetSheaf} predictor following the analysis framework of \citet{bodnarNeuralSheafDiffusion2022}. The aim is twofold: first, to make explicit how type conditioning changes the asymptotic cost relative to vanilla NSD; and second, to quantify the practical overhead of the sheaf-based approach relative to strong non-sheaf baselines.

Consider a heterogeneous graph with $n$ nodes, $m$ edges, $s$ node types, and $t$ edge types. Let $d$ denote the stalk dimension and $f$ the number of feature channels, so that $c = d \times f$. We distinguish between diagonal restriction maps, which require only $d$ parameters per map, and non-diagonal maps, which require $d^2$ parameters and may incur an additional $\mathcal{O}(d^3)$ orthogonalisation cost in the orthogonal case. \Cref{tab:complexity} summarises the resulting per-layer complexities.

\begin{table}[!htbp]
    \centering
    \caption{
        \textbf{Per-layer computational complexities of heterogeneous sheaf
        predictors and baselines.}
        Notation: $n$ nodes, $m$ edges, $s$ node types, $t$ edge types,
        stalk dimension $d$, channels $f$, $c = df$ (or $c = f$ for
        non-sheaf models), attention heads $h$.
        Diagonal restriction maps are denoted (diag); general or
        orthogonal maps are denoted (non-diag).
    }
    \label{tab:complexity}
    \renewcommand{\arraystretch}{1.15}
    \setlength{\tabcolsep}{6pt}
    \resizebox{0.4\textwidth}{!}{\begin{tabular}{@{}l l@{}}
        \toprule
        \textbf{Model} & \textbf{Computational complexity} \\
        \midrule
        \multicolumn{2}{@{}l}{\textit{Non-sheaf baselines}} \\[2pt]
        GCN
            & $\mathcal{O}(nc^2 + mc)$ \\
        GAT
            & $\mathcal{O}(nc^2h + mch)$ \\
        HAN
            & $\mathcal{O}(tnc^2h + tmch)$ \\
        \midrule
        \multicolumn{2}{@{}l}{\textit{HetSheaf predictors (diag)}} \\[2pt]
        NSD
            & $\mathcal{O}(nc^2 + mdc)$ \\
        TE
            & $\mathcal{O}(nc^2 + md(c + s + t))$ \\
        NE
            & $\mathcal{O}(nc^2 + md(c + s))$ \\
        EE
            & $\mathcal{O}(nc^2 + md(c + t))$ \\
        types
            & $\mathcal{O}(nc^2 + md(s + t))$ \\
        NT
            & $\mathcal{O}(nc^2 + mds)$ \\
        ET
            & $\mathcal{O}(nc^2 + mdt)$ \\
        ensemble
            & $\mathcal{O}(nc^2 + mdct)$ \\
        \midrule
        \multicolumn{2}{@{}l}{\textit{HetSheaf predictors (non-diag)}} \\[2pt]
        NSD
            & $\mathcal{O}(n(c^2 + d^3) + md^2(c + d))$ \\
        TE
            & $\mathcal{O}(n(c^2 + d^3) + md^2(c + s + t + d))$ \\
        NE
            & $\mathcal{O}(n(c^2 + d^3) + md^2(c + s + d))$ \\
        EE
            & $\mathcal{O}(n(c^2 + d^3) + md^2(c + t + d))$ \\
        types
            & $\mathcal{O}(n(c^2 + d^3) + md^2(s + t))$ \\
        NT
            & $\mathcal{O}(n(c^2 + d^3) + md^2 s)$ \\
        ET
            & $\mathcal{O}(n(c^2 + d^3) + md^2 t)$ \\
        ensemble
            & $\mathcal{O}(n(c^2 + d^3) + md^2(c + dt))$ \\
        \bottomrule
    \end{tabular}}
\end{table}

Several observations follow. First, the additive cost introduced by type-aware predictors relative to NSD is governed by the number of appended type indicators, and is therefore typically negligible since heterogeneous benchmarks usually satisfy $s,t \ll c$. Second, diagonal maps scale linearly in $d$ per edge, while general or orthogonal maps scale quadratically in $d$ because of the full matrix-valued transport. Third, the ensemble predictor is the only variant whose cost scales multiplicatively with the number of edge types, reflecting the use of independent edge-type-specific MLPs. Even so, for the small stalk dimensions and modest type cardinalities used in this work, the practical overhead remains limited. Finally, regarding scalability with the graph size: for fixed stalk dimension $d$ and channel size $f$, every diagonal \textsc{HetSheaf} predictor in \Cref{tab:complexity} runs in $\mathcal{O}\!\big(nc^2 + md(c+s+t)\big) = \mathcal{O}(n+m)$ time per layer, where $n$ and $m$ are the numbers of nodes and edges and $s$ and $t$ the numbers of node and edge types. The type conditioning we introduce therefore enters only as an \emph{additive edge-wise cost} governed by $s+t$, rather than as a factor that grows with the graph size. While the largest benchmark evaluated in this work contains roughly 63K nodes, this linear scaling makes the diagonal variants natural candidates for much larger graphs; an empirical validation at million-node scale is left to future work.

To complement the asymptotic analysis, \Cref{tab:nc_overhead} reports parameter counts and wall-clock costs on the node classification benchmarks. Despite not yet using the most heavily optimised PyG kernels, the sheaf-based models remain substantially smaller than the strongest heterogeneous baselines in terms of parameters, while often matching or surpassing them in predictive performance.

\begin{table}[htbp]
    \centering
    \caption{
        \textbf{\textsc{HetSheaf} computational overhead on node
        classification tasks.}
        The highest performing heterogeneous baseline architectures are
        \colorbox{Top1!30!cyan!40}{highlighted} with the highest performing
        \textsc{HetSheaf} model shown in \textbf{bold}.
    }
    \label{tab:nc_overhead}
    \maxsizebox{\textwidth}{!}{
	\begin{tabular}{lScScSc}
		\toprule
		{}                                                    & \multicolumn{2}{c}{\thead{ACM}}       & \multicolumn{2}{c}{\thead{DBLP}}    & \multicolumn{2}{c}{\thead{IMDB}}                                                                                                                          \\
		\cmidrule(lr){2-3}\cmidrule(lr){4-5}\cmidrule(lr){6-7}
		{}                                                    & {\thead{Runtime (\unit{\second})}}    & {\thead{\# params}}                 & {\thead{Runtime (\unit{\second})}}   & {\thead{\# params}}                  & {\thead{Runtime (\unit{\second})}}   & {\thead{\# params}}                  \\
		\midrule
		R-GCN                                                 & 22.1                                  & 43M                                 & \cellcolor{Top1!30!cyan!40} 30.3                                 & \cellcolor{Top1!30!cyan!40} 87.8M                                & \cellcolor{Top1!30!cyan!40} 25.9                                 &  \cellcolor{Top1!30!cyan!40} 72.1M                                \\
		HAN                                                   & \cellcolor{Top1!30!cyan!40} 28.9            & \cellcolor{Top1!30!cyan!40} 2.2M          & 18.5                                 & 1.9M                                 & 15.8                                 & 3.4M                                 \\
		HGT                                                   & 1025.8                                & 7.2M                                & 648.3                                & 6.2M                                 & 939.9                                & 10.5M                                \\
		\midrule
		Sheaf-NSD                                             & 47.6                                  & 155K                                & 73.3                                 & 1.7M                                 & 40.5                                 & 1.6M                                 \\
		\midrule
		\textbf{HetSheaf-TE (ours)}                           & 44.7                                  & 155K                                & \bfseries 79.1                       & \bfseries 1.7M                       & 40.1                                 & 1.6M                                 \\
		\textbf{HetSheaf-ensemble (ours)}                     & 55.7                                  & 174K                                & 83.2                                 & 1.8M                                 & 44.9                                 & 1.9M                                 \\
		\textbf{HetSheaf-NE (ours)}                           & 47.3                                  & 155K                                & 89.6                                 & 1.7M                                 & 12.0                                 & 1.1M                                 \\
		\textbf{HetSheaf-EE (ours)}                           & \bfseries 49.6                        & \bfseries 155K                       & 85.9                                 & 1.7M                                 & 41.8                                 & 1.6M                                 \\
		\textbf{HetSheaf-NT (ours)}                           & 42.0                                  & 154K                                & 82.4                                 & 1.7M                                 & \bfseries 38.6                       & \bfseries 1.6M                       \\
		\textbf{HetSheaf-ET (ours)}                           & 45.1                                  & 154K                                & 80.5                                 & 1.7M                                 & 39.8                                 & 1.6M                                 \\
		\bottomrule
	\end{tabular}
}
\end{table}

\subsubsection{Heterogeneous Sheaf Attention Networks}
\label{app:sheafan}

Sheaf Attention Networks (SheafANs)~\citep{barberoSheafAttentionNetworks2022} replace Laplacian-based propagation with attention-weighted aggregation over sheaf transports. Since \textsc{HetSheaf} is intended to be model-agnostic, it is important to verify that the inferred heterogeneous sheaf can be consumed not only by NSD-style diffusion layers but also by attention-based sheaf architectures. This subsection, therefore, describes how SheafAN is extended to the heterogeneous setting.

Their geometric backbone is the sheaf adjacency operator $\hat{A}_{\gF} \in \sR^{nd \times nd}$, whose off-diagonal blocks are defined as
\begin{equation}
    P_{ij} = \gF_{i \trianglelefteq e}^{\top}\gF_{j \trianglelefteq e} \in \sR^{d\times d},
\end{equation}
for $e=(i,j)$. In the original homogeneous formulation, the scalar attention coefficients are
\begin{equation}
    \Lambda_{ij}
    =
    \frac{
        \exp\!\left(
            \mathrm{LeakyReLU}\!\left(
                \mathbf{a}^{\top}[\rmW\rvx_i \| \rmW\rvx_j]
            \right)
        \right)
    }{
        \sum_{k \in \mathcal{N}_i}
        \exp\!\left(
            \mathrm{LeakyReLU}\!\left(
                \mathbf{a}^{\top}[\rmW\rvx_i \| \rmW\rvx_k]
            \right)
        \right)
    }.
\end{equation}
To adapt SheafAN to heterogeneous graphs, we  consider the channel projection $\rvz_* = \mathcal P_{\phi(*)}(\rvx_*)$ and augment the attention input with node- and edge-type embeddings:
\begin{equation}
    \Lambda_{ij} =
    \frac{
        \exp\!\left(
            \mathrm{LeakyReLU}\!\left(
                \rva^\top[
                    \rmW\rvz_i \| \rmW\rvz_j \|
                    \rve_{\phi(i)} \| \rve_{\phi(j)} \| \rve_{\psi((i,j))}
                ]
            \right)
        \right)
    }{
        \sum_{k \in \mathcal{N}_i}
        \exp\!\left(
            \mathrm{LeakyReLU}\!\left(
                \rva^\top[
                    \rmW\rvz_i \| \rmW\rvz_k \|
                    \rve_{\phi(i)} \| \rve_{\phi(k)} \| \rve_{\psi((i,k))}
                ]
            \right)
        \right)
    }.
\end{equation}
At the same time, the transport maps $P_{ij}$ are computed from the heterogeneous restriction maps inferred by \textsc{HetSheaf}. Hence, both geometric transport and attention become aware of the typed structure of the graph, while the overall attentive sheaf architecture itself remains unchanged.

\subsubsection{Heterogeneous Polynomial Neural Sheaf Diffusion}
\label{app:polynsd}

Polynomial Neural Sheaf Diffusion (PolyNSD)~\citep{borgi2026polynomialneuralsheafdiffusion} extends first-order sheaf diffusion by replacing repeated local propagation with a learnable spectral polynomial of the sheaf Laplacian. Since PolyNSD is structurally quite different from NSD, its heterogeneous extension provides a strong demonstration of the model-agnosticity of \textsc{HetSheaf}.

For a sheaf Laplacian $L_{\gF}$ with spectral decomposition $L_{\gF} = U\Lambda U^\top$, a polynomial filter $p(L_{\gF}) = \sum_{k=0}^K c_k L_{\gF}^k$ acts diagonally in the sheaf Fourier basis, scaling the $i$-th mode by $p(\lambda_i)$. PolyNSD uses a Chebyshev parameterisation on a rescaled operator
\begin{equation}
    \tilde{L}_{\gF} = \frac{2}{\lambda_{\max}} L_{\gF} - \rmI,
\end{equation}
so that the polynomial remains numerically stable on $[-1,1]$. The filter is then written as
\begin{equation}
    p(\tilde{L}_{\gF}) = \sum_{k=0}^K \alpha_k T_k(\tilde{L}_{\gF}),
    \qquad \{\alpha_k\} = \mathrm{softmax}(\boldsymbol{\eta})
\end{equation}
with the Chebyshev basis evaluated through the usual three-term recurrence.

Our framework extends PolyNSD to heterogeneous graphs without altering this spectral machinery. The only difference is that the Laplacian is now built from a heterogeneous sheaf whose restriction maps are predicted by
\begin{equation}
    \gF_{u \trianglelefteq e\coloneq (u,v)}
    =
    \boldsymbol{\Phi}(\rvx_u,\rvx_v,\phi(u),\phi(v),\psi(e))
\end{equation}
The resulting heterogeneous sheaf Laplacian $L_{\gF}$ is then used directly inside the polynomial filter. In this way, the same higher-order spectral diffusion mechanism extends naturally to the heterogeneous setting, again illustrating that once heterogeneity is absorbed into the inferred sheaf, the downstream sheaf architecture itself does not require ad hoc redesign.

\section{Limitations}
\label{sec:limitations}
While \textsc{HetSheaf} provides a robust topological framework for learning on heterogeneous graphs, it possesses a few notable limitations. First, the computational complexity of the learned restriction maps scales quadratically with their dimension $d$ when using general or bundle parametrisations. Furthermore, these stalks have a fixed dimension $d$ across the entire network for all node and edge types, which may restrict the model's capacity to fully capture the varying complexities of a heterogeneous environment, constituting a well-known limitation of the Sheaf Laplacian. Second, the theoretical guarantees of the \textsc{SheafPool} gauge canonicalisation rely on the assumption that node embeddings possess full row rank, while in practice, network layers may induce rank-deficient stalks, necessitating a regularisation term  $\epsilon I_d$. Consequently, the whitening operation acts as a stable approximation rather than an exact $O(d)$ canonicalisation.


\subsection{Broader Impact}
\label{sec:broader impact}

\textsc{HetSheaf} and \textsc{SheafPool} provide a general framework for learning on heterogeneous relational data by representing type-dependent structure through cellular sheaves rather than through increasingly specialised architectures. This may have positive impact in domains where heterogeneous graphs naturally arise, including biology, drug discovery, recommender systems, social networks, and communication networks. In these settings, modelling different node and edge types through type-aware local spaces may lead to more accurate and parameter-efficient systems, reducing the computational cost of training and deployment. At the same time, heterogeneous graph learning methods can also have negative societal impacts when applied to sensitive relational data. In recommender systems or social networks, improved relational modelling may amplify existing biases, reinforce filter bubbles, or enable more effective profiling of users. In biomedical or decision-support settings, inaccurate predictions or biased training data may lead to unreliable conclusions if the model is used without domain-specific validation. Moreover, because heterogeneous graphs often encode rich relational information, privacy risks may arise if sensitive node or edge attributes are present. Our work is primarily methodological and does not target a specific high-stakes deployment. Nevertheless, practical use of \textsc{HetSheaf} should be accompanied by dataset-specific fairness, privacy, robustness, and interpretability analyses. In particular, when applied to human-centred domains, users should carefully assess whether type information introduces or amplifies protected-attribute correlations, and whether the resulting predictions are reliable across different subgroups.

\end{document}